\documentclass[10pt,twocolumn,letterpaper]{article}

\usepackage{wacv}
\usepackage{times}
\usepackage{epsfig}
\usepackage{graphicx}
\usepackage{amsmath}
\usepackage{amssymb}
\usepackage{booktabs}
\usepackage{multirow}
\usepackage[ruled,vlined]{algorithm2e}  %
\usepackage{tabularx}
\usepackage{xcolor,colortbl}
\usepackage{adjustbox}

\usepackage{caption}
\usepackage{subcaption}

\usepackage{listings}
\usepackage{xcolor}

\definecolor{codegreen}{rgb}{0,0.6,0}
\definecolor{codegray}{rgb}{0.5,0.5,0.5}
\definecolor{codepurple}{rgb}{0.58,0,0.82}
\definecolor{backcolour}{rgb}{0.95,0.95,0.92}

\lstdefinestyle{mycodestyle}{
  backgroundcolor=\color{backcolour}, commentstyle=\color{codegreen},
  keywordstyle=\color{magenta},
  numberstyle=\tiny\color{codegray},
  stringstyle=\color{codepurple},
  basicstyle=\ttfamily\footnotesize,
  breakatwhitespace=false,         
  breaklines=true,                 
  captionpos=b,                    
  keepspaces=true,                 
  numbers=left,                    
  numbersep=5pt,                  
  showspaces=false,                
  showstringspaces=false,
  showtabs=false,                  
  tabsize=2
}

\makeatletter
\@namedef{ver@everyshi.sty}{}
\makeatother
\usepackage{tikz,pgfplots}
\usetikzlibrary{shapes.arrows}

\colorlet{myGray}{black!5}
\newcolumntype{a}{>{\columncolor{myGray}}c}

\def\wacvPaperID{1060} %

\wacvalgorithmstrack   %
\wacvfinalcopy %

\ifwacvfinal
\usepackage[breaklinks=true,bookmarks=false]{hyperref}
\else
\usepackage[pagebackref=true,breaklinks=true,colorlinks,bookmarks=false]{hyperref}
\fi

\usepackage[capitalize,nameinlink]{cleveref}
\crefname{section}{Sec.}{Secs.}
\crefname{table}{Table}{Tables}
\crefname{appendix}{App.}{Apps.}

\renewcommand{\paragraph}[1]{\smallskip\noindent{\bf #1}~~}

\usepackage{xspace}
\def\eg{\textit{e.g.}\xspace}
\def\ie{\textit{i.e.}\xspace}

\def\etal{\textit{et al.}\xspace}
\usepackage[capitalize,nameinlink]{cleveref}
\crefname{section}{Sec.}{Secs.}
\Crefname{section}{Section}{Sections}
\Crefname{table}{Table}{Tables}
\crefname{table}{Table}{Tables}
\crefname{algorithm}{Alg.}{Algs.}

\usetikzlibrary{quotes,shapes.arrows,shapes.geometric,arrows.meta,decorations.markings,decorations.text}

\definecolor{mycolor0}{rgb}{0.2667,0.4471,0.7098}
\definecolor{mycolor1}{rgb}{0.1647,0.6706,0.3804}
\definecolor{mycolor2}{rgb}{0.8275,0.2627,0.3059}
\definecolor{mycolor3}{rgb}{0.5216,0.4392,0.7176}
\definecolor{mycolor4}{rgb}{0.8118,0.7255,0.4118}
\definecolor{mycolor5}{rgb}{0.2745,0.7176,0.8157}

\definecolor{mylcolor0}{rgb}{0.6902,0.7686,0.8863}
\definecolor{mylcolor1}{rgb}{0.5451,0.8902,0.6941}
\definecolor{mylcolor2}{rgb}{0.9412,0.7490,0.7647}
\definecolor{mylcolor3}{rgb}{0.8627,0.8392,0.9176}
\definecolor{mylcolor4}{rgb}{0.9569,0.9373,0.8667}
\definecolor{mylcolor5}{rgb}{0.7529,0.9020,0.9373}
\definecolor{mylcolor6}{rgb}{0.8750,0.8750,0.8750}
\begin{document}

\title{Expansion of Visual Hints for Improved Generalization in Stereo Matching}

\author{
Andrea Pilzer$^1$\thanks{Work done while at Aalto University} \qquad Yuxin Hou$^{2,3}$ \qquad Niki Loppi$^1$ \qquad Arno Solin$^2$ \qquad Juho Kannala$^2$ \\[6pt]
\begin{minipage}{.3\textwidth}\centering
$^1$NVIDIA \\
\end{minipage}
\hfill
\begin{minipage}{.3\textwidth}\centering
$^2$Aalto University \\
\end{minipage}
\hfill
\begin{minipage}{.3\textwidth}\centering
$^3$Niantic \\

\end{minipage}%
\\[6pt] 
{\small\tt \{apilzer,nloppi\}@nvidia.com, \{yuxin.hou, arno.solin, juho.kannala\}@aalto.fi}
}

\maketitle
\thispagestyle{empty}

\begin{abstract}
We introduce visual hints expansion for guiding stereo matching to improve generalization. Our work is motivated by the robustness of Visual Inertial Odometry (VIO) in computer vision and robotics, where a sparse and unevenly distributed set of feature points characterizes a scene. To improve stereo matching, we propose to elevate 2D hints to 3D points. These sparse and unevenly distributed 3D visual hints are expanded using a 3D random geometric graph, which enhances the learning and inference process. We evaluate our proposal on multiple widely adopted benchmarks and show improved performance without access to additional sensors other than the image sequence. To highlight practical applicability and symbiosis with visual odometry, we demonstrate how our methods run on embedded hardware.
\end{abstract}

\section{Introduction}
\label{sec:intro}

Accurate depth estimation is an important task for many 3D applications such as AR/VR and robotics navigation. Technological advances have made active depth sensors (\eg, LiDaR) affordable. 
Yet, they have the drawback of providing only sparse depth maps.
Algorithmic techniques are still more common for dense depth predictions, where deep learning approaches \cite{eigen2015predicting, garg2016unsupervised, liang2018learning, chang2018pyramid, Guo_2019_CVPR, Duggal2019ICCV, kendall2017end, godard2017unsupervised, khamis2018stereonet} have overtaken traditional matching techniques \cite{boykov2001fast, marroquin1987probabilistic, scharstein1998stereo, Hannah1974ComputerMO, jisct1993stereo, 10.1007/BFb0028345}, as their accuracy has kept improving with bigger annotated data sets available \cite{Menze2015ISA, Menze2018JPRS, sceneflow}.  Nevertheless, techniques based on geometric computer vision are still viable options for scarce data or for ensuring good out-of-domain performance. 

Purely image-based dense 3D reconstructions can be computed using state-of-the-art photogrammetry software (\eg, Metashape, ReCap Pro). However, many of these photogrammetry tools do not perform robustly if the scene has numerous textureless surfaces, such as in typical indoor office spaces. They also require high-resolution images to match fine surfaces resulting in high computational costs. %
In practice, visual-inertial odometry (VIO) and simultaneous localization and mapping (SLAM) techniques \cite{murORB2, seiskari_2021_arxiv} are typically employed for real-time camera motion estimation (\eg, ARKit, ARCore, \etc) as well as in classical computer vision pipelines, which reconstruct sparse point cloud models based on matching of local features between multiple registered images (\eg, COLMAP \cite{schoenberger2016mvs, schoenberger2016sfm}).

\begin{figure}[t]
    \centering
    \resizebox{\columnwidth}{!}{%
    \input{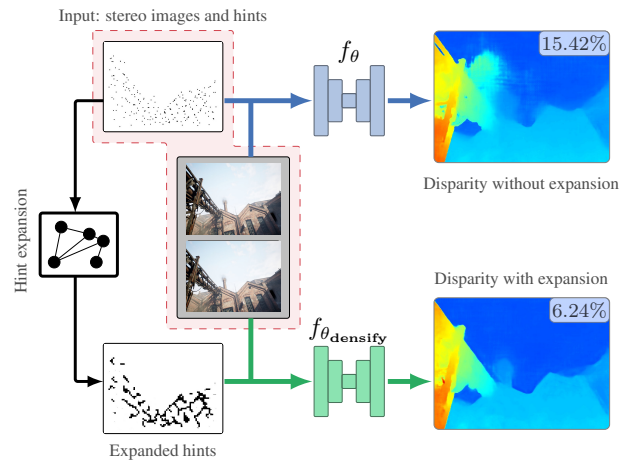}}
    \vspace{-6pt}
    \caption{Visual hints expansion for deep stereo matching. (\textbf{top}) Inference with a model $\theta$ trained with sparse visual hints, and  (\textbf{bottom}) with a model $\theta_{\mathbf{densify}}$ trained on expanded visual hints. 3D expanded hints lead to more accurate predictions: the overlay labels show the error rate ${>}$3.}
    \label{fig:teaser}
    \vspace{-4pt}
\end{figure}

\begin{figure*}[t]
    \centering
    \resizebox{\textwidth}{!}{%
    \input{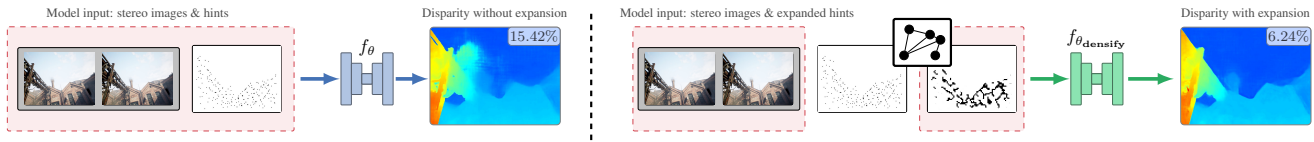}}
    \vspace{-18pt}
    \caption{Guided Stereo Matching Pipeline. \textbf{Left} vanilla guided stereo matching, model inputs are stereo images and hints (from VIO or LiDaR). \textbf{Right} expanded hints for guided stereo matching, model inputs are stereo images and expanded hints.}
    \label{fig:pipeline}
    \vspace{-4pt}
\end{figure*}

In this work, we propose VIO guidance for improved robustness and more accurate predictions of stereo matching pipelines on data with domain shift (see \cref{fig:teaser}). We argue that by exploiting synthetic data sets, stereo matching algorithms have increased their performance but may need costly fine tuning---typically on real data. Our work hinges on the realization, that visual localization methods (\eg, SLAM) can provide a valuable source of sparse 3D world information to guide our algorithm towards accurate dense predictions. Seminal work in this direction was presented by Poggi \etal \cite{Poggi_CVPR_2019}, who considered sparse uniformly distributed guidance (\eg, from a LiDaR), to guide feature matching at lower scale (\ie to build the cost volume) \cref{fig:pipeline}\textit{-left}. However, our work has two key differences. First, visual guidance is sparse and non-uniformly distributed, ranging from tens to a few hundred points. Second, visual guidance may be imprecise at some locations, which requires filtering to improve robustness. %
Sparse and uneven VIO hints at lower scale could be discarded due to downsampling. Therefore, an expansion is used to improve guidance and our quantitative and qualitative experiments prove its effectiveness. 

To address the previously mentioned challenges, we propose a 3D graph based hints expansion \cref{fig:pipeline}\textit{-right}. Expansion of sparse hints has been considered before by, \eg, Huang \etal \cite{Huang_2021_CVPR} who proposed a constant expansion for close points. We argue this is a too strict condition, and a slanted linear expansion as in \cite{bleyer2011patchmatch} is more suitable. %
Therefore, we consider the guidance points not in 2D but in 3D. Intuitively, close points on the 2D image plane may be very far in their 3D position. %
With this in mind, we turn hints into nodes and connect them with edges only if they are close in the 3D world. After obtaining the graph, we linearly approximate the disparity with a 3D slanted line following the edges. %
In our case, slanted 3D lines do not add any additional computational overhead other than assigning the corresponding disparity value to the pixels on the graph edges. 
Furthermore, we devise a 3D linear approximation of our graph expansion. By building upon  a heuristic that searches for pixels only along vertical and horizontal lines and dividing the images in non-overlapping patches, our approach becomes efficient in connecting hints with 3D slanted lines. %

We leverage our expanded hints for deep stereo matching with DeepPruner \cite{Duggal2019ICCV}, where their role is to guide the differentiable patch match in a narrow range instead of the full disparity range. We demonstrate that sparse visual guidance is accurate enough to lead the model towards correct predictions. Extensive experimental results show that expanded guidance improves the performance of deep stereo matching on unseen data sets at training time. Following \cite{Poggi_CVPR_2019}, we also demonstrate the proposed contributions on PSMNet \cite{chang2018pyramid}. However, expanded visual hints may contain errors, and to this end, we propose a confidence-based guidance filtering. In PSMNet, feature activation maps of both stereo images are extracted to build the 3D cost volume. We filter hints if the feature of the depth hint in the reference frame (\ie\ left stereo view) is not similar-enough to the corresponding feature in the right stereo view. %

We summarize our contributions as follows:
{\em (i)}~We propose a novel 3D graph based hints expansion which circumvents pitfalls in previous expansion methods by considering the 3D neighbourhood of the guidance.
{\em (ii)}~We devise a 3D linear approximation of our graph guidance, and show that this is an efficient heuristic.
{\em (iii)}~We leverage our expanded hints on DeepPruner \cite{Duggal2019ICCV}, where their role is to guide the differentiable patch match in a narrow range instead of the full disparity range.
{\em (iv)}~Expanded visual hints may still be error prone, and we propose a confidence based guidance filtering method for improved robustness. 

\section{Related Work}
\label{sec:related}

\textbf{Stereo Matching} has a long history in computer vision and has well established benchmarks \cite{middlebury, schoeps2017cvpr}. Traditional methods are based on local \cite{Hannah1974ComputerMO, jisct1993stereo, 10.1007/BFb0028345} or global matching \cite{boykov2001fast, marroquin1987probabilistic, scharstein1998stereo}. Local algorithms are faster and work well in textured regions, while global matching algorithms are computationally more expensive and work well also in textureless regions. Recently, deep learning based methods \cite{liang2018learning, chang2018pyramid, Guo_2019_CVPR, Duggal2019ICCV, kendall2017end, pami2020pilzer} have shown superior performance over traditional methods. Deep architectures typically use convolutional neural networks (CNN) as feature extractors, U-Net style architectures \cite{garg2016unsupervised, eigen2015predicting, godard2017unsupervised}, and cost volumes aggregation \cite{kendall2017end, khamis2018stereonet, chang2018pyramid}. In order to improve cost volume construction, Guo \etal \cite{Guo_2019_CVPR} proposed a feature correlation method. At the same time, Duggal \etal \cite{Duggal2019ICCV} revisited patch matching in a differentiable way allowing end-to-end learning of CNNs. HITNet \cite{Tankovich_2021_CVPR} showed that slanted surfaces allow for smooth and accurate disparity prediction. We take inspiration from their work in devising a hint expansion algorithm that %
exploits the local slanted nature of disparity.

\textbf{Guided Stereo Matching.} CNNs offer accurate dense predictions but suffer of overconfidence and domain-shift with out-of-distribution (OOD) data. Poggi \etal \cite{Poggi_CVPR_2019} proposed a first attempt to address these issues through a feature enhancement method that exploits sparse LiDaR guidance. Later, \cite{Huang_2021_CVPR} further extended \cite{Poggi_CVPR_2019} by expanding sparse guidance and learning where to confidently use it. Unlike them, we employ visual hints as a sparse and unevenly distributed guidance. This poses a greater challenge as we cannot assume neighbouring pixels to have a similar disparity. On the other hand, long term tracking of sparse feature points in VIO can provide accurate triangulation and help solving ambiguous regions in two-view stereo. VIO algorithms find strong features to match on edges or corners, where also disparity quickly transitions. Quick transitions are harder to model. Bleyer \etal \cite{bleyer2011patchmatch} proposed slanted planes as a locally linear approximation.

Beyond stereo matching, Sinha \etal \cite{sinha2020deltas} proposed a learning-based triangulation method for dense multi-view stereo depth. Wong \etal \cite{wong2020unsupervised} use \emph{scaffolding}, based on convex hulls, as part of  monocular depth completion. However, we do not use convex hulls but propose our own 3D graph approximation. %
For 3D object detection, \cite{you2020pseudo} developed a LiDaR to pseudo-LiDaR depth refinement based on graph neural networks. Nonetheless, our algorithm is orthogonal to theirs as 3D hint expansion are not learned.

We leverage existing VIO algorithms. We focus on how to effectively exploit sparse visual hint guidance for deep stereo matching. VIO \cite{bloesch2015robust, seiskari_2021_arxiv, sun2018robust, rosinol2020kimera} and SLAM \cite{bai2019novel, rosinol2020kimera, usenko2019visual, Schops_2019_CVPR} are core for navigation. The CV community has  pushed to make these methods fast, robust, and general. Thus, the sparse hints are agnostic to use case and generally their errors are uncorrelated to learned depth methods. 

\section{Methods}
\label{sec:method}

Expanding hints is not trivial since VIO algorithms return hints at image keypoints, \eg~edges or corner points. These areas are difficult to expand, because of the geometry around the feature point. While a wrong value could be filtered (as in \cite{Huang_2021_CVPR}), it is important to note that it will also not have positive impact.
Hints expansion is further motivated by the structure of deep CNNs for stereo matching. Feature representations of the input images have smaller spatial size compared to the image itself, leading to a difficult, and in some case impossible, alignment of the sparse hints with the downsampled image grid. This can be avoided by expanding the hints over a larger area. 

In our case, in addition to the stereo image pair, sparse unevenly distributed hints are available as guidance for our model. They are encoded in sparse matrices $H_n$, each corresponding to a stereo image pair $(I^\mathrm{R}_n,I^\mathrm{L}_n)$. We aim at thoroughly exploiting the rich information regarding the 3D world encoded in the hints by lifting them to 3D points and not simply modelling them as disparity values on a 2D matrix. The 3D hypothesis prompts us to more clearly discriminate hints that are close to each other. 

\textbf{Notation} Consider a data set $\mathcal{D}=\{({I}^{\mathrm{L}}_n,{I}^{\mathrm{R}}_n, H_n)\}_{n=1}^{N}$, where $N$ is the number of data samples, ${I}^{\mathrm{L}}_n,{I}^{\mathrm{R}}_n$ are stereo image pairs, and $H_n$ are sparse disparity hints. We learn a model with parameters $\theta$ to infer a dense disparity map $\mathbf{disp}$ such that $\mathbf{disp}_n = f_{\theta}(I^{\mathrm{L}}_n,I^{\mathrm{R}}_n, H_n)$. We assume $H_n$ to be a sparse matrix of visual hints, and that an expansion function $\mathbf{densify}(H)$ exists such that the number of hints increases. The function $\mathbf{densify}$ should also lead to more accurate estimation of the disparity $\mathbf{disp}_n = f_{\theta_{\mathbf{densify}}}(I^{\mathrm{L}}_n,I^{\mathrm{R}}_n, \mathbf{densify}(H_n))$.

\subsection{3D Linear Hints Expansion (Lin 3D)}
\label{subsec:fast_3d}

We propose a linear densification based on two core steps. {\em (i)}~The first, is to connect two hints if they are found on the same horizontal or vertical axis. Hints are connected with a slanted linear interpolation as in \cite{bleyer2011patchmatch}. {\em (ii)}~Secondly, the image is split into non-overlapping square patches of fixed size. In this way, the expansion process transforms into a patch-wise line-by-line (first horizontal and then vertical) search of hints that can be connected. %
Detailed algorithm and implementation is available in the supplementary. %

\begin{figure}[t!]
    \centering
    \includegraphics[width=0.8\columnwidth]{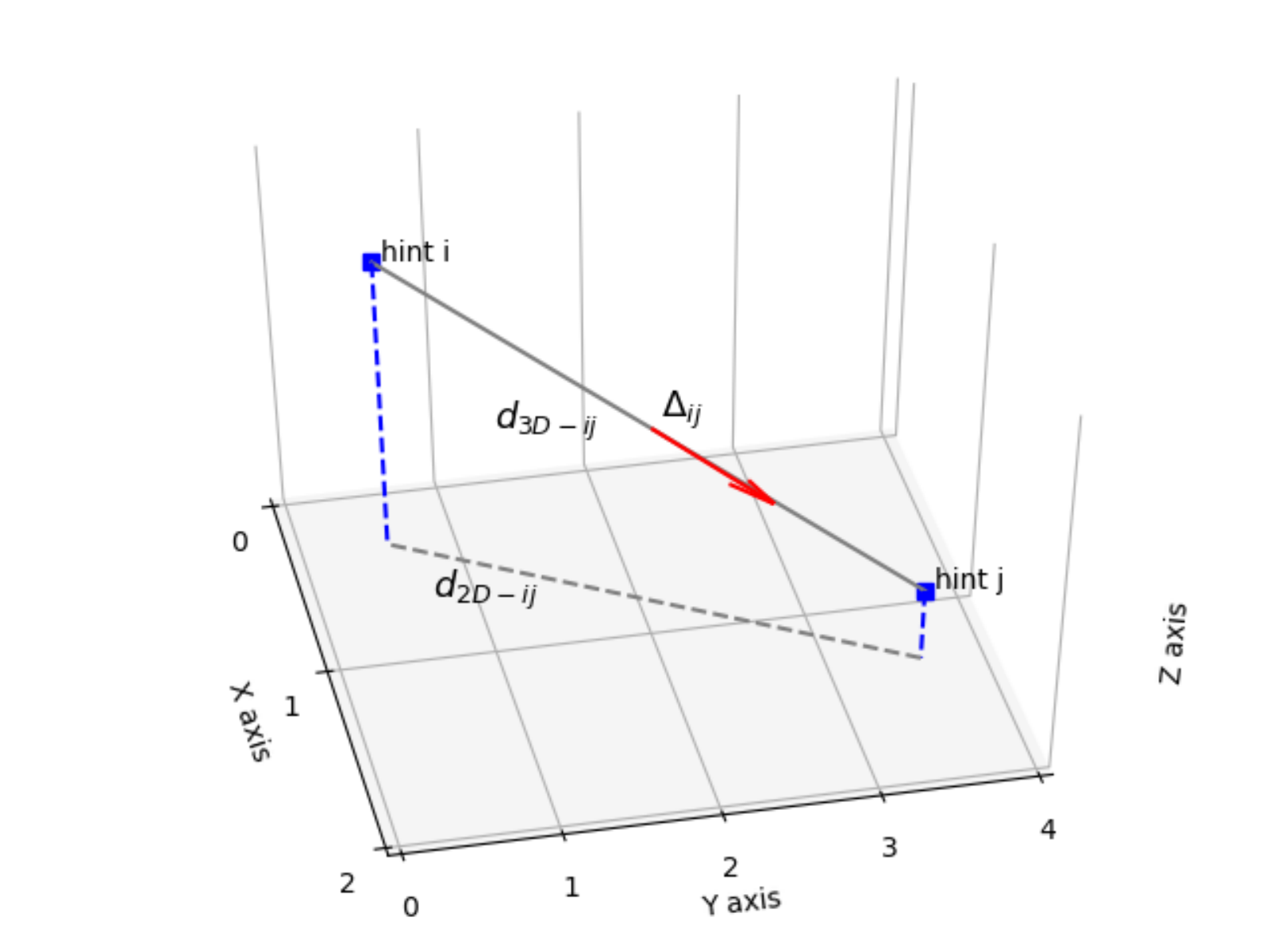}
    \caption{Hints $i$ and $j$ are shown as blue dots in the center of their corresponding pixels, their distance in 3D space and 2D are $d_{3D-ij}$ (gray line) and $d_{\mathrm{2D}-ij}$ (gray dashed line), respectively. The red arrow represents the direction $\Delta_{ij}$ used to travel between the hints. $Z$ represents the depth.}
    \vspace{-9pt}
    \label{fig:3d_segment}
\end{figure}

\subsection{3D Graph Hints Expansion (3D $\mathcal{G}$)}
\label{subsec:expansion}
To improve over 3D linear hints expansion, we propose to build a 3-dimensional Random Geometric Graph ($\textbf{RGG}$, \cite{geom_graph, CLARK1990165}), where the hints are nodes and edges connect hints that are close in 3D. Connections are further constrained by color similarity between nodes (each node has a corresponding RGB value in ${I}^{\mathrm{L}}_n$, ${I}^{\mathrm{L}}_n$ and $H_n$ are aligned). Formally, an edge will be created if
\begin{equation}
    E_{ij}: (d_{\mathrm{3D}-ij} < R) \land (RGB_i \odot RGB_j > \tau),
\end{equation}
where $E_{ij}$ is the edge connecting hints $i$ and $j$, $d_{\mathrm{3D}-ij}$ is their (Euclidean) distance, and $R$ is the maximum distance. $RGB_{i,j}$ are the RGB values in the left image at the location of the hints, $\odot$ is the cosine similarity between the two color vectors and $\tau$ is a threshold of similarity ($\tau=0.9$). Note that $d_{\mathrm{3D}-ij}$ is calculated in 3D coordinates, and we will refer to it also as volumetric distance to help distinguish it from 2D spatial distance in the 2D image plane, denoted as $d_{\mathrm{2D}-ij}$. The process is visualized in \cref{fig:3d_segment}, where disparity is shown along the $Z$-axis. Once the edges are created, they are {\em (i)}~sorted by volumetric 3D distance and {\em (ii)}~the adjacent ones are discarded. While {\em (ii)} is trivial, between two spatially (2D) adjacent pixels there are no further pixels to possibly assign an expanded disparity value. The first step {\em (i)} is grounded on a locally linear approximation of the real disparity, a similar hypotheses to \cite{bleyer2011patchmatch}, \cite{wong2020unsupervised}. Therefore, the shorter edges are expanded first as they are more likely to accurately model the real disparity.

The expansion for each edge is performed by travelling one spatial unit distances $d_{\mathbf{1}-\mathrm{2D}}$ at a time between the two hints in the 2D image plane, where their 2D Euclidean distance is denoted $d_{\mathrm{2D}-ij}$. We define $E_{ij}$ be the edge between hints $i$ and $j$, with distances $(d_{\mathrm{3D}-ij},d_{\mathrm{2D}-ij})$ and coordinates $\{x_i,y_i,z_i\}$, $\{x_j,y_j,z_j\}$, respectively. They are connected by a slanted 3D line with a slope $\Delta_{ij}$ (red arrow in \cref{fig:3d_segment}) 
\begin{equation}
    \Delta_{ij} = \begin{cases}
        \delta_x = \cos(\arctan((y_i-y_j)/(x_i-x_j))), \\
        \delta_y = \sin(\arctan((y_i-y_j)/(x_i-x_j))), \\
        \delta_z = (z_i-z_j)/((y_i-y_j)/(x_i-x_j)).
    \end{cases}
\end{equation}
Moving by $m$ steps of unit distances $d_{\mathbf{1}-\mathrm{2D}}$ until $d_{\mathbf{1}-\mathrm{2D}} m < d_{\mathrm{2D}-ij}$, the corresponding $z$ disparity value is assigned to the closest hint pixel $\mathbf{h}=\mathcal{H}[\delta_x m,\delta_y m]$ by rounding the computed coordinates to the closest integer. Formally, the loop writes
\begin{equation} \label{eq:loop}
    \!\!\!\{H[\textbf{r}(\delta_x m), \textbf{r}(\delta_y m)] = \delta_z m ~| ~m < d_{\mathrm{2D}-ij}, m \in \mathcal{N} \},
\end{equation}
where $\textbf{r}$ is the rounding to the closest integer operation, and $m$ is the set of unit distances. Note that a value is assigned only if the corresponding pixel $\mathbf{h}$ is empty, in order to preserve original hints and values already assigned in \cref{eq:loop}. We present the graph expansion algorithm and the implementation in the supplementary material for more clarity. %

\subsection{DeepPruner}
\label{subsec:generalize}

Our aim is to show that expanded VIO hints are effective in enhancing model performance on unseen sequences. Given expanded sparse hints, a model can consider fewer candidate disparities around the suggested values and return more accurate predictions. To demonstrate this, we implemented a novel framework based on DeepPruner \cite{Duggal2019ICCV}. 
DeepPruner proposed pipeline consists of four parts: {\em (i)}~a PatchMatch-based module first compute matching scores from a small subset of disparities. {\em (ii)}~ Then given the estimation from PatchMatch, there is a confidence range prediction network to adjust search range for each pixel. {\em (iii)}~A sparse cost volume will be constructed and aggregated within the predicted range. {\em (iv)}~Finally, an image guided refinement module uses low-level details to further improve the prediction. 

Since we have sparse hints $H$ as extra inputs, we can modify the first step and skip the iterative PatchMatch; instead, we use a simpler sampling strategy to compute the sparse matching scores. For a sparse hints matrix $H_n$
\begin{equation}
    \!\!
    \mathbf{range}_n {:}
    \begin{cases}
          d_\mathrm{low} {=} (1 - V_n) \, d_{\min} + V_n \, H_n \, (1 - \alpha) \\
          d_\mathrm{high} {=} (1 - V_n) \, d_{\max} + V_n \, H_n \, (1 + \alpha)
    \end{cases}
\end{equation}
where $d_\mathrm{low}, d_\mathrm{high}$ are the lower bound and the upper bound of the search range $\mathbf{range}_n$, respectively. $V=(H>0)$ indicates if the pixels have hints and $\alpha$ is a hyper-parameter to control the relative range. We set $\alpha = 0.2$ to accommodate a small margin of error for our densified visual hints, and $(d_{\min},d_{\max})$ are the minimum and maximum disparities allowed for the $\mathbf{range}$ during training and testing. $(d_{\min},d_{\max})$ are used in case no hints is available at a given location. Then we sample disparity candidates uniformly from $[d_\mathrm{low}, d_\mathrm{high}]$ for each pixel to compute the matching scores. %
Even if it falls out of the scope of this paper, we believe such guidance approach is straight-forward to apply to several traditional stereo matching methods based on patch matching.

\subsection{3D Expansion for Guided Stereo Matching}
\label{subsec:guided}

\paragraph{Guided Stereo Matching for PSMNet}
Guided stereo matching  (GSM) \cite{Poggi_CVPR_2019} proposed to guide learning with sparse supervision points, which we refer to as hints in this work. Their method is based on a scaling factor that promotes learning for the pixels of the image where the ground truth is present. Namely, in a model with cost volume, this is formulated as
\begin{equation}
    G_n = \bigg(1 - V_n + V_n \, k \, \exp\bigg(- \frac{d-H_{n}}{2c}\bigg)\bigg) F_n ,
\end{equation}
where $F_n$ are the original cost volume features, $G_n$ are the enhanced output features, $H_{n}$ are the known hints, and $V_n = (H_n > 0)$ is a binary mask specifying which pixel will be enhanced. The Gaussian modulation is parameterized by magnitude $k=10$ and variance $c=1$.

\paragraph{Confidence-guided Stereo Matching for PSMNet}
Although guided stereo matching proved to be effective on an uniformly sampled ground-truth, in our setting VIO would be the source of the guiding points. This leads to relevant differences to the previous setting. 
First, our hints will be localized in areas with keypoint presence, effectively invalidating the assumption of uniformly sampled hints. 
Second, VIO hints and the expanded hints generated by the proposed expansions may be imprecise. 
Third, the density is lower than with sparse supervision. %

\begin{figure*}[t!]
  \centering
  \setlength{\mywidth}{0.33\textwidth}%
  
  \newcommand{\plot}[4]{%
    \node[anchor=north west,draw=black!50,minimum width=\mywidth,
      minimum height=.53\mywidth,rounded corners=2pt,path picture={
      \node at (path picture bounding box.north west){
        \includegraphics[width=\mywidth]{#3}};
      }] (fig) at (#1,#2) {};
    \node[anchor=north west,draw=black!50,align=left,inner sep=1pt,
      fill=white,rounded corners=2pt,minimum height=10pt,opacity=.75] at 
      (fig.north west) {\footnotesize #4};
  }

    \begin{subfigure}[t]{\textwidth}
      \centering
      \begin{tikzpicture}[inner sep=0,outer sep=0]      
        \plot{0}{0}{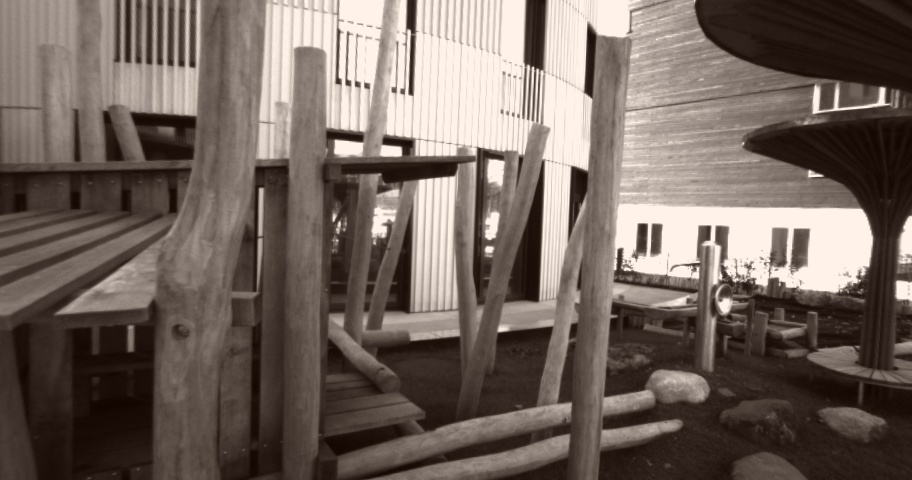}{RGB}
        \plot{1.02\mywidth}{0}{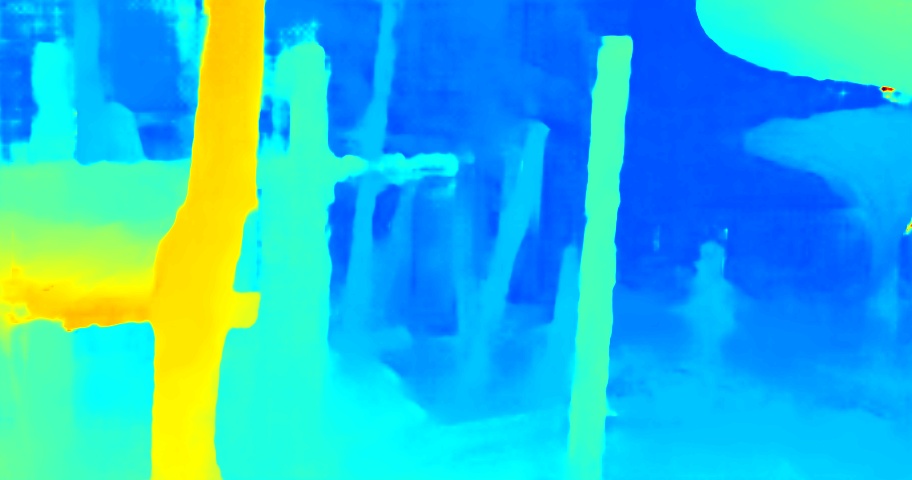}{PSMNet}
        \plot{2.04\mywidth}{0}{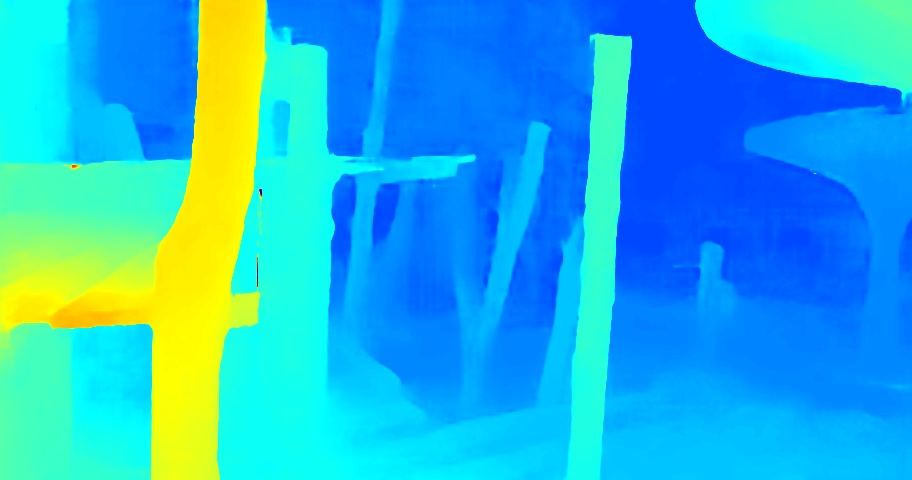}{\textit{3D $\mathcal{G}$ exp}}
      \end{tikzpicture}
    \end{subfigure}
    \caption{PSMNet qualitative results on {\sc ETH3D}. Our expansion sharpens details. %
    }
    \label{fig:psmnet_qualitative}
    \vspace{-6pt}
\end{figure*}
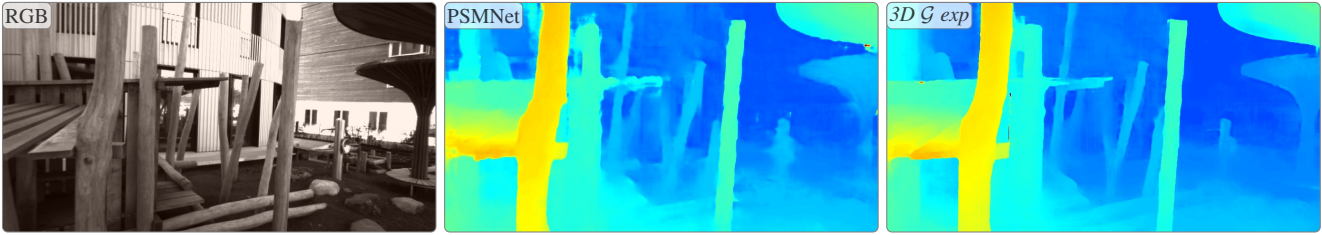

\begin{table*}[t]
    \centering
    \footnotesize
    \setlength{\tabcolsep}{4pt}
    \begin{tabularx}{\textwidth}{l|c|cc|ccc|ccc}
    \toprule
         \sc Data set & \sc Img size & \sc Vgd [density] & MAE & \sc 3D $\mathcal{G}$ exp [density] & MAE & Param $R$ & \sc Lin 3D exp [density] & MAE & Param $W$ \\
    \midrule
         \sc SceneFlow& (256,512)$^\circ$& 440 [0.33\%]& 1.59 & 8690 [6.6\%]& 1.89 & $R=8\phantom{0}$ & 8840 [6\%] & 2.08 & $W=(8,16)$ \\
         \sc Tartan& (480,640)& 13 [0.004\%]&1.43 & 849 [0.27\%]&1.57 & $R=25$ & 113 [0.03\%] & 1.63 & $W=(8,16)$ \\
         \sc ETH3D& (544,960)& 263 [0.05\%]&2.06 & 3703 [0.7\%]&2.16 & $R=8\phantom{0}$ & 4627 [0.88\%] & 2.34 & $W=(8,16)$\\
         \sc KITTI15& (368,1232)& 562 [0.1\%]& 7.02 & 40494 [8.9\%]& 7.47 & $R=20$ & 5271 [1.1\%] & 7.77 & $W=(8,16)$ \\
   \bottomrule
    \end{tabularx}
    \caption{Sparsity of the visual guidance hints for each data set. Average hints per image over all the data sets, [density in brackets], MAE hints error, and expansion parameters. `Img size' is the image size in pixels, `$\circ$' refers to training data. %
    The proposed expansions do not introduce noticeable additional error while greatly improving model prediction performance.
    }
    \vspace{-6pt}
    \label{tab:exp_ablation}
\end{table*}

To address the noisy hints, we devise a confidence-based hints filtering for 3D cost volume based architectures as PSMNet \cite{chang2018pyramid}. In our scarce supervision scenario, it is particularly important to ensure that hints are positively contributing in the pipeline. The confidence is implemented as an Euclidean distance between feature vectors. Given the normalized feature maps of the left stereo view $f_\mathrm{L}$ and the feature maps of the right stereo view $f_\mathrm{R}$, we compare the left feature at hint position $(x_H,y_H)$ with the corresponding right feature as suggested by the hint,
\begin{multline}
    \mathrm{conf}_{H[x_H,y_H]} = 1 - \tanh(\lVert f_\mathrm{L}(x_H,y_H) - \\ f_\mathrm{R}(x_{H}+H[x_{H},y_{H}],y_H) \rVert^2),
\end{multline}
where $H[x_H,y_H]$ is one of our hints. This process is trivially repeated for all hints. Subsequently, the final mask is obtained as follows
\begin{equation}
    V = \begin{cases}
1, &\text{where $\mathrm{conf}_{H} > \tau \land H > 0$,}\\
0, &\text{where $\mathrm{conf}_{H} < \tau \lor H > 0$,}
\end{cases}
\end{equation}
and it is directly applied to the hints for filtering noisy ones ($\tau=0.9$). A further minor modification to PSMNet is an additional loss term. To mitigate the over-smoothing problem \cite{Chen_2019_ICCV}, we add a classification loss on the disparity. It is implemented as negative log-likelihood (NLL) loss between the ground truth disparity and the model prediction. Both of them are rounded to the closest integer in order to satisfy the classification objective constraints.

\section{Experiments}
\label{sec:exp}

We evaluate the proposed expansion methods in \cref{subsec:exp_ablation} and employ it as guidance for DeepPruner in \cref{par:deeppruner} and PSMNet in \cref{par:psmnet}. %
We also deploy it on embedded devices in \cref{subsec:embedded} and compare against guided methods in \cref{subsec:sota}.

\textbf{Data sets} We use {\sc SceneFlow} \cite{sceneflow}, {\sc ETH3D} \cite{schoeps2017cvpr}, \cite{Schops_2019_CVPR}, {\sc Tartan} \cite{tartanair2020iros}, {\sc KITTI} \cite{Menze2015ISA}, \cite{Menze2018JPRS} data sets for our experiments. Detailed information about each data set is available in the supplementary material.
As in \cite{chang2018pyramid}, \cite{Poggi_CVPR_2019}, \cite{Huang_2021_CVPR} models are evaluated with MAE and threshold error rate. They are computed for every pixel where ground-truth is available, with thresholds $t \in \{2,3,4,5\}$. %

\textbf{Implementation} We implement our method with PyTorch. Training is performed on a single NVIDIA A100 GPU. With {\sc Sceneflow} \cite{sceneflow} training takes about 25 hours for PSMNet \cite{chang2018pyramid} and 125 hours for DeepPruner \cite{Duggal2019ICCV}. We follow training guidelines from the respective papers. We did not notice any training time difference with sparse hints or expanded hints. Testing is performed on the same device and on a NVIDIA Jetson AGX Xavier device. The capability of the method to run on embedded devices is of particular interest for future practical uses. %

\textbf{Model Details} \textit{PSMNet} \cite{chang2018pyramid} and \textit{DeepPruner} \cite{Duggal2019ICCV} are the baseline models. \textit{Vgd-test} and \textit{Lgd-test} refer to visual guidance and LiDaR guidance at test time applied to the original model. \textit{Vgd} refers to visual hints guided training and testing as introduced in \cite{Poggi_CVPR_2019}. Note that a crucial difference lies in the hints used, we have sparse visual hints and not the LiDaR hints of \cite{Poggi_CVPR_2019}. %
\textit{3D $\mathcal{G}$ exp} and \textit{Lin 3D exp} are the two expansion algorithms proposed in our work. PSMNet models that use expansion also use our confidence filtering. %

\subsection{Expansion Details}
\label{subsec:exp_ablation}

\begin{figure*}[t!]
	\centering
	\setlength{\mywidth}{0.335\columnwidth}%
	
	\newcommand{\plot}[4]{%
		\node[anchor=north west,draw=black!50,minimum width=\mywidth,
		minimum height=.75\mywidth,rounded corners=2pt,path picture={
			\node at (path picture bounding box.north west){
				\includegraphics[width=\mywidth]{#3}};
		}] (fig) at (#1,#2) {};
		\node[anchor=north east,draw=black!50,align=left,inner sep=1pt,
		fill=white,rounded corners=2pt,minimum height=10pt,opacity=.75] at 
		(fig.37) {\footnotesize #4};
	}
	
	\begin{subfigure}[t]{0.49\textwidth}
		\centering
		\begin{tikzpicture}[inner sep=0,outer sep=0]      
			\plot{0}{0}{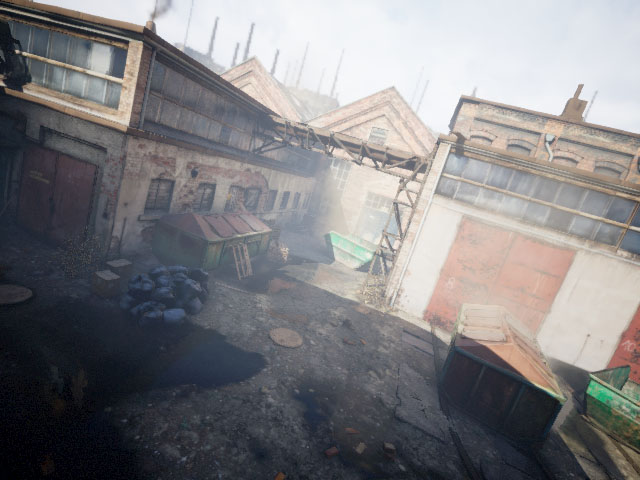}{RGB}
			\plot{1.02\mywidth}{0}{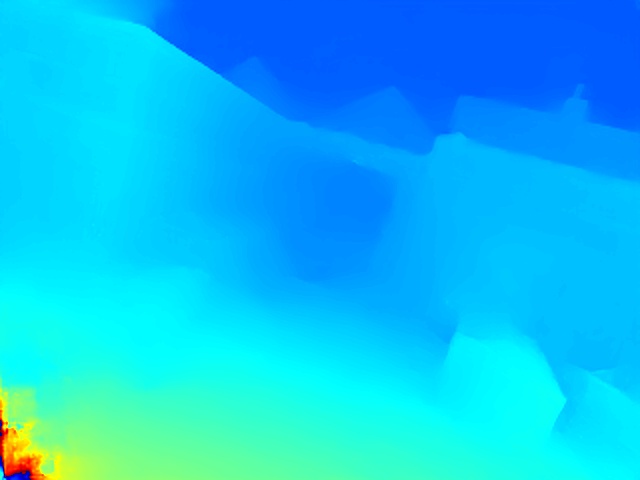}{DP}
			\plot{2.04\mywidth}{0}{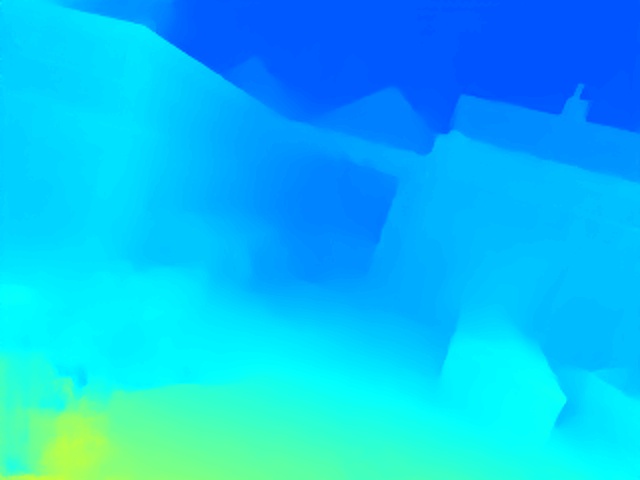}{\textit{3D $\mathcal{G}$ exp}}
		\end{tikzpicture}
	\end{subfigure}
	\hfill
	\begin{subfigure}[t]{0.49\textwidth}
		\centering
		\begin{tikzpicture}[inner sep=0,outer sep=0]      
			\plot{0}{1}{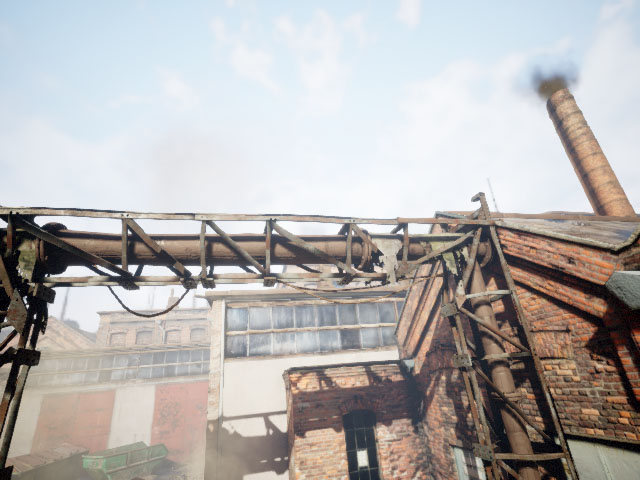}{RGB}
			\plot{1.02\mywidth}{1}{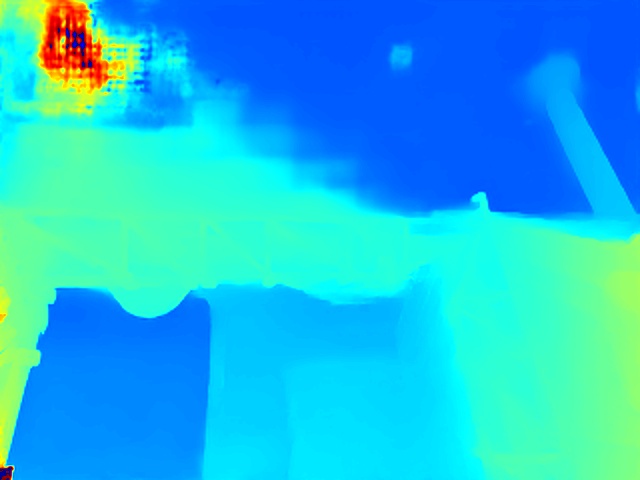}{DP}
			\plot{2.04\mywidth}{1}{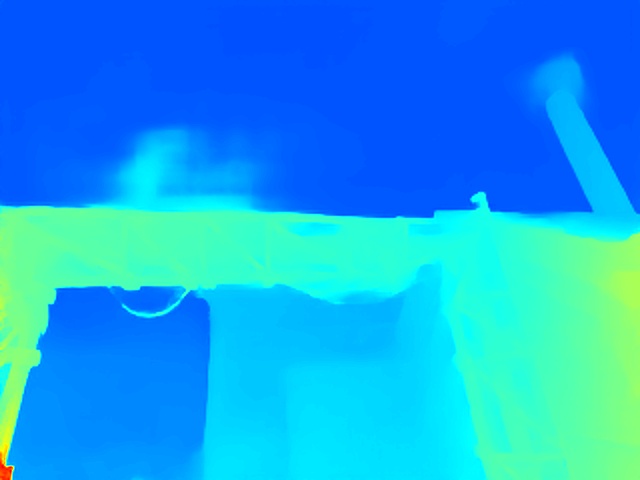}{\textit{3D $\mathcal{G}$ exp}}
		\end{tikzpicture}    
	\end{subfigure}
	\caption{DeepPruner (DP) qualitative results on the {\sc Tartan} data set. Above, \textit{3D $\mathcal{G}$ exp} sharpens some building edges, below, removes domain-shift artifacts.}
	\label{fig:dp_qualitative}
	\vspace{-12pt}
\end{figure*}
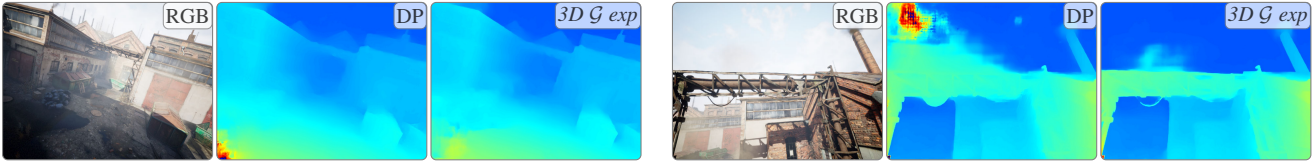

Before looking into actual model performance, we analyze the impact of the proposed expansions. VIO does not provide a uniformly spaced grid of sparse hints as a LiDaR would. Nevertheless, thanks to our densification we are able to achieve a much higher hints density. In \cref{tab:exp_ablation} (upper half) we denote the original VIO hints as \textit{Vgd}, our proposed 3D Graph expansion as \textit{3D $\mathcal{G}$ exp} and linear expansion as \textit{Lin 3D exp}. Generally a $10 {\times}$ increase in hints can be observed when our proposed densification algorithms are applied. Particularly interesting is the case of the {\sc Tartan} data set, where the average hints over all the images is only $13$ hints, but after expansion it grows to $113$ and $849$ hints, for \textit{Lin 3D exp} and \textit{3D $\mathcal{G}$ exp}, respectively. In square brackets we report the density of hints over the total number of pixels in the images. Density better conveys the sparsity or to some extent `rare' nature of visual hints that our method expands. For example, even after expansion both {\sc Tartan} and {\sc ETH3D} do not pass the $1\,\%$ threshold, while {\sc SceneFlow} and {\sc KITTI} do. 
In \cref{tab:exp_ablation} (bottom half), we detail expansion parameters. Both \textit{3D $\mathcal{G}$ exp} and \textit{Lin 3D exp} algorithms are implemented as functions that take in input hints of one image and output expanded hints. \textit{3D $\mathcal{G}$ exp} has one parameter, the maximum 3D radius $R$, which specifies the search range for nodes to connect. \textit{Lin 3D exp} has one parameter, which specifies the size of the patches $W$ to use for potential hints expansion. We assign these two parameters to obtain a good trade-off between final density, proximity of the hints and the chance of even finding a hint due to the low guidance density. \textit{3D $\mathcal{G}$ exp} is applied once, whereas \textit{Lin 3D exp} is applied in two iterations the first with $W_1=8$ and the second with $W_2=16$. %

In \cref{tab:exp_ablation} we report the proposed expansions MAE errors, it is interesting to note that the expansions introduce a minor increase in MAE errors but not significantly higher than the initial VIO hints noise. {\sc KITTI15} is the only data sets that stands out with a higher MAE error. This is due to the short sequences that compose the data sets where VIO hits are more challenging to extract and align compared to the longer sequences of the other data sets.

\subsection{Ablation Studies}
\label{subsec:ablation}

\textbf{DeepPruner} \label{par:deeppruner} \cref{tab:exp_deeppruner} demonstrates that expanded guidance is beneficial for DeepPruner \cite{Duggal2019ICCV}. The \textit{Vanilla} model uses a checkpoint and code from the authors official repository. Visual guidance \textit{Vgd} improves on {\sc KITTI15} only, meaning our guided PatchMatch needs denser hints ({\sc KITTI15} has higher density among the three data sets). In fact, expanded guidance improves over all the test data sets, proving that our intuition is correct. %
On {\sc KITTI15} all error rate thresholds improve by roughly a factor of $4$, while MAE decreases from $3.47$ to $1.46$ and $1.44$ for \textit{Lin 3D exp} and \textit{3D $\mathcal{G}$ exp}, respectively. On {\sc ETH3D}, \textit{3D $\mathcal{G}$ exp} lowers MAE from $0.87$ to $0.54$ (38\% decrease). Also error rates are all improved marginally relative to \textit{Lin 3D exp}. Boost on {\sc Tartan} is also competitive with a $60\%$ lower MAE error. We could not observe a significant difference between \textit{Lin 3D exp} and \textit{3D $\mathcal{G}$ exp} densification, possibly due to the very sparse visual hints for this data set (see \cref{tab:exp_ablation}). Qualitative results are shown in \cref{fig:dp_qualitative}, which highlights that expanded hints are able to produce sharper predictions and remove inference artifacts (more examples in the supplement). 

\begin{table}[t]
    \centering
    \footnotesize
    \setlength{\tabcolsep}{6pt}
    \begin{tabularx}{\columnwidth}{l|c|c|c c c c}
    \toprule
          & \sc Expansion & MAE & ${>}2$ & ${>}3$ & ${>}4$ & ${>}5$ \\
    \midrule
         \multirow{4}{*}{ETH3D} & Vanilla $^\triangleleft$& 0.87 & 5.11 & 3.63 & 2.9 & 2.39 \\
         & \textit{Vgd} & 1.01 & 5.41 & 3.82 & 3.01 & 2.52 \\
         & \textit{Lin 3D exp} & 0.56 & 3.88 & 2.54 & 1.93 & 1.54 \\
         & \textit{3D $\mathcal{G}$ exp} & \textbf{0.54} & \textbf{3.76} & \textbf{2.43} & \textbf{1.85} & \textbf{1.47} \\
    \hline
         \multirow{4}{*}{\sc Tartan} &Vanilla $^\triangleleft$& 5.63 & 20.63 & 16.70 & 14.50 & 12.98 \\
         & \textit{Vgd} & 8.65 & 20.58 & 17.33 & 15.46 & 14.15 \\
         & \textit{Lin 3D exp} & 2.18 & 14.78 & 11.36 & 9.47 & 8.15 \\
         & \textit{3D $\mathcal{G}$ exp} & \textbf{2.17} & \textbf{14.77} & \textbf{11.34} & \textbf{9.45} & \textbf{8.13} \\
    \hline
         \multirow{4}{*}{KITTI15}  &Vanilla $^\triangleleft$& 3.47 & 34.76 & 23.59 & 17.76 & 14.10 \\
         & \textit{Vgd} & 1.94 & 10.98 & 6.37 & 4.68 & 3.79 \\
         & \textit{Lin 3D exp} & 1.46 & 9.46 & 5.27 & 3.78 & 3.00 \\
         & \textit{3D $\mathcal{G}$ exp} & \textbf{1.44} & \textbf{9.30} & \textbf{5.01} & \textbf{3.51} & \textbf{2.75} \\
    \bottomrule
    \end{tabularx}
    \caption{Ablation study of our guidance on DeepPruner. Expanded VIO guidance improves accuracy on unseen data. \textit{Vgd} is a sparse visual hints guided model, %
			\textit{3D $\mathcal{G}$ exp} is the 3D graph expansion (\cref{subsec:expansion}) and \textit{Lin 3D exp} the linear 3D expansion (\cref{subsec:fast_3d}). $\triangleleft$ Original DeepPruner-best {\sc Sceneflow} model.}
    \vspace{-12pt}
    \label{tab:exp_deeppruner}
\end{table}

\begin{table*}[t]
	\setlength{\tabcolsep}{9pt}
	\centering
		\begin{tabularx}{\textwidth}{l|ccccc|ccccc}
			\toprule\noalign{\smallskip}
			\multirow{2}{*}{\sc Model} & \multicolumn{5}{c|}{\sc ETH3D}& \multicolumn{5}{c}{\sc TartanAIR}\\ 
			
			& MAE & ${>}2$ & ${>}3$ &${>}4$ & ${>}5$ & MAE & ${>}2$ &${>}3$ &${>}4$ & ${>}5$ \\
			\noalign{\smallskip}
			\hline
			\noalign{\smallskip}
			\textit{PSMNet} \cite{chang2018pyramid}  & 5.25 & 16.99 &7.62 & 5.82& 5.10 & 5.51 & 21.30 &14.09&11.34& 9.81 \\ 
			\textit{PSMNet} \cite{chang2018pyramid} \textit{Vgd-test} & 5.25 & 16.78 &7.65 & 5.80& 5.10 & 5.51 & 21.29 &14.09 & 11.30& 9.80\\
			\textit{Vgd} & 1.54 & 8.59 &5.47&4.17& 3.41 & 5.53 & 21.03 &16.62&14.14& 12.45 \\
			\textit{Vgd} + \textit{Lin 3D exp} & 1.19 & 8.8 &5.16&3.63& 2.77 & 2.74 & 17.86 &13.29&10.81& 9.24 \\
			\textit{Vgd} + \textit{3D $\mathcal{G}$ exp}  & \textbf{1.04} & \textbf{7.91} &\textbf{4.41}&\textbf{3.01}& \textbf{2.27} & \textbf{2.54} & \textbf{17.00} &\textbf{12.72}&\textbf{10.48}& \textbf{9.04} \\
			\bottomrule
		\end{tabularx}
	\caption{Ablation study of our proposed methods on PSMNet \cite{chang2018pyramid}. \textit{Vgd} is a sparse visual hints guided model, %
			\textit{3D $\mathcal{G}$ exp} is the 3D graph expansion (\cref{subsec:expansion}) and \textit{Lin 3D exp} the linear 3D expansion (\cref{subsec:fast_3d}).}
	\vspace{-12pt}
	\label{tab:ablation_psmnet}
\end{table*}

\textbf{PSMNet} \label{par:psmnet} We test on {\sc Tartan}, {\sc ETH3D} and {\sc KITTI}. Results are illustrated in \cref{tab:ablation_psmnet}. Initial results confirm the findings of \cite{Poggi_CVPR_2019}, there is a minor improvement on both {\sc ETH3D} and {\sc Tartan} when using guidance \textit{Vgd-test} at test time, and a clear improvement across all metrics with \textit{Vgd} at training time as well. This is a clue that our sparse visual hints can positively contribute to improve generalization performance. %
For {\sc ETH3D}, \textit{Vgd} MAE is $1.54$, with \textit{Vgd + Lin 3D exp} MAE lowers to $1.19$ ($23\%$ lower), even better with \textit{Vgd + 3D $\mathcal{G}$ exp} down to $1.04$ ($30\%$ lower). %
Qualitative results on challenging {\sc ETH3D} scenes in \cref{fig:psmnet_qualitative} confirm the benefits of using the proposed guidance for sharper and more accurate disparity predictions. More examples are available in the supplementary. 
{\sc Tartan} is peculiar to have very sparse visual hints, which makes it a very challenging test-bed for our method. This is highlighted by the minor metric changes for \textit{Vgd-test}. \textit{Vgd} obtains a major improvement of $38\%$ in MAE, increasing to $50\%$ and $54\%$ for  \textit{Vgd + Lin 3D exp} and \textit{Vgd + 3D $\mathcal{G}$ exp}, respectively. %

{\sc KITTI} We test on the $2011\_09\_26\_0011$ {\sc KITTI Velodyne} sequence. \cref{tab:kitti_velodyne} presents on the upper half reference results of PSMNet and finetuned PSMNet. The bottom half is split in LiDaR guided ({\sc Lgd}) and VIO guided ({\sc Vgd}) stereo matching. LiDaR expansion improves over \cite{Poggi_CVPR_2019} for MAE error while paying a minor loss in ${<} 2\%$. This is a notable achievement as the performance of \cite{Poggi_CVPR_2019} are strong and shows expansion can benefit LiDaR guided stereo matching. Notably, VIO expansion as well achieves a slight improvement in MAE error. To summarize, hints guide cost volume construction, leading to better overall performace (MAE) but expansion in this experiment is not yet able to improve accuracy (${<} 2\%$) likely due to expansion noise.

In addition to  {\sc KITTI Velodyne}, we evaluate the performance on a model pre-trained on {\sc Sceneflow} without any guidance, and test it on {\sc KITTI15}. %
We report numbers in \cref{tab:expansion_pretr_model}. Starting from the publicly available {\sc SceneFlow} pre-trained model from PSMNet authors, we obtain MAE $4.24$. With \textit{PSMNet Vgd-test} guidance, the performance does not change except for a minor improvement on the error rates. Hints expansion is more effective, \textit{Lin 3D exp} leads to a MAE improvement of $0.5\%$ and a decrease on the ${>}2$ error rate, meaning that the model is more precise in some areas with small errors. \textit{3D $\mathcal{G}$ exp} is even more effective with a $2.5\%$ MAE decrease but shows mixed results on the errors rates. A possible reason, as happened in  {\sc KITTI Velodyne}, is that guided areas improve accuracy (thus lower MAE) but unguided areas do not. We speculate the reason is the feature modulation in the guided cost volume.  %
Finally, we include GSM \cite{Poggi_CVPR_2019} as a sensor-based reference which gives more competitive results. Again our expansions improve the {\sc Lgd} baseline by around $3\%$ MAE and similarly all the error rates. %
Yet, we emphasize that our starting hint density is much lower ($0.1\%$ vs.\ their $5\%$). %
To summarize, in \cref{tab:expansion_pretr_model} performance gain is limited because the model was not trained to exploit guidance. However, guidance positively contributes out-of-the-box to improve accuracy.

\begin{table}[t]
    \centering
    \setlength{\tabcolsep}{2pt}
    \begin{tabularx}{\columnwidth}{c|cc|cc|cc|cc}
    \toprule
         & \multicolumn{2}{c|}{${<} 2\%$} & \multicolumn{2}{c|}{MAE} &\multicolumn{2}{c|}{${<} 2\%$} & \multicolumn{2}{c}{MAE}\\
         & All & NoG & All & NoG & All & NoG & All & NoG \\
         \midrule
         PSMNet \cite{chang2018pyramid} & 38.60 & 38.86 & 2.36 & 2.37 &--&--&--&-- \\
         PSMNet-ft & 1.71 & 1.73 & 0.72 & 0.73 &--&--&--&-- \\
         \midrule
         & \multicolumn{4}{c|}{LiDaR guidance (\sc Lgd)} & \multicolumn{4}{c}{VIO guidance (\sc Vgd)} \\
         \midrule
         {\sc gd} & \textbf{0.67} & \textbf{0.67} & 0.47 & 0.47 & \textbf{1.71} & \textbf{1.73} & 0.72 & 0.73 \\
         \textit{Lin 3D exp} & 0.79 & 0.77 & 0.44 & 0.45 & 2.08 & 2.08 & 0.72 & 0.72\\
         \textit{3D $\mathcal{G}$ exp} & 0.73 & 0.82 & \textbf{0.41} & \textbf{0.44} & 1.90 & 1.91& \textbf{0.70} & \textbf{0.71}\\
         \bottomrule
    \end{tabularx}
    \caption{Ablation study of our expansions on {\sc KITTI Velodyne}. Top half, reference PSMNet results. Bottom half, expanded LiDaR and VIO guidance reduce MAE error for guided stereo matching, proving expansion is effective not only on VIO but also on LiDaR. %
    }
    \vspace{-9pt}
    \label{tab:kitti_velodyne}
\end{table}

\begin{table}[t]
    \centering\footnotesize
    \setlength{\tabcolsep}{2pt}
    \begin{tabularx}{\columnwidth}{l|c|c|c |c c c c}
    \toprule
         \sc \sc Guidance & \sc Vgd & \sc Lgd & MAE & ${>}2$ & ${>}3$ & ${>}4$ & ${>}5$ \\
    \midrule
         \textit{PSMNet}\cite{chang2018pyramid}$^{\triangleleft,\oplus}$ &&& 4.24 & 46.54 & 29.61 & 21.26 & 16.41 \\
     \midrule
         \textit{GSM Lgd-test}\cite{Poggi_CVPR_2019}&&\checkmark& 3.90 & 33.38 & 23.12 & 17.59 & 14.01 \\
         \textit{GSM Lgd-test} \textit{Lin 3D exp}$^{\triangleleft}$&&\checkmark& 3.82 & 32.83 & 22.45 & 17.09 & 13.19 \\
         \textit{GSM Lgd-test} \textit{3D $\mathcal{G}$ exp}$^{\triangleleft}$&&\checkmark&  \textbf{3.79} & \textbf{32.34} & \textbf{22.08} & \textbf{17.01} & \textbf{13.03} \\
      \midrule
         \textit{PSMNet Vgd-test}$^{\triangleleft}$ &\checkmark&& 4.24 & 46.45 & \textbf{29.57} & \textbf{21.24} & \textbf{16.40} \\
         \textit{Lin 3D exp}$^{\triangleleft}$ &\checkmark&& 4.21 & \textbf{46.28} & 29.62 & 21.29 & 16.44 \\
         \textit{3D $\mathcal{G}$ exp}$^{\triangleleft}$ &\checkmark&& \textbf{4.13} & 47.13 & 30.65 & 22.04 & 16.98 \\
    \bottomrule
    \end{tabularx}
    \caption{Pre-trained PSMNet agnostic to guidance, tested on {\sc KITTI15} with hints guidance. %
    The model was not trained to exploit guidance. However, results confirm expansion is effective on both LiDaR (\textit{Lgd}) and VIO (\textit{Vgd}) hints guidance out-of-the-box. $\triangleleft$ authors original checkpoint and $\oplus$ code.}
    \vspace{-12pt}
    \label{tab:expansion_pretr_model}
\end{table}

\subsection{Comparison with Guided Stereo Methods}
\label{subsec:sota}

\begin{table}[t]
    \centering\footnotesize
    \setlength{\tabcolsep}{6pt}
    \begin{tabularx}{1\columnwidth}{l|c|c|c c c c}
    \toprule
         \sc Model & \sc Lgd & MAE & ${>}2$ & ${>}3$ & ${>}4$ & ${>}5$ \\
    \midrule
         \textit{PSMNet}\cite{chang2018pyramid} && 4.24 & 46.54 & 29.61 & 21.26 & 16.41 \\
         \textit{GSM}\cite{Poggi_CVPR_2019}&\checkmark& \textbf{1.39} & \textbf{12.31} & \textbf{3.89} & \textbf{2.23} & \textbf{1.60} \\
    \hline
         \textit{Lin 3D exp} && 3.44 & 41.63 & 29.23 & 22.19 & 17.66 \\
         \textit{3D $\mathcal{G}$ exp}& & \textit{3.21} & \textit{38.27} & \textit{26.78} & \textit{20.35} & \textit{16.23} \\
    \toprule
         \textit{GSM-ft}\cite{Poggi_CVPR_2019} &\checkmark& 0.763 & 2.73 & 1.82 & 1.51 & 1.33 \\
         \textit{$S^3$-ft}\cite{Huang_2021_CVPR} &\checkmark& \textbf{0.443} & \textbf{1.65} & \textbf{0.96} & \textbf{0.71} & \textbf{0.57} \\
    \hline
         \textit{Lin 3D exp-ft} && \textit{0.95} & \textit{6.27} & 3.29 & 2.35 & 1.87 \\
         \textit{3D $\mathcal{G}$ exp-ft} && 0.98 & 6.28 & \textit{3.25} & \textit{2.35} & \textit{1.89} \\
    \bottomrule
    \end{tabularx}
    \caption{Comparison with guided stereo methods. Training on Sceneflow and testing on KITTI15 (upper half), and finetuned on KITTI12 (lower half).}
    \vspace{-6pt}
    \label{tab:sota_kitti}
\end{table}

We compare with state-of-the-art guided stereo matching methods \cite{Poggi_CVPR_2019,Huang_2021_CVPR}. In the upper half of \cref{tab:sota_kitti}, models are trained on {\sc SceneFlow} and tested on {\sc KITTI15} (as in \cref{tab:expansion_pretr_model}). Our improvement over PSMNet is particularly evident in average error and for pixels with small error rate (threshold ${>}2$). The \textit{Lin 3D exp} model suffers a minor accuracy drop on the higher error rates (thresholds ${>}4, {>}5$) while improving $18\%$ on MAE. \textit{3D $\mathcal{G}$ exp} gains on all the metrics, and \textit{GSM} \cite{Poggi_CVPR_2019} obtains the best absolute performance. It is worth noting they exploit LiDaR guidance, which as already discussed is evenly distributed and accurate, leading to a clear performance advantage. 
If additional sensors are available, a LiDaR can be effective. Models finetuned (suffix `-ft') on {\sc KITTI12} are in the bottom half of \cref{tab:sota_kitti}. All models obtain clear gains thanks to reduced domain-shift. However, in this case \textit{$S^3$-ft}\cite{Huang_2021_CVPR} improved upon \textit{GSM-ft} \cite{Poggi_CVPR_2019} to obtain state-of-the-art results. Finetuning reduces the gap of our methods with \textit{GSM-ft}. Before finetuning the average error of \textit{GSM} is $57\%$ lower and after it is $23\%$ lower. %

\subsection{Embedded Devices}
\label{subsec:embedded}

\begin{figure}
    \centering
    \includegraphics[width=\columnwidth]{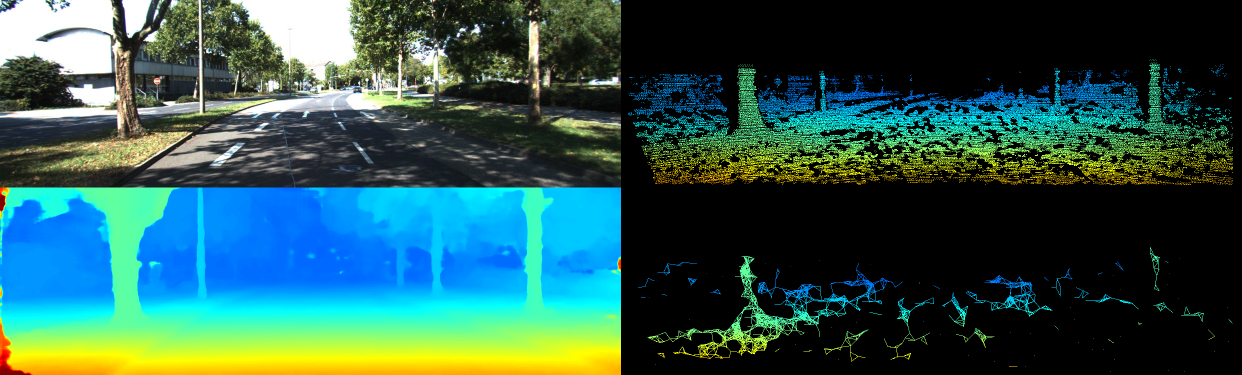}
    \caption{Jetson board results on {\sc KITTI} running at ${\sim}1\,$fps, $8\times$ faster than the standard model, with further engineering it could work real-time. RGB (top left), LiDaR GT (top right), predicted disparity (bottom left) and \textit{3D $\mathcal{G}$ exp} visual hints (bottom right).}
    \label{fig:jetson}
    \vspace{-12pt}
\end{figure}
To demonstrate our method on an embedded device, we perform inference on an NVIDIA Jetson AGX Xavier device with {\sc KITTI Velodyne} sequence \\$2011\_09\_26\_0011$ at high resolution ($384 {\times} 1280$ pixels) using PSMNet\footnote{We were unable to deploy DeepPruner on the device yet, as some of the operations in the model could not be optimized/converted by TRTorch}. Our \textit{3D $\mathcal{G}$ exp} has an inference time of $8.607$ seconds per sample, while the MAE error on the full sequence is $0.70$. Furthermore, we optimized the model to use half-precision (fp16) and target the aarch64 platform using NVIDIA TensorRT inference optimizer via TRTorch compiler. This resulted in an inference time of $1.062$ seconds per sample (a speed-up factor of $8\times$ relative to the unoptimized model),  while maintaining the original accuracy MAE $ = 0.70$. \cref{fig:jetson} illustrates a qualitative example of inference. An interesting observation is that the densification of visual hints on vertical structures like trees works particularly well. The predicted disparity is accurate except a typical stereo artifact on the right.

We profiled our code (after applying NVIDIA TensorRT optimization \cref{tab:jetson}) and the execution time is divided as follows: data loading and VIO hints extraction $12\%$, \textit{3D $\mathcal{G}$} expansion $10\%$, and deep CNN inference $78\%$. Most of the cost comes from the deep CNN employed. Expansion on Jetson comes at a time cost similar to data loading, but improves MAE from $2.36$ to $0.70$ as reported in \cref{tab:kitti_velodyne}. In the case of \textit{Lin 3D} expansion, the execution time is slightly faster because the expansion is more lightweight. Resulting in an inference time of $0.9708$ seconds per sample and $MAE = 0.72$. The detailed execution time breakdown is $13\%$ for data loading, $1.5\%$ for \textit{Lin 3D} expansion, and $85.5\%$ for deep CNN inference. To summarize, \textit{Lin 3D} is $~8.5\%$ faster, while \textit{3D $\mathcal{G}$} is $~4\%$ more accurate. %

\begin{table}[t]
    \centering\footnotesize
    \setlength{\tabcolsep}{4pt}
    \begin{tabularx}{1\columnwidth}{l|c|c|c c c }
    \toprule
         \sc Guidance & \sc Time (sec/sample) & MAE & DL+VIO & \sc exp & CNN   \\
    \midrule
         \textit{Lin 3D exp} & \textbf{0.9708} & 0.72 & 13\% & 1.5\% & 88.5\% \\
         \textit{3D $\mathcal{G}$ exp} & 1.062 & \textbf{0.70} & 12\% & 10\% & 85.5\%  \\
    \bottomrule
    \end{tabularx}
    \caption{Inference on an NVIDIA Jetson AGX Xavier and {\sc KITTI Velodyne} ($384 {\times} 1280$ pixels). Breakdown of time (in \%) employed for Data Loading and VIO extraction (DL+VIO), expansion ({\sc exp}) and inference in CNN (CNN).}
    \vspace{-12pt}
    \label{tab:jetson}
\end{table}

\section{Conclusion}

This work tackled a challenging scenario for stereo-matching methods. Without any additional sensors, improving their performance on unseen sequences (i.e. from different data distributions). To this end, we demonstrated the utility of VIO hints guidance for deep stereo matching. In particular, 3D visual hints expansion seamlessly works for existing pre-trained models and guidance-aware models, and across different architectures. Our technique does not need additional sensors and exploits well studied and robust odometry techniques. Nevertheless, we show that also LiDaR benefits from our expansion. Extensive experiments on distinct and challenging data sets support our findings, and practical use was also investigated by successfully deploying our algorithm on embedded devices.

\paragraph{Acknowledgements.} We acknowledge funding from Academy of Finland (No.\ 339730, 324345), and the Finnish Center for Artificial Intelligence (FCAI). 
We acknowledge the computational resources by the Aalto Science-IT project and CSC -- IT Center for Science, Finland. \sloppy

{\small
\bibliographystyle{ieee_fullname}

\begin{thebibliography}{10}\itemsep=-1pt

\bibitem{bai2019novel}
Jinqiang Bai, Junqiang Gao, Yimin Lin, Zhaoxiang Liu, Shiguo Lian, and Dijun
  Liu.
\newblock A novel feedback mechanism-based stereo visual-inertial slam.
\newblock {\em IEEE Access}, 7:147721--147731, 2019.

\bibitem{bleyer2011patchmatch}
Michael Bleyer, Christoph Rhemann, and Carsten Rother.
\newblock Patchmatch stereo-stereo matching with slanted support windows.
\newblock In {\em Proceedings of the British Machine Vision Conference (BMVC)},
  volume~11, pages 1--11, 2011.

\bibitem{bloesch2015robust}
Michael Bloesch, Sammy Omari, Marco Hutter, and Roland Siegwart.
\newblock Robust visual inertial odometry using a direct ekf-based approach.
\newblock In {\em Proceedings of the IEEE/RSJ International Conference on
  Intelligent Robots and Systems (IROS)}, pages 298--304. IEEE, 2015.

\bibitem{jisct1993stereo}
Robert Bolles, Harlyn Baker, and M Hannah.
\newblock The jisct stereo evaluation.
\newblock In {\em DARPA Image Understanding Worshop}, 1993.

\bibitem{boykov2001fast}
Yuri Boykov, Olga Veksler, and Ramin Zabih.
\newblock Fast approximate energy minimization via graph cuts.
\newblock {\em IEEE Transactions on Pattern Analysis and Machine Intelligence
  (T-PAMI)}, 23(11):1222--1239, 2001.

\bibitem{chang2018pyramid}
Jia-Ren Chang and Yong-Sheng Chen.
\newblock Pyramid stereo matching network.
\newblock In {\em Proceedings of the IEEE/CVF Conference on Computer Vision and
  Pattern Recognition (CVPR)}, pages 5410--5418, 2018.

\bibitem{Chen_2019_ICCV}
Chuangrong Chen, Xiaozhi Chen, and Hui Cheng.
\newblock On the over-smoothing problem of cnn based disparity estimation.
\newblock In {\em Proceedings of the IEEE/CVF International Conference on
  Computer Vision (ICCV)}, 2019.

\bibitem{CLARK1990165}
Brent~N Clark, Charles~J Colbourn, and David~S Johnson.
\newblock Unit disk graphs.
\newblock {\em Discrete Mathematics}, 86(1-3):165--177, 1990.

\bibitem{geom_graph}
Jesper Dall and Michael Christensen.
\newblock Random geometric graphs.
\newblock {\em Physical Review E}, 66(1):016121, 2002.

\bibitem{Duggal2019ICCV}
Shivam Duggal, Shenlong Wang, Wei-Chiu Ma, Rui Hu, and Raquel Urtasun.
\newblock Deeppruner: Learning efficient stereo matching via differentiable
  patchmatch.
\newblock In {\em Proceedings of the IEEE/CVF International Conference on
  Computer Vision (ICCV)}, pages 4384--4393, 2019.

\bibitem{eigen2015predicting}
David Eigen and Rob Fergus.
\newblock Predicting depth, surface normals and semantic labels with a common
  multi-scale convolutional architecture.
\newblock In {\em Proceedings of the IEEE/CVF International Conference on
  Computer Vision (ICCV)}, pages 2650--2658, 2015.

\bibitem{garg2016unsupervised}
Ravi Garg, Vijay~Kumar Bg, Gustavo Carneiro, and Ian Reid.
\newblock Unsupervised cnn for single view depth estimation: Geometry to the
  rescue.
\newblock In {\em Proceedings of the European Conference on Computer Vision
  (ECCV)}, pages 740--756. Springer, 2016.

\bibitem{godard2017unsupervised}
Cl{\'e}ment Godard, Oisin Mac~Aodha, and Gabriel~J Brostow.
\newblock Unsupervised monocular depth estimation with left-right consistency.
\newblock In {\em Proceedings of the IEEE/CVF Conference on Computer Vision and
  Pattern Recognition (CVPR)}, pages 270--279, 2017.

\bibitem{Guo_2019_CVPR}
Xiaoyang Guo, Kai Yang, Wukui Yang, Xiaogang Wang, and Hongsheng Li.
\newblock Group-wise correlation stereo network.
\newblock In {\em Proceedings of the IEEE/CVF Conference on Computer Vision and
  Pattern Recognition (CVPR)}, pages 3273--3282, 2019.

\bibitem{Hannah1974ComputerMO}
Marsha~Jo Hannah.
\newblock Computer matching of areas in stereo images.
\newblock In {\em PhD Thesis, Stanford University}, 1974.

\bibitem{Huang_2021_CVPR}
Yu-Kai Huang, Yueh-Cheng Liu, Tsung-Han Wu, Hung-Ting Su, Yu-Cheng Chang,
  Tsung-Lin Tsou, Yu-An Wang, and Winston~H Hsu.
\newblock S3: Learnable sparse signal superdensity for guided depth estimation.
\newblock In {\em Proceedings of the IEEE/CVF Conference on Computer Vision and
  Pattern Recognition (CVPR)}, pages 16706--16716, 2021.

\bibitem{kendall2017end}
Alex Kendall, Hayk Martirosyan, Saumitro Dasgupta, Peter Henry, Ryan Kennedy,
  Abraham Bachrach, and Adam Bry.
\newblock End-to-end learning of geometry and context for deep stereo
  regression.
\newblock In {\em Proceedings of the IEEE/CVF International Conference on
  Computer Vision (ICCV)}, pages 66--75, 2017.

\bibitem{khamis2018stereonet}
Sameh Khamis, Sean Fanello, Christoph Rhemann, Adarsh Kowdle, Julien Valentin,
  and Shahram Izadi.
\newblock Stereonet: Guided hierarchical refinement for real-time edge-aware
  depth prediction.
\newblock In {\em Proceedings of the European Conference on Computer Vision
  (ECCV)}, pages 573--590, 2018.

\bibitem{liang2018learning}
Zhengfa Liang, Yiliu Feng, Yulan Guo, Hengzhu Liu, Wei Chen, Linbo Qiao, Li
  Zhou, and Jianfeng Zhang.
\newblock Learning for disparity estimation through feature constancy.
\newblock In {\em Proceedings of the IEEE Conference on Computer Vision and
  Pattern Recognition (CVPR)}, pages 2811--2820, 2018.

\bibitem{marroquin1987probabilistic}
Jose Marroquin, Sanjoy Mitter, and Tomaso Poggio.
\newblock Probabilistic solution of ill-posed problems in computational vision.
\newblock {\em Journal of the American Statistical Association},
  82(397):76--89, 1987.

\bibitem{sceneflow}
Nikolaus Mayer, Eddy Ilg, Philip Hausser, Philipp Fischer, Daniel Cremers,
  Alexey Dosovitskiy, and Thomas Brox.
\newblock A large dataset to train convolutional networks for disparity,
  optical flow, and scene flow estimation.
\newblock In {\em Proceedings of the IEEE/CVF Conference on Computer Vision and
  Pattern Recognition (CVPR)}, pages 4040--4048, 2016.

\bibitem{Menze2015ISA}
Moritz Menze, Christian Heipke, and Andreas Geiger.
\newblock Joint 3d estimation of vehicles and scene flow.
\newblock {\em ISPRS Annals of the Photogrammetry, Remote Sensing and Spatial
  Information Sciences}, 2:427, 2015.

\bibitem{Menze2018JPRS}
Moritz Menze, Christian Heipke, and Andreas Geiger.
\newblock Object scene flow.
\newblock {\em ISPRS Journal of Photogrammetry and Remote Sensing}, 140:60--76,
  2018.

\bibitem{murORB2}
Raul Mur-Artal and Juan~D Tard{\'o}s.
\newblock Orb-slam2: An open-source slam system for monocular, stereo, and
  rgb-d cameras.
\newblock {\em IEEE Transactions on Robotics}, 33(5):1255--1262, 2017.

\bibitem{pami2020pilzer}
Andrea Pilzer, St{\'e}phane Lathuili{\`e}re, Dan Xu, Mihai~Marian Puscas, Elisa
  Ricci, and Nicu Sebe.
\newblock Progressive fusion for unsupervised binocular depth estimation using
  cycled networks.
\newblock {\em IEEE Transactions on Pattern Analysis and Machine Intelligence
  (T-PAMI)}, 42(10):2380--2395, 2019.

\bibitem{Poggi_CVPR_2019}
Matteo Poggi, Davide Pallotti, Fabio Tosi, and Stefano Mattoccia.
\newblock Guided stereo matching.
\newblock In {\em Proceedings of the IEEE/CVF Conference on Computer Vision and
  Pattern Recognition (CVPR)}, pages 979--988, 2019.

\bibitem{rosinol2020kimera}
Antoni Rosinol, Marcus Abate, Yun Chang, and Luca Carlone.
\newblock Kimera: an open-source library for real-time metric-semantic
  localization and mapping.
\newblock In {\em 2020 IEEE International Conference on Robotics and Automation
  (ICRA)}, pages 1689--1696. IEEE, 2020.

\bibitem{middlebury}
Daniel Scharstein, Heiko Hirschm{\"u}ller, York Kitajima, Greg Krathwohl, Nera
  Ne{\v{s}}i{\'c}, Xi Wang, and Porter Westling.
\newblock High-resolution stereo datasets with subpixel-accurate ground truth.
\newblock In {\em Proceedings of the German Conference on Pattern Recognition
  (GCPR)}, pages 31--42. Springer, 2014.

\bibitem{scharstein1998stereo}
Daniel Scharstein and Richard Szeliski.
\newblock Stereo matching with nonlinear diffusion.
\newblock {\em International Journal of Computer Vision (IJCV)},
  28(2):155--174, 1998.

\bibitem{schoenberger2016sfm}
Johannes~L Schonberger and Jan-Michael Frahm.
\newblock Structure-from-motion revisited.
\newblock In {\em Proceedings of the IEEE/CVF Conference on Computer Vision and
  Pattern Recognition (CVPR)}, pages 4104--4113, 2016.

\bibitem{schoenberger2016mvs}
Johannes~L Sch{\"o}nberger, Enliang Zheng, Jan-Michael Frahm, and Marc
  Pollefeys.
\newblock Pixelwise view selection for unstructured multi-view stereo.
\newblock In {\em Proceedings of the European Conference on Computer Vision
  (ECCV)}, pages 501--518. Springer, 2016.

\bibitem{Schops_2019_CVPR}
Thomas Schops, Torsten Sattler, and Marc Pollefeys.
\newblock Bad slam: Bundle adjusted direct rgb-d slam.
\newblock In {\em Proceedings of the IEEE/CVF Conference on Computer Vision and
  Pattern Recognition (CVPR)}, pages 134--144, 2019.

\bibitem{schoeps2017cvpr}
Thomas Schops, Johannes~L Schonberger, Silvano Galliani, Torsten Sattler,
  Konrad Schindler, Marc Pollefeys, and Andreas Geiger.
\newblock A multi-view stereo benchmark with high-resolution images and
  multi-camera videos.
\newblock In {\em Proceedings of the IEEE/CVF Conference on Computer Vision and
  Pattern Recognition (CVPR)}, pages 3260--3269, 2017.

\bibitem{seiskari_2021_arxiv}
Otto Seiskari, Pekka Rantalankila, Juho Kannala, Jerry Ylilammi, Esa Rahtu, and
  Arno Solin.
\newblock Hybvio: Pushing the limits of real-time visual-inertial odometry.
\newblock {\em ArXiv Preprint}, 2021.

\bibitem{sinha2020deltas}
Ayan Sinha, Zak Murez, James Bartolozzi, Vijay Badrinarayanan, and Andrew
  Rabinovich.
\newblock Deltas: Depth estimation by learning triangulation and densification
  of sparse points.
\newblock In {\em Proceedings of the European Conference on Computer Vision
  (ECCV)}, pages 104--121. Springer, 2020.

\bibitem{sun2018robust}
Ke Sun, Kartik Mohta, Bernd Pfrommer, Michael Watterson, Sikang Liu, Yash
  Mulgaonkar, Camillo~J Taylor, and Vijay Kumar.
\newblock Robust stereo visual inertial odometry for fast autonomous flight.
\newblock {\em IEEE Robotics and Automation Letters (RAL)}, 3(2):965--972,
  2018.

\bibitem{Tankovich_2021_CVPR}
Vladimir Tankovich, Christian Hane, Yinda Zhang, Adarsh Kowdle, Sean Fanello,
  and Sofien Bouaziz.
\newblock Hitnet: Hierarchical iterative tile refinement network for real-time
  stereo matching.
\newblock In {\em Proceedings of the IEEE/CVF Conference on Computer Vision and
  Pattern Recognition (CVPR)}, pages 14362--14372, 2021.

\bibitem{usenko2019visual}
Vladyslav Usenko, Nikolaus Demmel, David Schubert, J{\"o}rg St{\"u}ckler, and
  Daniel Cremers.
\newblock Visual-inertial mapping with non-linear factor recovery.
\newblock {\em IEEE Robotics and Automation Letters (RAL)}, 5(2):422--429,
  2019.

\bibitem{tartanair2020iros}
Wenshan Wang, Delong Zhu, Xiangwei Wang, Yaoyu Hu, Yuheng Qiu, Chen Wang, Yafei
  Hu, Ashish Kapoor, and Sebastian Scherer.
\newblock Tartanair: A dataset to push the limits of visual slam.
\newblock In {\em Proceedings of the IEEE/RSJ International Conference on
  Intelligent Robots and Systems (IROS)}, pages 4909--4916. IEEE, 2020.

\bibitem{wong2020unsupervised}
Alex Wong, Xiaohan Fei, Stephanie Tsuei, and Stefano Soatto.
\newblock Unsupervised depth completion from visual inertial odometry.
\newblock {\em IEEE Robotics and Automation Letters (RAL)}, 5(2):1899--1906,
  2020.

\bibitem{you2020pseudo}
Yurong You, Yan Wang, Wei-Lun Chao, Divyansh Garg, Geoff Pleiss, Bharath
  Hariharan, Mark Campbell, and Kilian~Q Weinberger.
\newblock Pseudo-lidar++: Accurate depth for 3d object detection in autonomous
  driving.
\newblock In {\em Proceedings of the International Conference on Learning
  Representations (ICLR)}, 2020.

\bibitem{10.1007/BFb0028345}
Ramin Zabih and John Woodfill.
\newblock Non-parametric local transforms for computing visual correspondence.
\newblock In {\em Proceedings of the European Conference on Computer Vision
  (ECCV)}, pages 151--158. Springer, 1994.

\end{thebibliography}

}

\clearpage
\appendix


\appendix
\twocolumn[\vspace*{5em}\centering\Large\bf%
{\Large Supplementary Material for} \\ {Expansion of Visual Hints for Improved Generalization in Stereo Matching}%
\vspace*{4em}]

\setcounter{table}{0}
\renewcommand{\thetable}{A\arabic{table}}%
\setcounter{figure}{0}
\renewcommand{\thefigure}{A\arabic{figure}}%
\setcounter{equation}{0}
\renewcommand{\theequation}{A\arabic{equation}}%

\section{Limitations}

We proposed visual hints as a mean to improve deep stereo matching performance on novel unseen data. 
VIO keypoint triangulation works best when there is sufficient camera motion that allows to extend the baseline for accurate triangulation, and, hence, in the initial $5$ to $10$ frames of a sequence a warm-up time may be beneficial to reach a higher number of accurate hints. 
Nowadays, real-time VIO algorithms are very effective and run up to $30fps$. Thus, our method is widely applicable on a wide variety of applications. So for practical use cases one should choose the VIO algorithm according to the application requirements. This is due to the fact that a generic VIO algorithm produces keypoints at a lower rate than the video frame rate.

There is scope for further improvement of the current 3D densification by leveraging the power of convex hulls and possibly integrating stereo and odometry pipelines. Finally, further steps towards real-time execution, extension to traditional stereo methods, extension to multi-view stereo and monocular depth estimation could be undertaken.

\section{Data Set Details}

We use SceneFlow \cite{sceneflow}, ETH3D \cite{schoeps2017cvpr}, \cite{Schops_2019_CVPR}, TartanAir \cite{tartanair2020iros}, KITTI Stereo 2012/2015 \cite{Menze2015ISA}, \cite{Menze2018JPRS} data sets for our experiments.

\textbf{SceneFlow} is a large synthetic stereo data set with ground truth used for model pretraining. It contains diverse scenes with a total of 35454 training images. For training we follow the guidelines in PSMNet and DeepPruner (\eg the image is randomly cropped to the size of $256 \times 512$ and data augmentation is applied).

\textbf{ETH3D} is a real-world stereo and SLAM benchmarking data set recorded in varied indoor and outdoor environments. We adopt the \textit{Low-res many-view} training split for testing our method. In total we have 1199 test images. For testing we use the largest possible center crop such that it works with PSMNet (size $480 \times 912$) and DeepPruner (size $480 \times 896$).

\textbf{TartanAir} is a real-world stereo data set for SLAM benchmarking, recorded with a flying drone in different environments. We evaluate our method on 18 stereo image sequences, one for each environment, totalling 3224 images. The sequences we used are \textit{abandonedfactory-easy-P010}, \textit{endofworld-easy-P000}, \textit{neighborhood-easy-P000}, \textit{oldtown-easy-P000}, \textit{soulcity-easy-P006}, \textit{abandonedfactory-night-easy-P013}, \textit{gascola-easy-P004}, \textit{ocean-easy-P000}, \textit{seasidetown-easy-P006}, \textit{westerndesert-easy-P004}, \textit{amusement-easy-P006}, \textit{hospital-easy-P018}, \textit{office2-easy-P009}, \textit{seasonsforest-easy-P003}, \textit{carwelding-easy-P002}, \textit{japanesealley-easy-P001}, \textit{office-easy-P003}, and \textit{seasonsforest-winter-easy-P000}. For this dataset, center cropping is not required as the original image size ($480 \times 640$) fits in both deep stereo matching architectures used in this study.

\textbf{KITTI15, 12} are two real-world stereo benchmarking data sets acquired from a car driving in German streets. {\sc KITTI15} is used only for testing and we finetune some of our models on {\sc KITTI12} to compare with state-of-the-art. For training with {\sc KITTI12} we follow author's guidelines with standard data augmentation. For testing, only center cropping to $368 \times 1232$ and $320 \times 1216$ is applied for PSMNet and DeepPruner, respectively.

\textbf{Sparse Hints} We used 
visual reconstruction based on COLMAP \cite{schoenberger2016mvs}, \cite{schoenberger2016sfm} to get the sparse hints for our data sets. All data sets we used are composed of sequences, thus we only had to provide the sequence as an input to the algorithm.

\section{3D Linear Graph Expansion}

In this section we report the algorithm and a Python implementation \cref{code:linear} of our \textit{Lin 3D exp} presented in \cref{subsec:fast_3d}. To densify the hints, first the patch iterator (\textit{patch\_iter}) function is called to split the image in patches. This is followed by a call to \textit{dense\_patch}, which performs the densification of the patches.

\begin{algorithm}[t]
\small
\SetAlgoLined\DontPrintSemicolon
\newcommand\mycommfont[1]{\hfill\textcolor{gray}{#1}}
\SetCommentSty{mycommfont}
\SetKwComment{Comment}{}{}%
\SetKwComment{tcp}{}{}%
\KwData{Hint matrix $H$ data set, patch size $W$}

 \For{ $\mathbf{h}$ in $H$}{
  Init expanded hints $\mathbf{h}_{\exp} = \mathbf{h}$\; 
  Create patches of size $W$: $p \leftarrow \mathbf{h}_{\exp}$\;
  \For{$p_i$ in patches $p$}{
    \For(\tcp*[f]{search horizontally}){$x = \{1,\dots,W\}$}{
        \If(\tcp*[f]{on x axis}){two hints not zero}{
            $p_i[x] \leftarrow$  linearly interpolate hints\;
        }
    }
    \For(\tcp*[f]{search vertically}){$y = \{1,\dots,W\}$}{
        \If(\tcp*[f]{on y axis}){two hints not zero}{
            $p_i[y] \leftarrow$  linearly interpolate hints\;
        }
    }
    $\mathbf{h}_{\exp} \leftarrow p_i$\;
    }
    $\mathbf{h} \leftarrow \mathbf{h}_{\exp}$\;
 }
 \caption{3D Linear Hints Expansion}
 \label{algo:lin_expansion}
\end{algorithm}

\begin{lstlisting}[language=Python, caption=Code for \textit{Lin 3D exp}., label={code:linear}]
import numpy as np

def dense_patch(patch):
"""
Function to expand sparse patches of hints with 3D Linear Interpolation. 
Input:
    patch: sparse patch of hints, hints are disparity values > 0, otherwise matrix values are 0
Ourput:
    new_patch: densified input hints
"""
    patch_size = patch.shape[0]
    new_patch = patch.copy()

    if np.count_nonzero(patch) >=3 :
        for u in range(patch_size):
            line = patch[u]
            if np.count_nonzero(line) >= 2:
                xp = [i for i in range(patch_size) if patch[u, i] > 0]
                fp = [patch[u, i] for i in xp]
                tmp = np.interp(range(patch_size) , xp, fp)
                new_patch[u] = tmp


        for v in range(patch_size):
            line = new_patch[:, v]
            if np.count_nonzero(line) >= 2:
                xp = [i for i in range(patch_size) if new_patch[i, v] > 0]
                fp = [new_patch[i, v] for i in xp]
                tmp = np.interp(range(patch_size) , xp, fp)
                new_patch[:, v] = tmp
    return new_patch

def patch_iter(hints, win):
"""
Function to expand sparse matrix of hints with 3D Linear Expansion. Creates patches and calls dense_patch to expand hints.
Input:
    hints: sparse matrix of hints, hints are disparity values > 0, otherwise matrix values are 0
    win: window size W to create patches.
Ourput:
    new: densified input hints
"""
    h, w = hints.shape
    new = hints.copy()
    for i in range(0, h//win):
        for j in range(0, w//win):
            window = hints[win*i:win*(i+1), win*j:win*(j+1)]
            new[win*i:win*(i+1), win*j:win*(j+1)] = dense_patch(dense_patch(window))
    return new

\end{lstlisting}

\section{3D Graph Expansion}

In this section, we report the algorithm and a Python implementation \cref{code:graph} of our \textit{3D $\mathcal{G}$ exp} presented in \cref{subsec:guided}. Hints densification is performed in one pass by calling the function \textit{expand\_with\_graph} over the sparse hints matrix.

\begin{algorithm}[h]
\small
\SetAlgoLined\DontPrintSemicolon
\newcommand\mycommfont[1]{\hfill\textcolor{gray}{#1}}
\SetCommentSty{mycommfont}
\SetKwComment{Comment}{}{}%
\SetKwComment{tcp}{}{}%
\KwData{Hints Matrices $H$ data set, Radius $R$}

 \For{ $\mathbf{h}$ in $H$}{
  Init.\ 3D graph $\mathcal{G}$\;
  Init.\ expanded Hints $\mathbf{h}_{\exp} = \mathbf{h}$\; 
  Init.\ nodes $\mathcal{N}_\mathcal{G} \leftarrow \mathbf{h}$, where $\mathbf{h}$ not empty\;
  Create edges: $\mathcal{E}_\mathcal{G} = \mathbf{RGG}(\mathcal{N}_\mathcal{G}, $R$ ~ \land  \tau)$\;
  Sort edges: $\mathcal{E}_\mathcal{G} = sort(\mathcal{E}_\mathcal{G}, key: d_{\mathrm{3D}-ij})$\;
  \For{edge $E_{ij}$ in edges $\mathcal{E}_\mathcal{G}$}{
    \If(\tcp*[f]{if not adjacent}){$d_{\mathrm{2D}-ij} > \sqrt{2}$}{
        Compute $\Delta_{ij}$\;
        $\mathcal{M} = \{1 \ldots m  \ldots M \mid M < d_{\mathrm{2D}-ij}\}$\;
        \For{$m \in \mathcal{M}$}{
            \If{$\mathbf{h}_{\exp}[\textbf{r}(\delta_x m), \textbf{r}(\delta_y m)]$ is empty}{
                $\mathbf{h}_{\exp}[\textbf{r}(\delta_x m), \textbf{r}(\delta_y m)] = \delta_z m$\;
                }
            }
        }
    }
    $\mathbf{h} \leftarrow \mathbf{h}_{\exp}$\;
 }
 \caption{3D Graph Hints Expansion}
 \label{algo:3d_graph}
\end{algorithm}

\begin{lstlisting}[language=Python, caption=Code for \textit{3D $\mathcal{G}$ exp}., label={code:graph}]
import numpy as np
import networkx as nx

def expand_with_graph(hints, imgL, radius=8, tau=0.9):
"""
Function to expand sparse matrix of hints with Random Geometric Graph.
Input:
    hints: sparse matrix of hints, hints are disparity values > 0, otherwise matrix values are 0.
    imgL: left stereo image, to which disp is aligned.
    radius: maximum R used to search neighbouring hints.
    tau: similarity between pixel color.
Ourput:
    exp_hints: densified input hints
"""

    x_hints, y_hints = hints.shape
    G = nx.Graph()
    exp_hints = hints.copy()
    positions = np.transpose(np.nonzero(hints))
    if len(positions) < 1:
        return exp_hints
    for i in range(len(positions)):
        G.add_node(i)
        G.nodes[i]['pos']=[positions[i][0], positions[i][1], hints[positions[i][0],positions[i][1]]]
        G.nodes[i]['d_pos']=[positions[i][0], positions[i][1]]
        G.nodes[i]['disp']=hints[positions[i][0],positions[i][1]]

    nodes = G.nodes()
    pos = nx.get_node_attributes(G, 'pos')
    d_pos = nx.get_node_attributes(G, 'd_pos')

    RGG = nx.random_geometric_graph(nodes, dim=3, radius=radius, pos=pos)
    if RGG.number_of_edges() > 0:
        edges = sorted(RGG.edges(), key=lambda t: np.sqrt((pos[t[0]][0] - pos[t[1]][0]) ** 2 + (pos[t[0]][1] - pos[t[1]][1]) ** 2))
        nodes = list(RGG.nodes(data=True))
        
        for node_1, node_2 in edges:
            dist = np.sqrt((pos[node_1][0] - pos[node_2][0]) ** 2 + (pos[node_1][1] - pos[node_2][1]) ** 2)
            color_dist = np.dot(imgL[:,pos[node_1][0],pos[node_1][1]], imgL[:,pos[node_2][0],pos[node_2][1]])/(np.linalg.norm((imgL[:,pos[node_1][0],pos[node_1][1])*np.linalg.norm()(imgL[:,pos[node_2][0],pos[node_2][1])
            ceil_dist = np.ceil(dist).astype('int')
            if dist <= 1.42 or color_dist < threshold: # approximate sqrt(2)
                continue
            else:
                space = np.linspace(0, ceil_dist, ceil_dist+1)
                values = np.interp(space, [0, dist], [pos[node_1][2], pos[node_2][2]])

                slope = (pos[node_1][1] - pos[node_2][1]) / (pos[node_1][0] - pos[node_2][0] + 1e-8) #1e-8 added for numerical stability (in case of vertical line).
                rad_slope = np.arctan(slope)
                dx = np.cos(rad_slope)
                dy = np.sin(rad_slope)
                for interval, val in zip(space, values):
                    x = (pos[node_1][0] + np.rint(interval * dx)).astype('int')
                    y = (pos[node_1][1] + np.rint(interval * dy)).astype('int')
                    if x < x_hints and y < y_hints and exp_hints[x][y] == 0:
                        exp_hints[x][y] = val
    return exp_hints
\end{lstlisting}





\begin{table*}[h]
	\setlength{\tabcolsep}{9pt}
	\centering
		\begin{tabularx}{\textwidth}{l|ccccc|ccccc}
			\toprule\noalign{\smallskip}
			\multirow{2}{*}{\sc Model} & \multicolumn{5}{c|}{\sc ETH3D}& \multicolumn{5}{c}{\sc TartanAIR}\\ 
			
			& MAE & ${>}2$ & ${>}3$ &${>}4$ & ${>}5$ & MAE & ${>}2$ &${>}3$ &${>}4$ & ${>}5$ \\
			\noalign{\smallskip}
			\hline
			\noalign{\smallskip}
			\textit{PSMNet} \cite{chang2018pyramid}  & 5.25 & 16.99 &7.62 & 5.82& 5.10 & 5.51 & 21.30 &14.09&11.34& 9.81 \\ 
			\textit{PSMNet} \cite{chang2018pyramid} \textit{Vgd-test} & 5.25 & 16.78 &7.65 & 5.80& 5.10 & 5.51 & 21.29 &14.09 & 11.30& 9.80\\
			\textit{Vgd} & 1.54 & 8.59 &5.47&4.17& 3.41 & 5.53 & 21.03 &16.62&14.14& 12.45 \\
			\textit{Vgd} + \textit{Scaffolding} \cite{wong2020unsupervised} & 1.10 & 9.44 & 5.50 & 3.85 & 2.94 & 2.98 & 18.23 & 13.68 & 11.23 & 9.64 \\
			\textit{Vgd} + \textit{Lin 3D exp} & 1.19 & 8.8 &5.16&3.63& 2.77 & 2.74 & 17.86 &13.29&10.81& 9.24 \\
			\textit{Vgd} + \textit{3D $\mathcal{G}$ exp}  & \textbf{1.04} & \textbf{7.91} &\textbf{4.41}&\textbf{3.01}& \textbf{2.27} & \textbf{2.54} & \textbf{17.00} &\textbf{12.72}&\textbf{10.48}& \textbf{9.04} \\
			\bottomrule
		\end{tabularx}
	\caption{Ablation study of our proposed methods on PSMNet \cite{chang2018pyramid}. \textit{Vgd} is a sparse visual hints guided model, 
			\textit{3D $\mathcal{G}$ exp} is the 3D graph expansion (\cref{subsec:expansion}) and \textit{Lin 3D exp} the linear 3D expansion (\cref{subsec:fast_3d})}
	\vspace{-8pt}
	\label{tab:ablation_psmnet_scaff}
\end{table*}

\section{\textit{Lin 3D exp} \& \textit{3D $\mathcal{G}$ exp} Comparison}

The \textit{Lin 3D exp} expansion is performed in two steps, unlike \textit{3D $\mathcal{G}$ exp}. The linear expansion is devised to connect with a slanted line points only on vertical or horizontal directions. Moreover, the points must be on the same vertical or horizontal line inside a small patch of the image. The chance of densifying sparse points with this method is lower compared to \textit{3D $\mathcal{G}$ exp}. One could imagine that in \textit{3D $\mathcal{G}$ exp} hints are connected with slanted lines if they are close in the 3D space, thus densification is a one-pass process. On the other hand, \textit{Lin 3D exp}, simply connects close points in based on their 2D location, for this reason two passes are performed (one vertical and one horizontal) to densify the hints.

\section{Comparison with Scaffolding}

In addition to the results presented in the PSMNet ablation study \cref{tab:ablation_psmnet} we compare with Scaffolding \cite{wong2020unsupervised} in \cref{tab:ablation_psmnet_scaff}. In \cite{wong2020unsupervised} they proposed an expansion based on convex hulls interpolation. Although, in our case it did not produce any improvement. A possible explanation is that our VIO guidance is too sparse and the expanded hints harm performance.

\section{Qualitative Results}

\begin{figure*}
	\vspace{-8pt}
	\centering
	\setlength{\mywidth}{.53\textwidth}%
	
	\newcommand{\plot}[4]{%
		\node[anchor=north west,draw=black!50,minimum width=\mywidth,
		minimum height=.75\mywidth,rounded corners=2pt,path picture={
			\node at (path picture bounding box.north west){
				\includegraphics[width=\mywidth]{#3}};
		}] (fig) at (#1,#2) {};
		\node[anchor=north east,draw=black!50,align=left,inner sep=1pt,
		fill=white,rounded corners=2pt,minimum height=10pt,opacity=.75] at 
		(fig.37) {\footnotesize #4};
	}
	\begin{tikzpicture}[inner sep=0,outer sep=0]  
		\plot{0}{0}{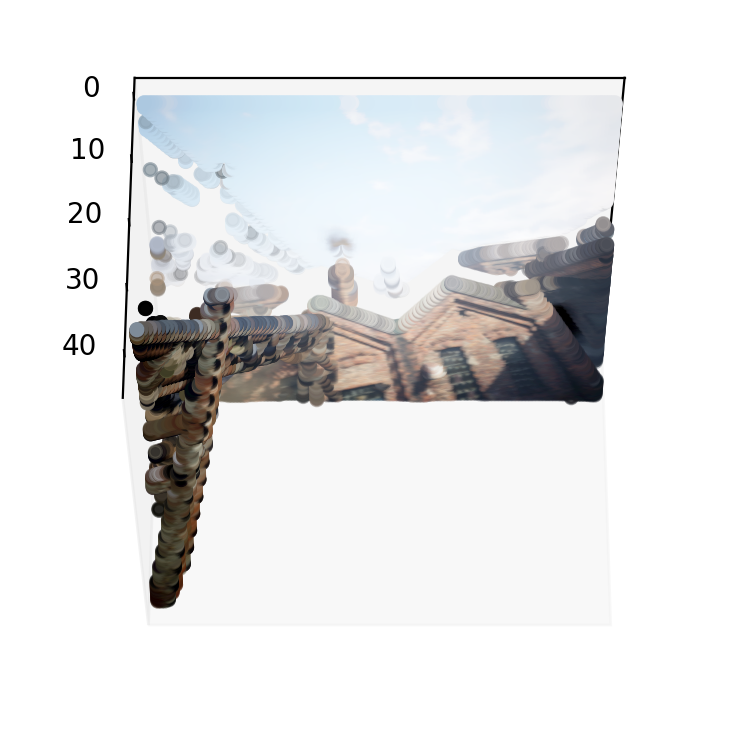}{GT}
		\plot{0}{-7}{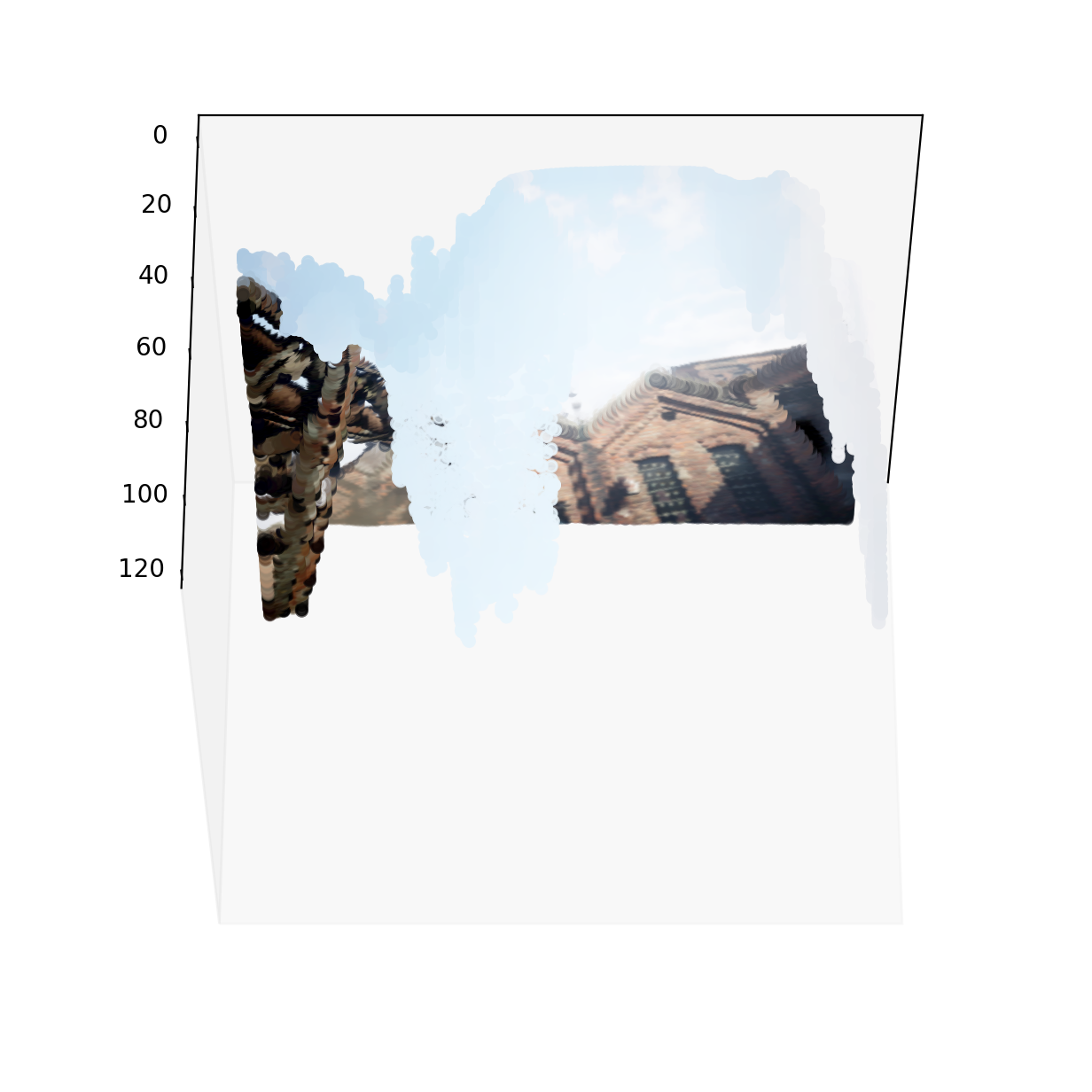}{Vanilla}
		\plot{0}{-14}{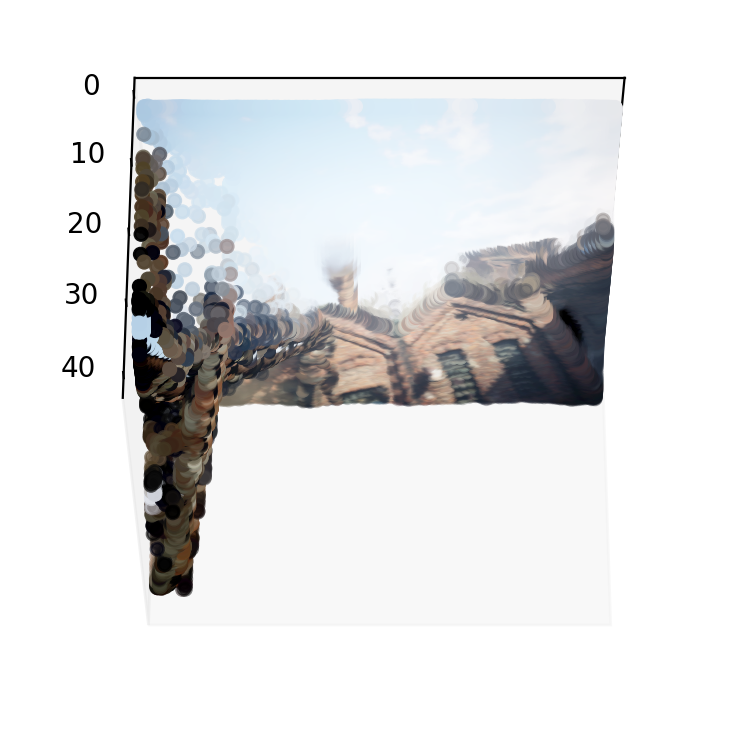}{$\mathrm{3D} \mathcal{G} \mathrm{exp}$}
	\end{tikzpicture}
    \caption{3D point cloud visualization of \cref{fig:teaser}, our method greatly reduces artifacts due to out-of-domain data. The numbers represent disparity values, in the case of vanilla model the disparity range is $3\times$ higher than with guidance $\mathrm{3D} \mathcal{G} \mathrm{exp}$ or \emph{GT}}
    \label{fig:supp_pointcloud}
\end{figure*}

\begin{figure*}
    \centering
    \setlength{\mywidth}{0.16\textwidth}%
  
  \newcommand{\plot}[4]{%
    \node[anchor=north west,draw=black!50,minimum width=\mywidth,
      minimum height=.75\mywidth,rounded corners=2pt,path picture={
      \node at (path picture bounding box.north west){
        \includegraphics[width=\mywidth]{#3}};
      }] (fig) at (#1,#2) {};
    \node[anchor=north east,draw=black!50,align=left,inner sep=1pt,
      fill=white,rounded corners=2pt,minimum height=10pt,opacity=.75] at 
      (fig.37) {\footnotesize #4};
  }

    \begin{tikzpicture}[inner sep=0,outer sep=0]  
        \plot{0}{0}{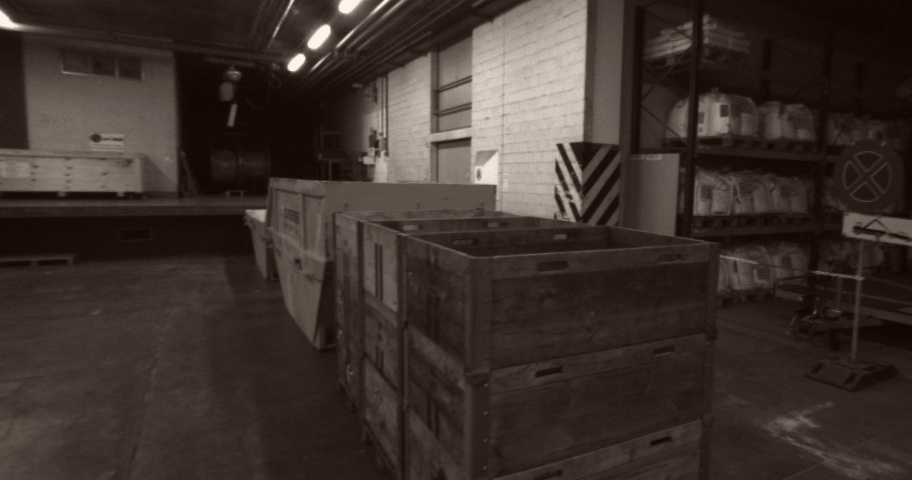}{RGB}
        \plot{1.02\mywidth}{0}{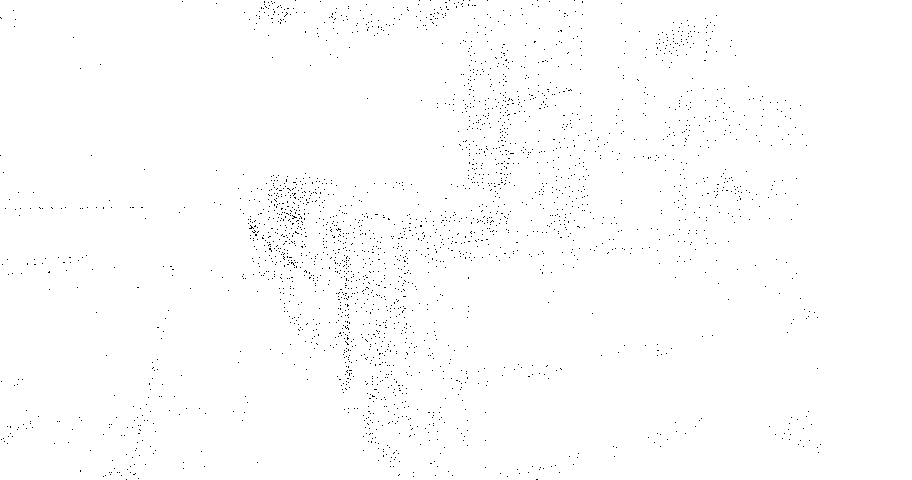}{H}
        \plot{2.04\mywidth}{0}{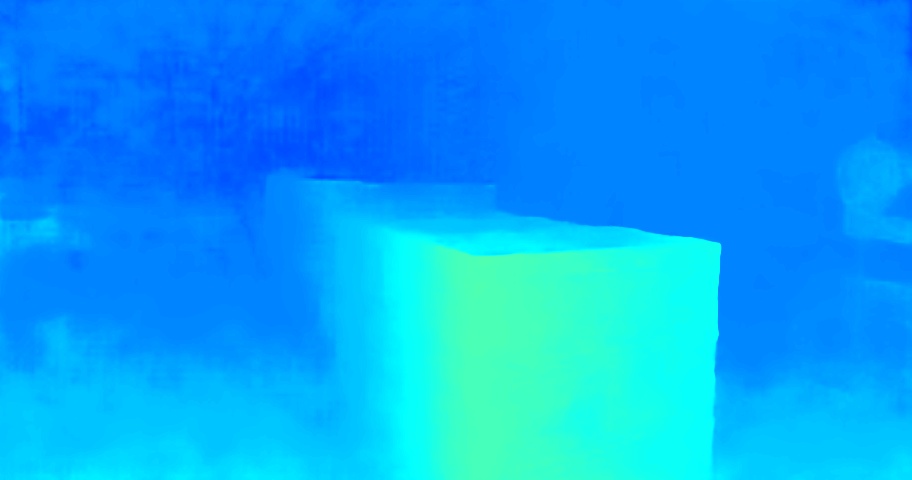}{disp}
        \plot{3.06\mywidth}{0}{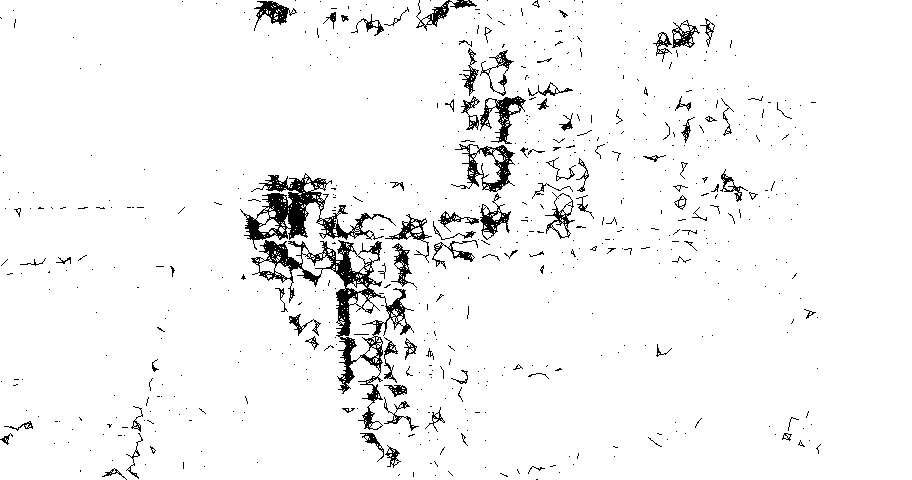}{H$_{\mathbf{3D} \mathcal{G} \mathbf{exp}}$}
        \plot{4.08\mywidth}{0}{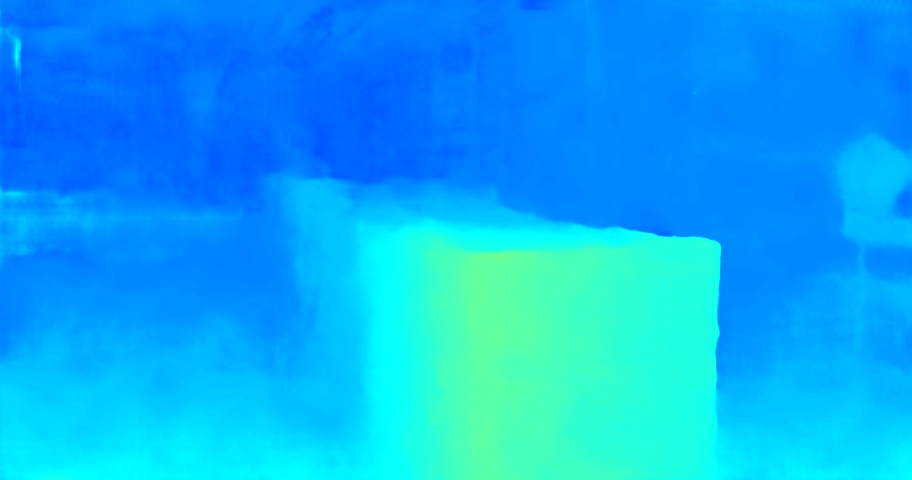}{disp$_{\mathrm{3D} \mathcal{G} \mathrm{exp}}$}
        \plot{5.10\mywidth}{0}{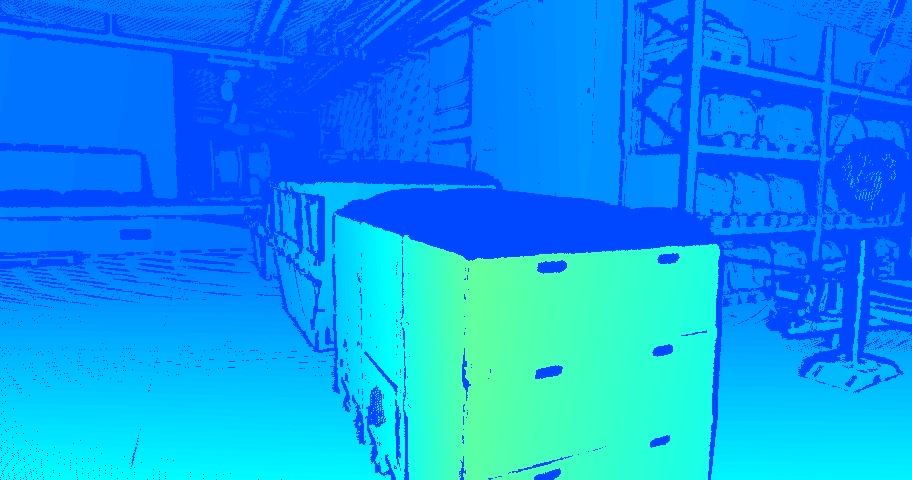}{GT}
        \plot{0}{-0.8\mywidth}{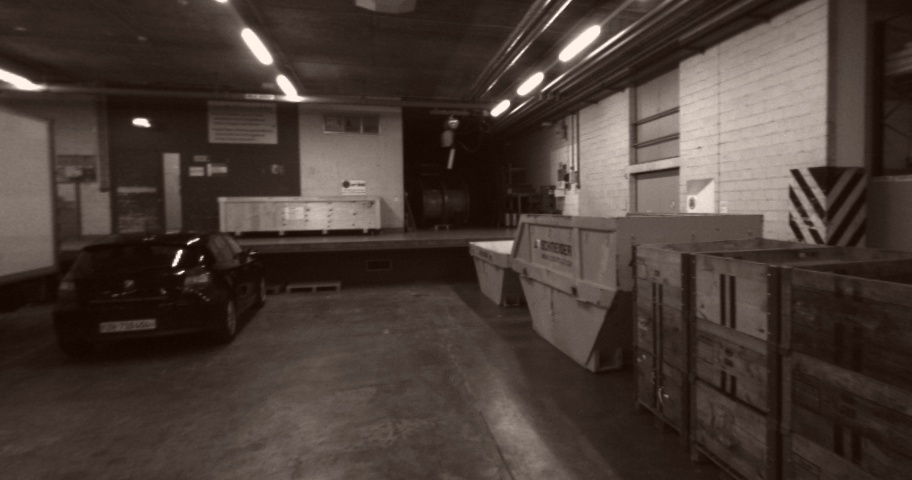}{RGB}
        \plot{1.02\mywidth}{-0.8\mywidth}{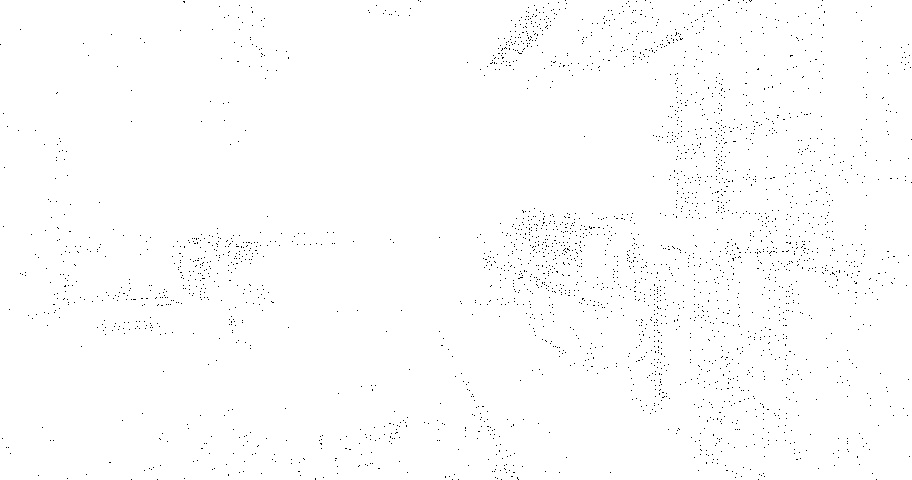}{H}
        \plot{2.04\mywidth}{-0.8\mywidth}{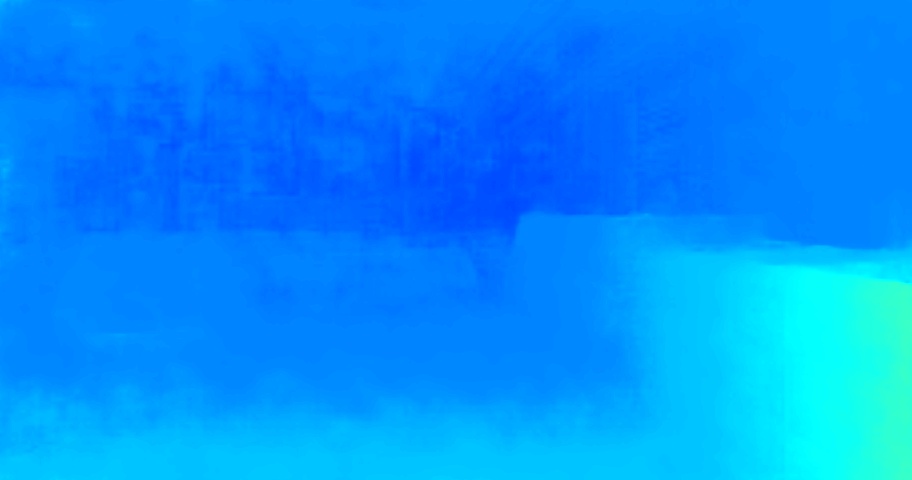}{disp}
        \plot{3.06\mywidth}{-0.8\mywidth}{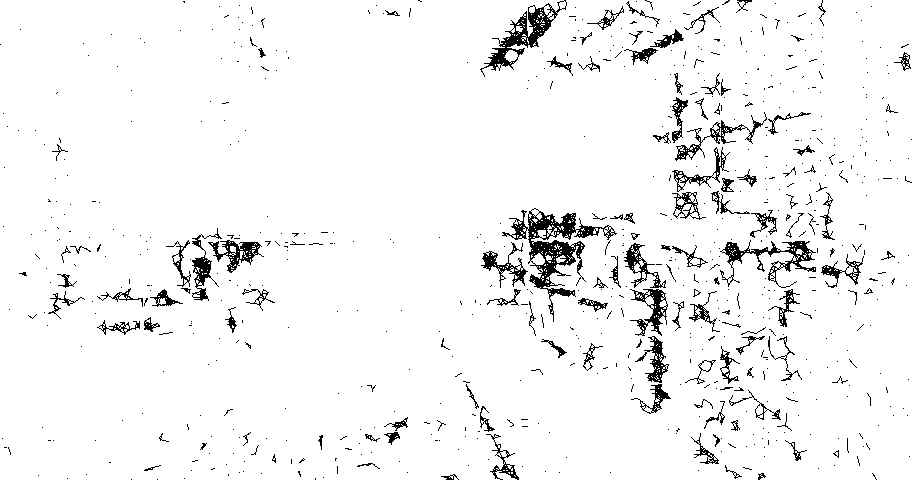}{H$_{\mathbf{3D} \mathcal{G} \mathbf{exp}}$}
        \plot{4.08\mywidth}{-0.8\mywidth}{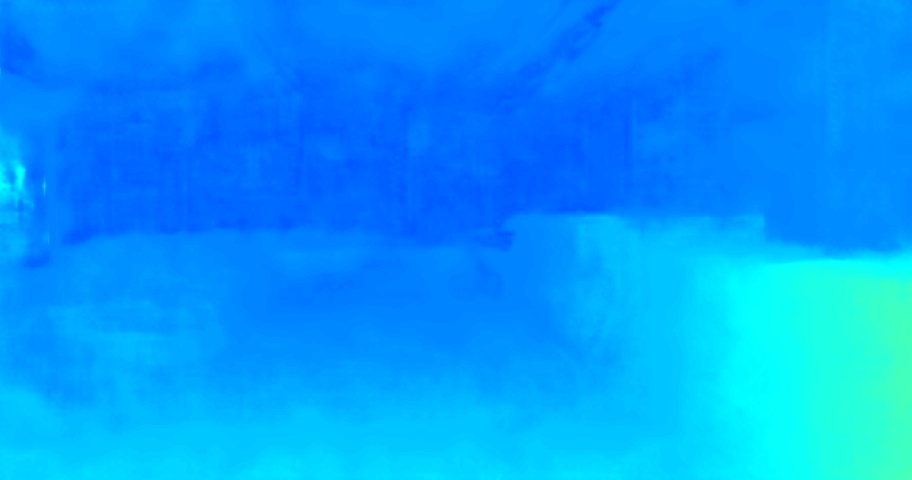}{disp$_{\mathrm{3D} \mathcal{G} \mathrm{exp}}$}
        \plot{5.10\mywidth}{-0.8\mywidth}{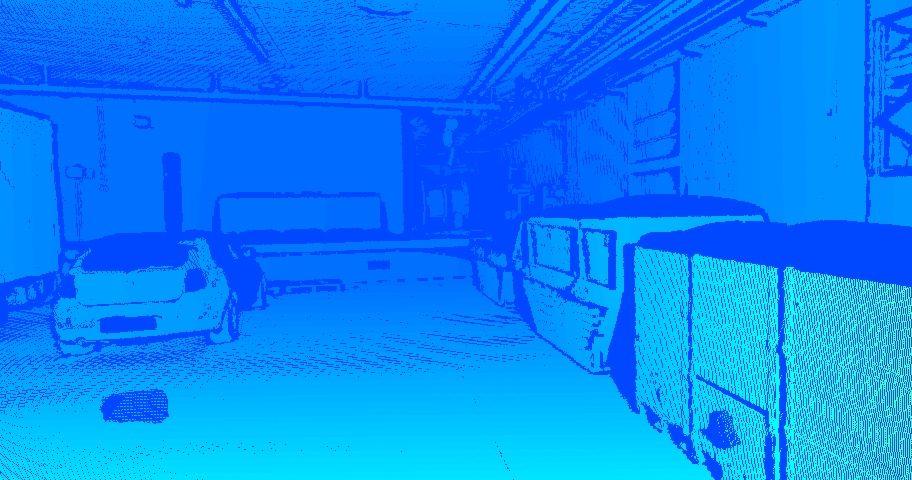}{GT}
        \plot{0}{-1.6\mywidth}{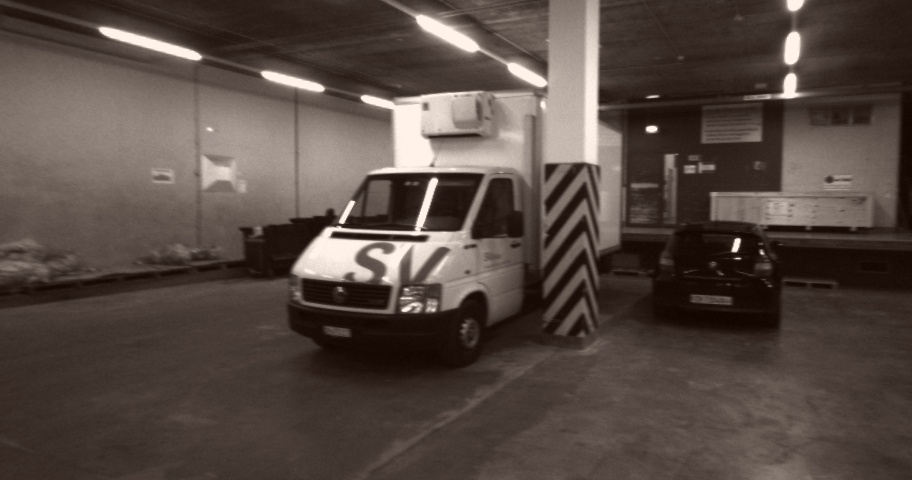}{RGB}
        \plot{1.02\mywidth}{-1.6\mywidth}{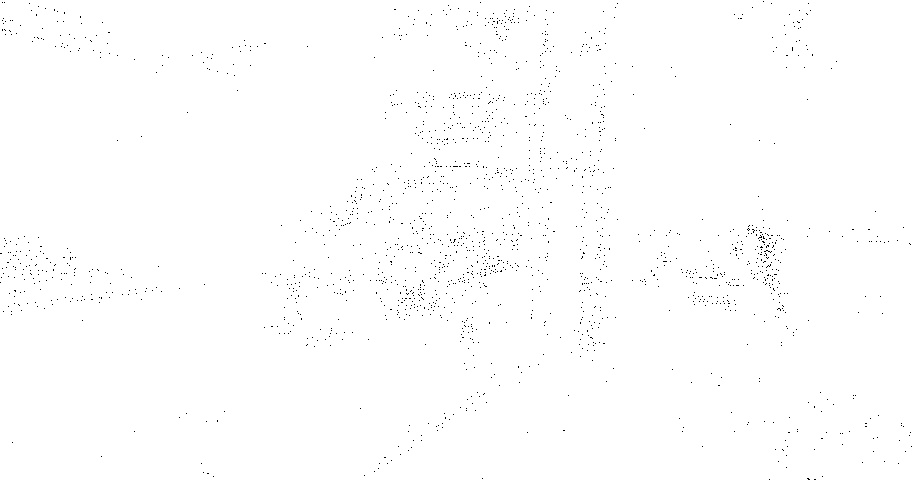}{H}
        \plot{2.04\mywidth}{-1.6\mywidth}{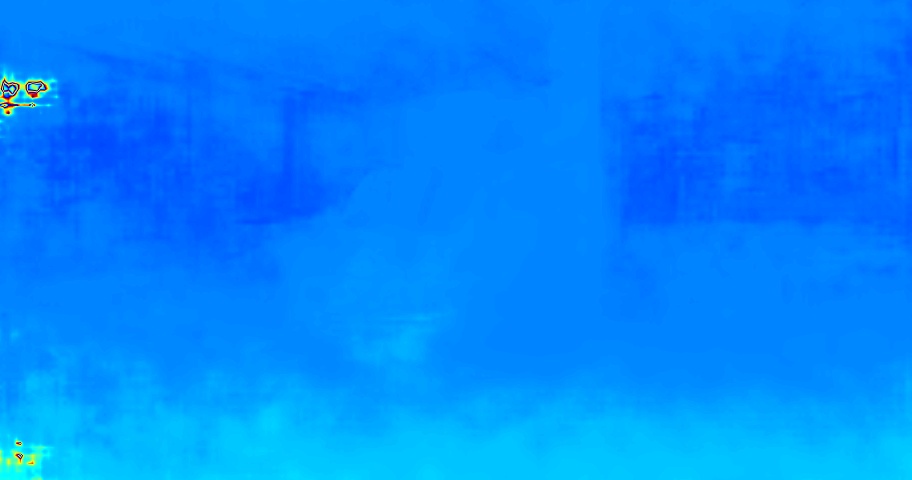}{disp}
        \plot{3.06\mywidth}{-1.6\mywidth}{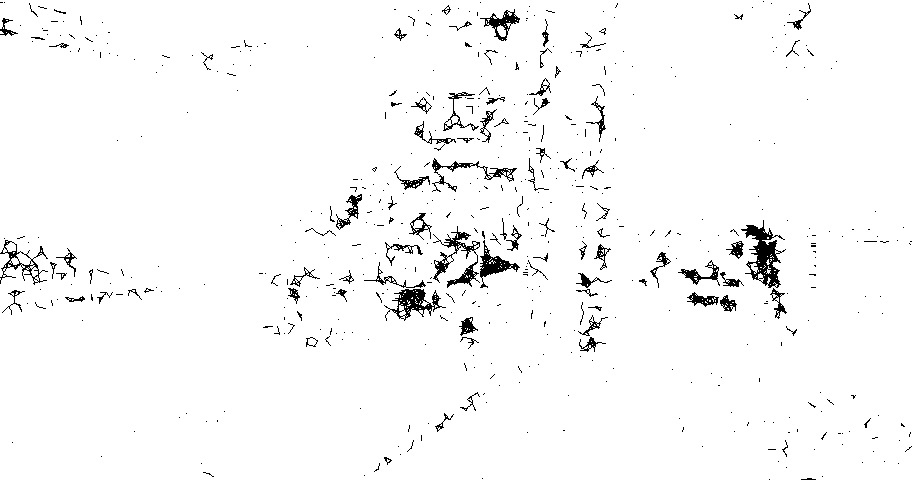}{H$_{\mathbf{3D} \mathcal{G} \mathbf{exp}}$}
        \plot{4.08\mywidth}{-1.6\mywidth}{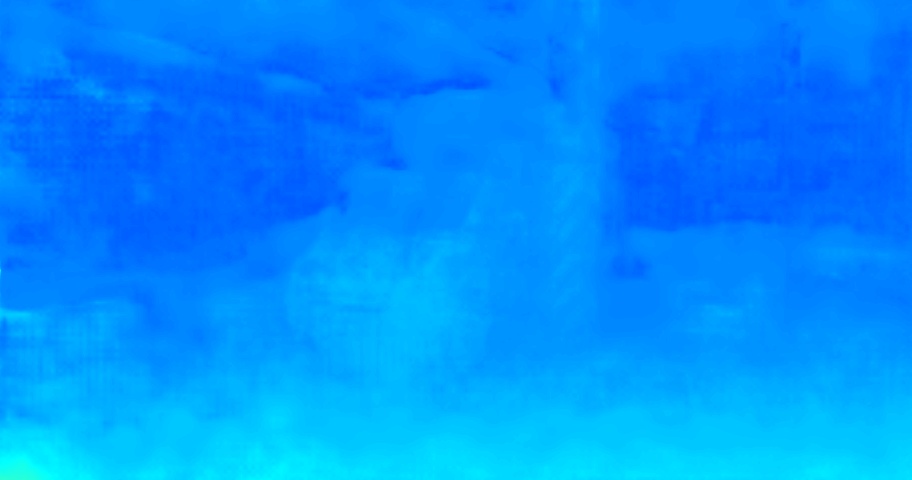}{disp$_{\mathrm{3D} \mathcal{G} \mathrm{exp}}$}
        \plot{5.10\mywidth}{-1.6\mywidth}{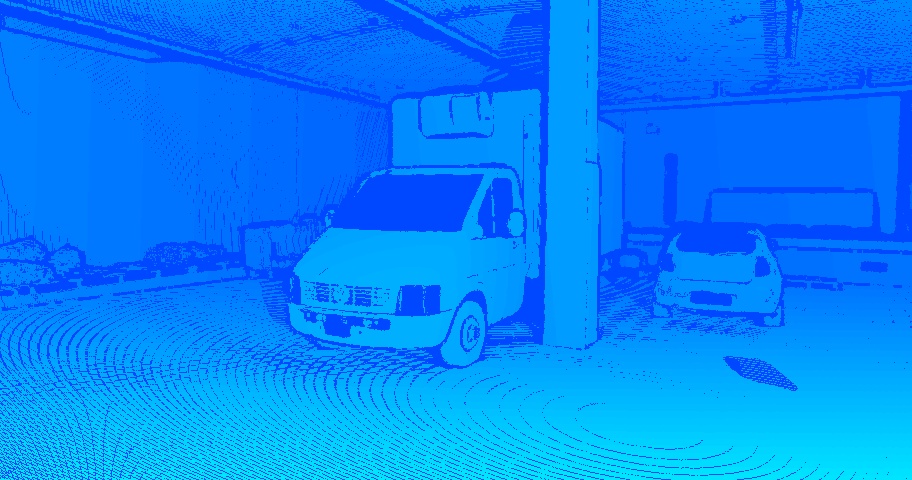}{GT}
        \plot{0}{-2.4\mywidth}{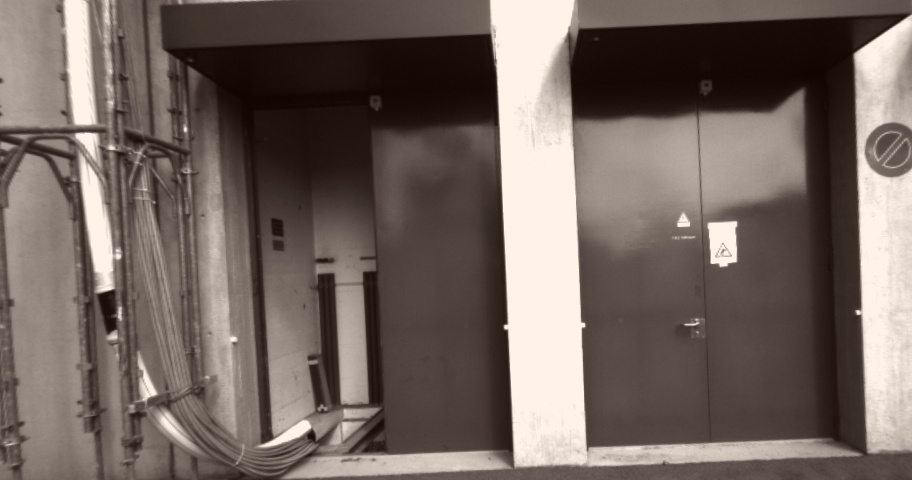}{RGB}
        \plot{1.02\mywidth}{-2.4\mywidth}{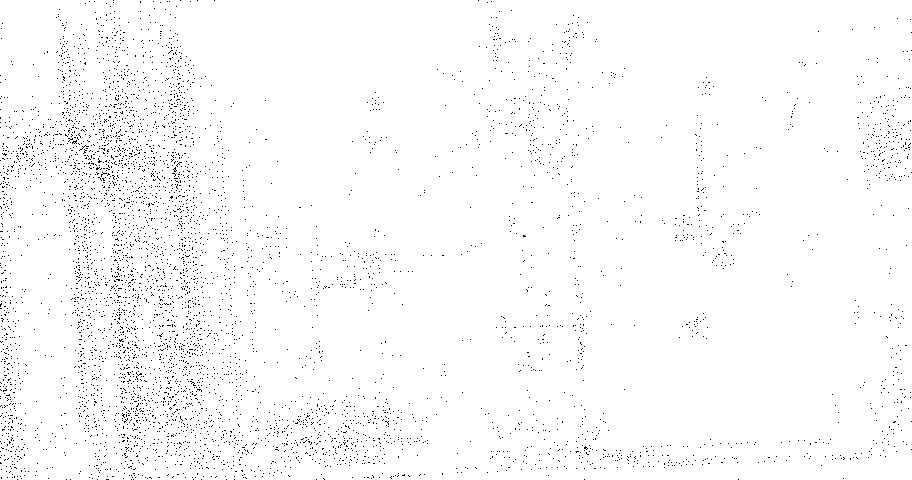}{H}
        \plot{2.04\mywidth}{-2.4\mywidth}{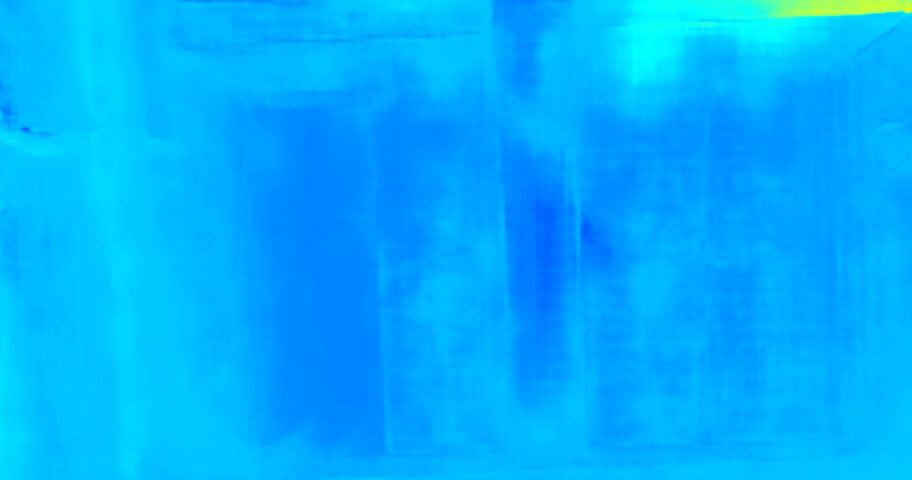}{disp}
        \plot{3.06\mywidth}{-2.4\mywidth}{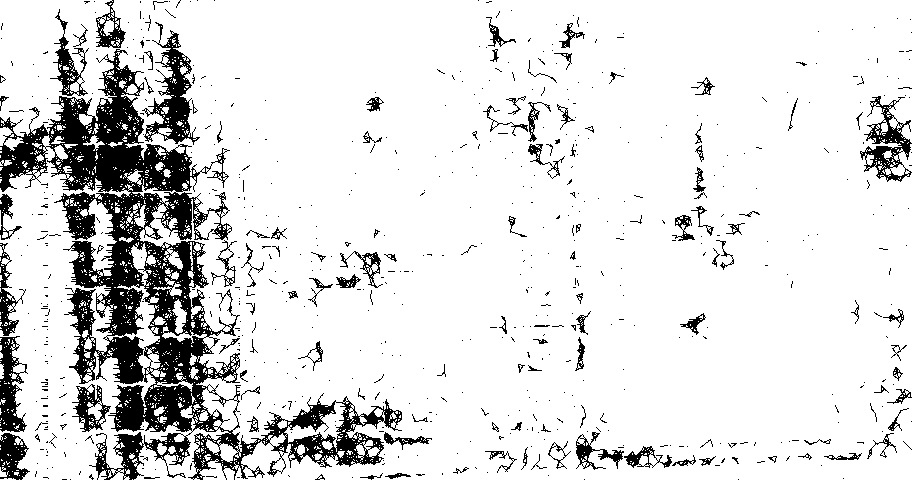}{H$_{\mathbf{3D} \mathcal{G} \mathbf{exp}}$}
        \plot{4.08\mywidth}{-2.4\mywidth}{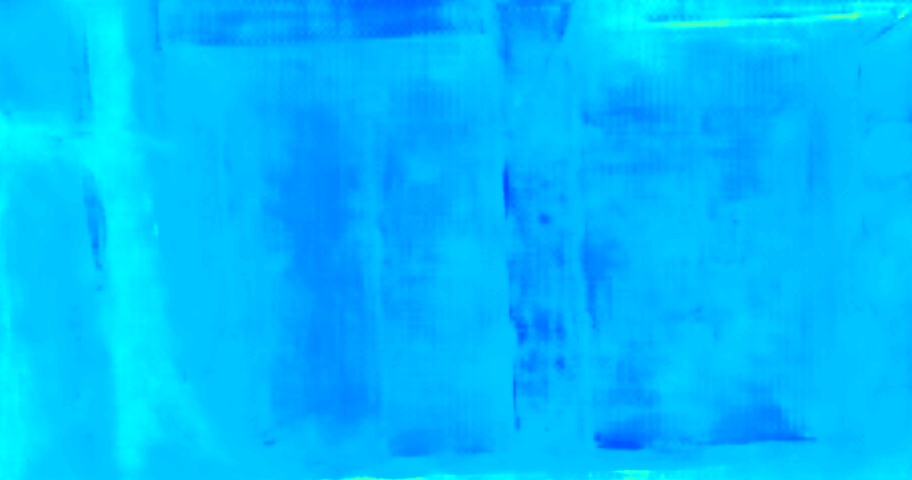}{disp$_{\mathrm{3D} \mathcal{G} \mathrm{exp}}$}
        \plot{5.10\mywidth}{-2.4\mywidth}{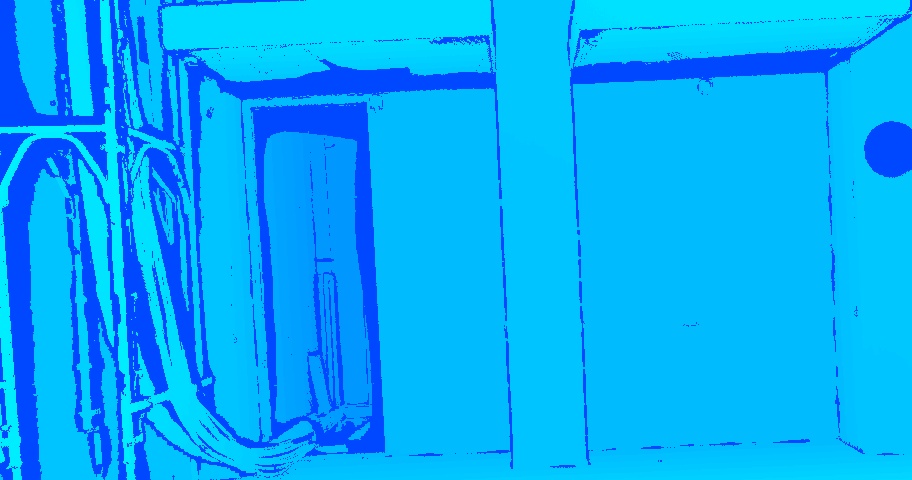}{GT}
        \plot{0}{-3.2\mywidth}{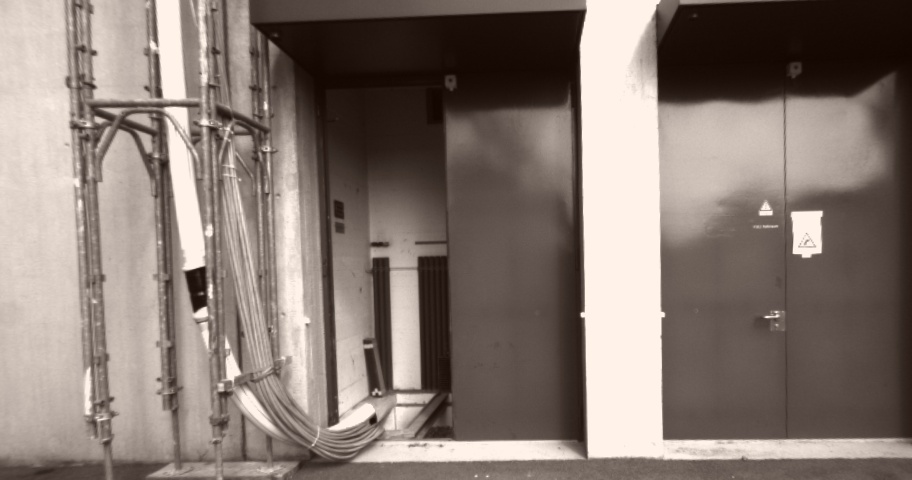}{RGB}
        \plot{1.02\mywidth}{-3.2\mywidth}{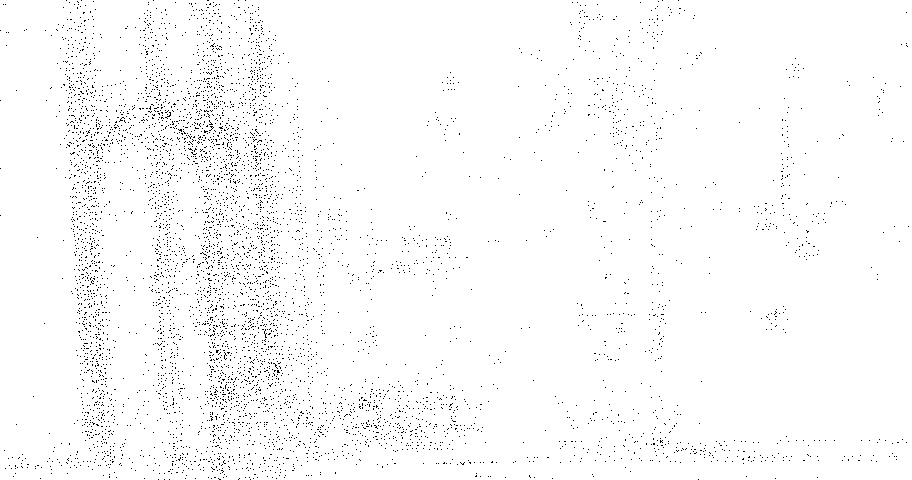}{H}
        \plot{2.04\mywidth}{-3.2\mywidth}{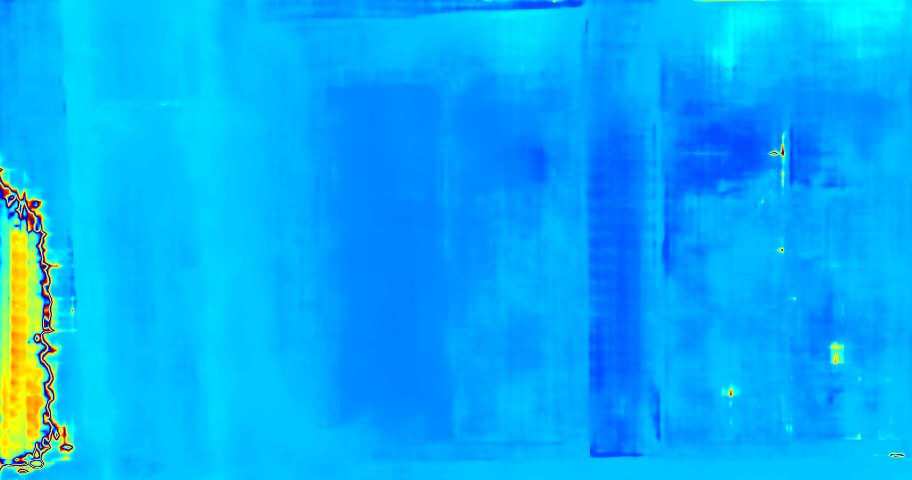}{disp}
        \plot{3.06\mywidth}{-3.2\mywidth}{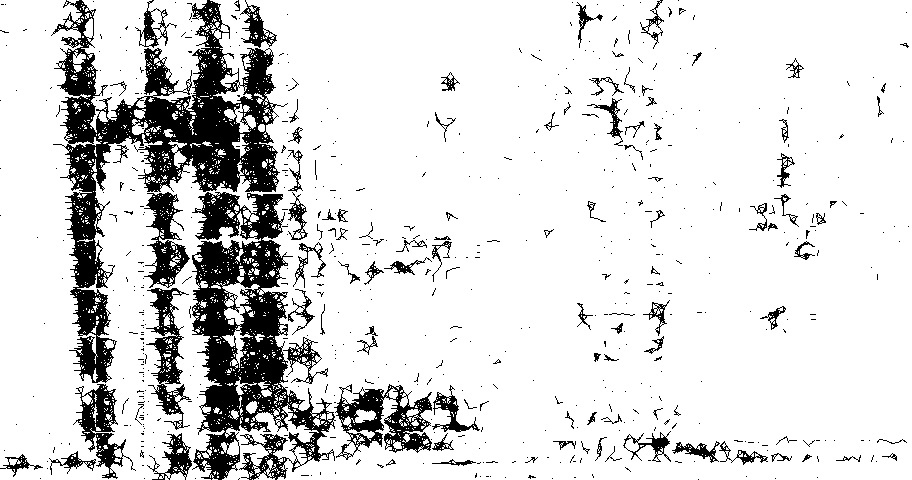}{H$_{\mathbf{3D} \mathcal{G} \mathbf{exp}}$}
        \plot{4.08\mywidth}{-3.2\mywidth}{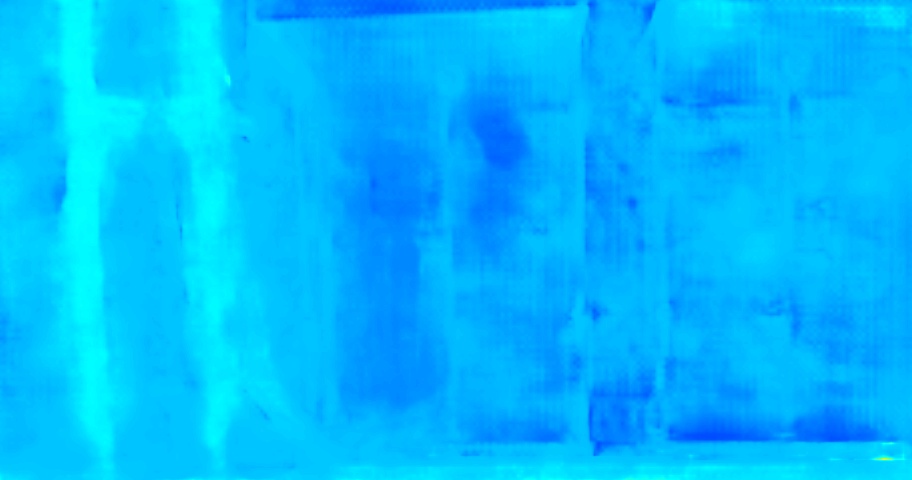}{disp$_{\mathrm{3D} \mathcal{G} \mathrm{exp}}$}
        \plot{5.10\mywidth}{-3.2\mywidth}{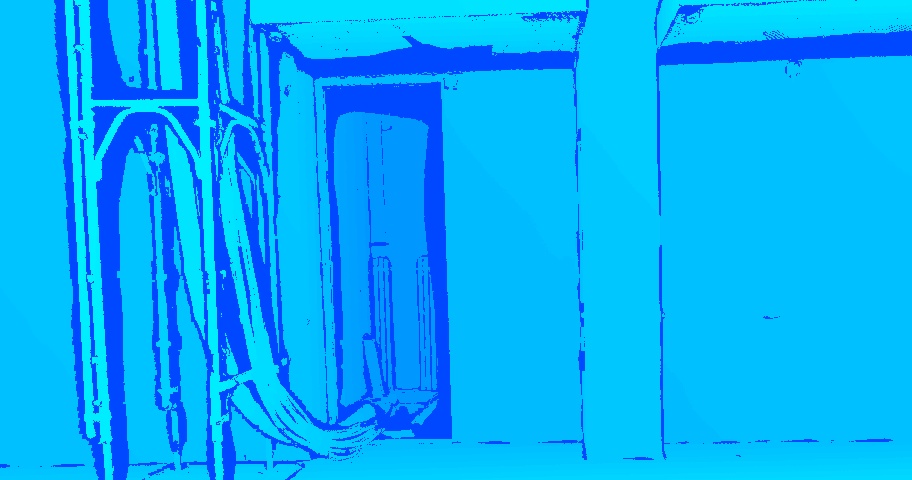}{GT}
        \plot{0}{-4\mywidth}{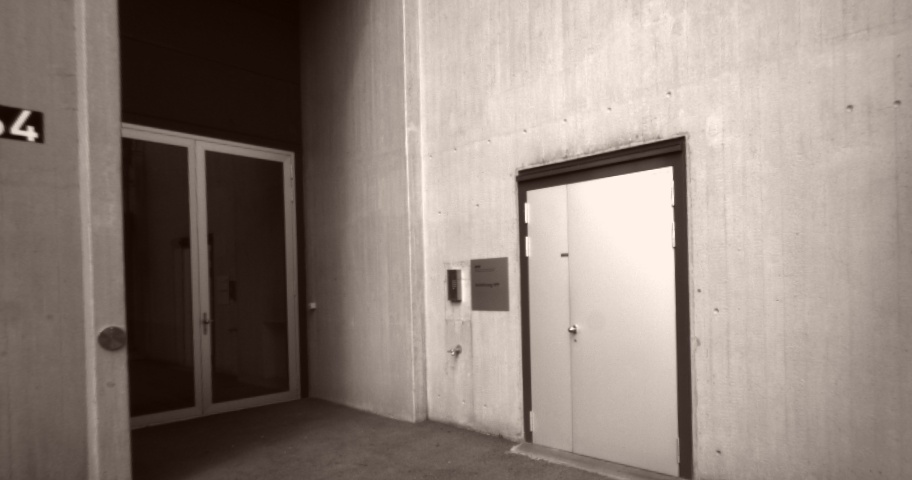}{RGB}
        \plot{1.02\mywidth}{-4\mywidth}{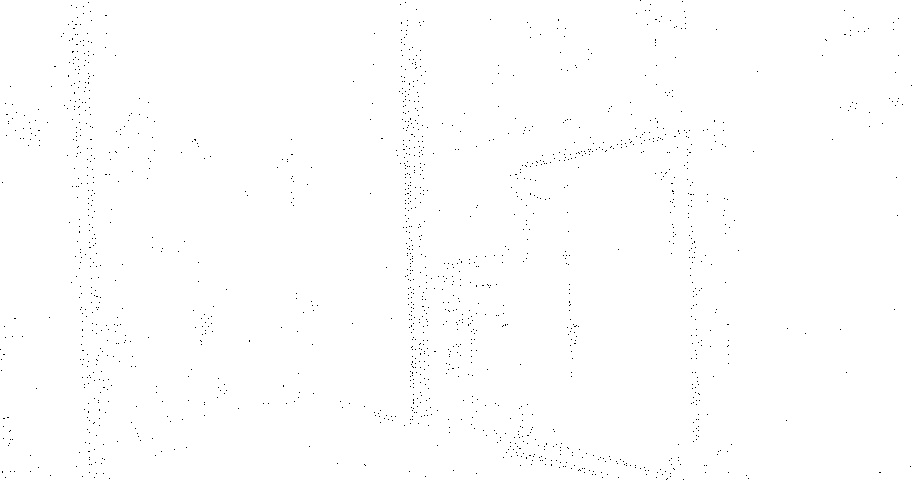}{H}
        \plot{2.04\mywidth}{-4\mywidth}{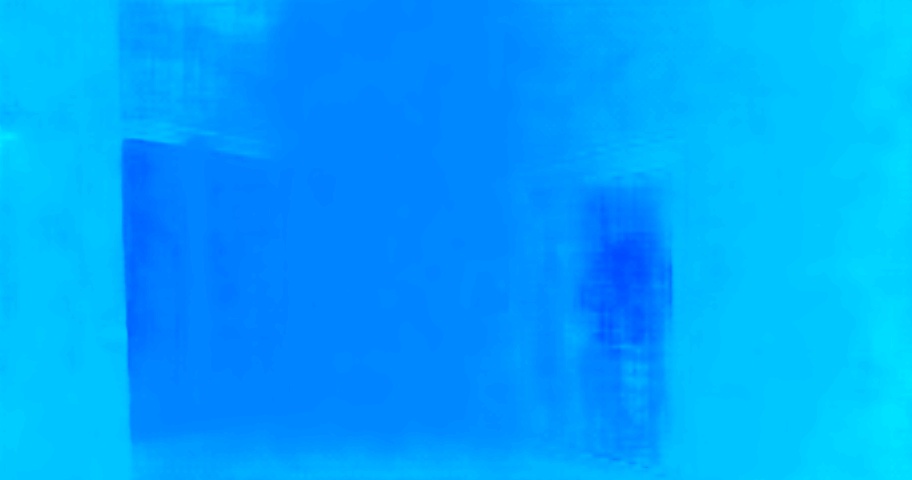}{disp}
        \plot{3.06\mywidth}{-4\mywidth}{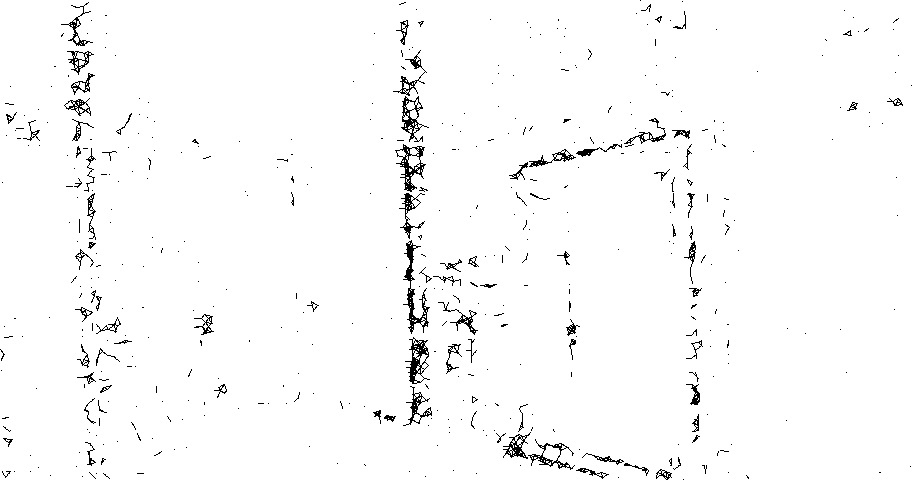}{H$_{\mathbf{3D} \mathcal{G} \mathbf{exp}}$}
        \plot{4.08\mywidth}{-4\mywidth}{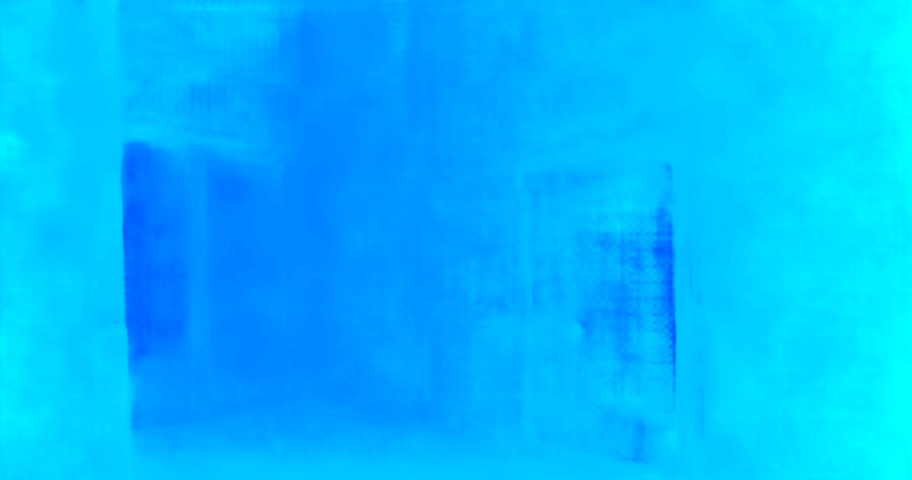}{disp$_{\mathrm{3D} \mathcal{G} \mathrm{exp}}$}
        \plot{5.10\mywidth}{-4\mywidth}{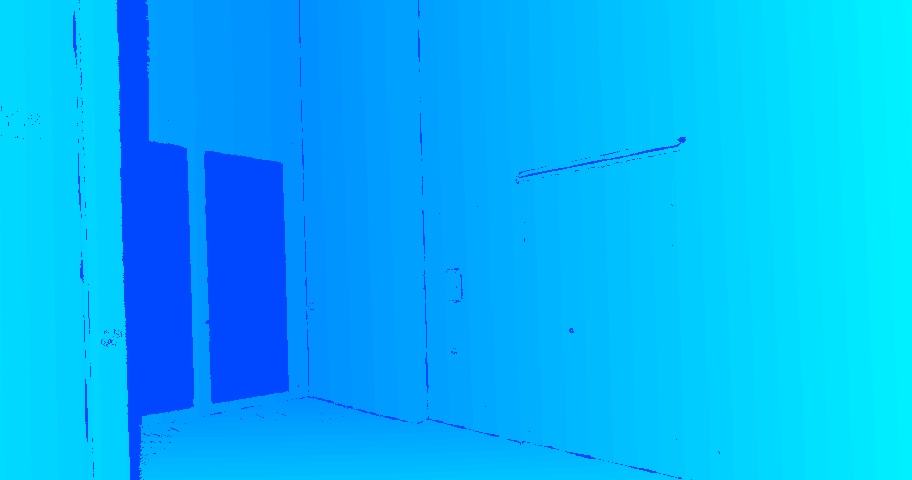}{GT}
        \plot{0}{-4.8\mywidth}{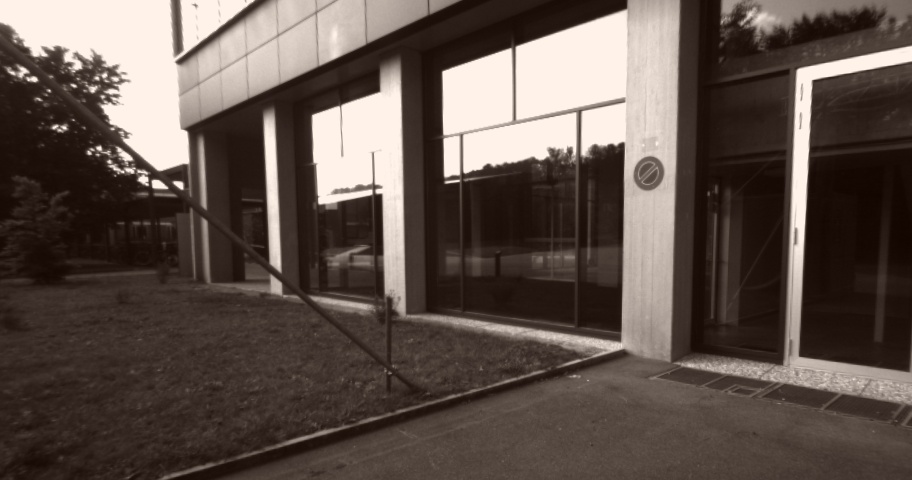}{RGB}
        \plot{1.02\mywidth}{-4.8\mywidth}{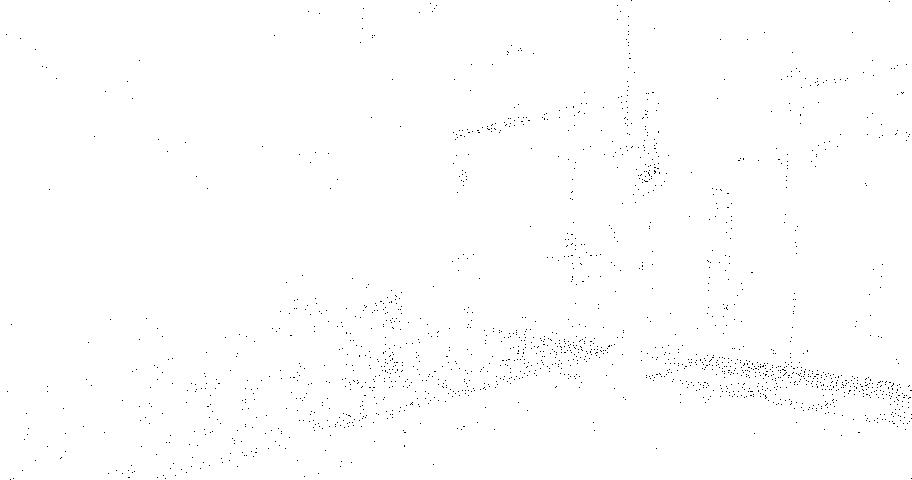}{H}
        \plot{2.04\mywidth}{-4.8\mywidth}{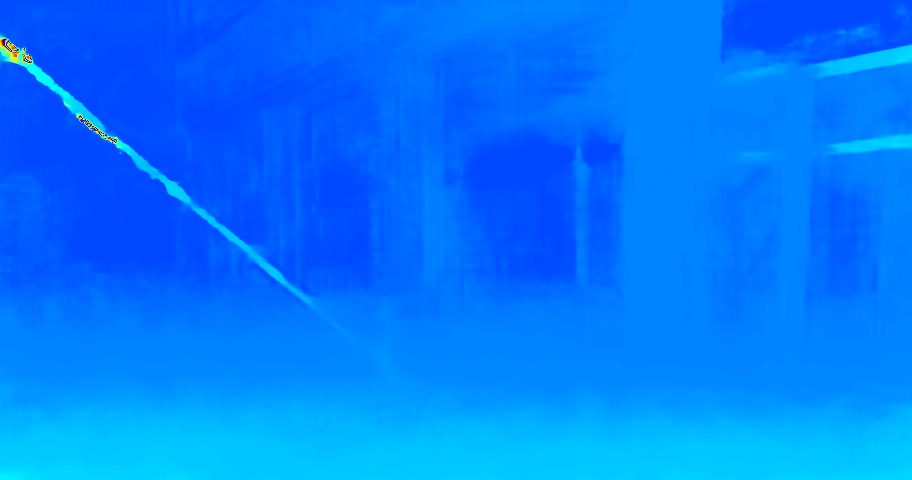}{disp}
        \plot{3.06\mywidth}{-4.8\mywidth}{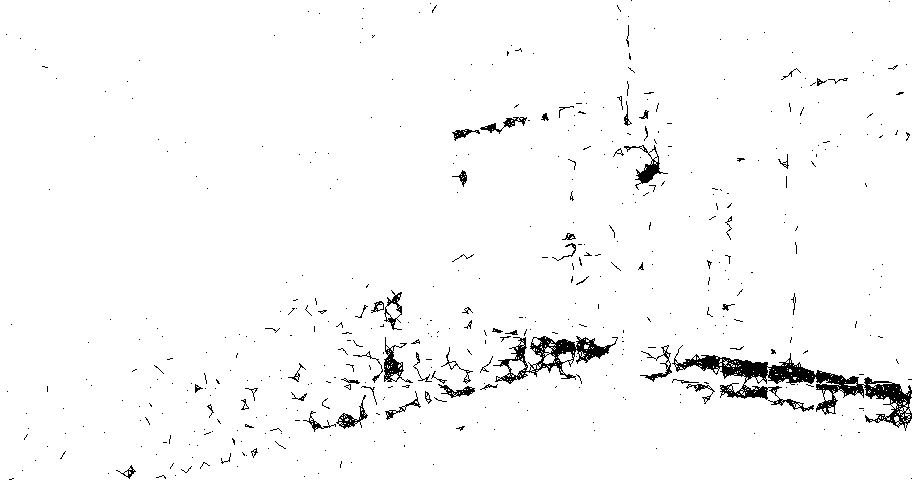}{H$_{\mathbf{3D} \mathcal{G} \mathbf{exp}}$}
        \plot{4.08\mywidth}{-4.8\mywidth}{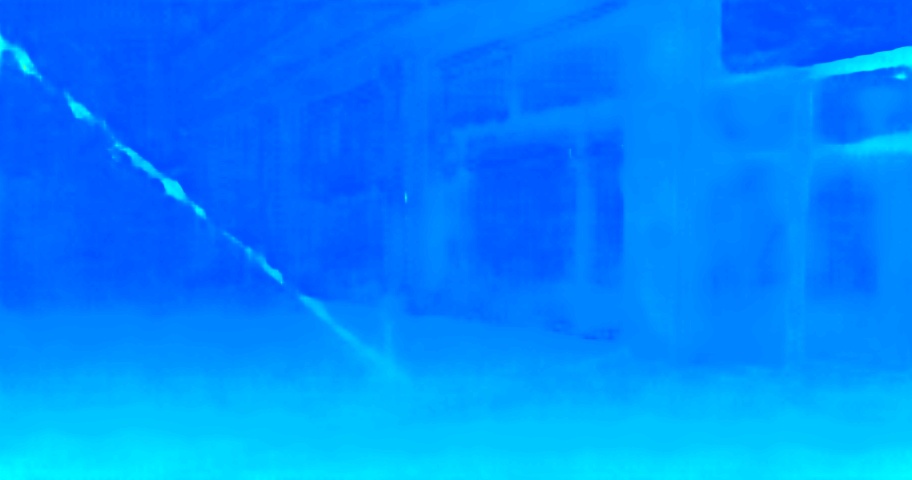}{disp$_{\mathrm{3D} \mathcal{G} \mathrm{exp}}$}
        \plot{5.10\mywidth}{-4.8\mywidth}{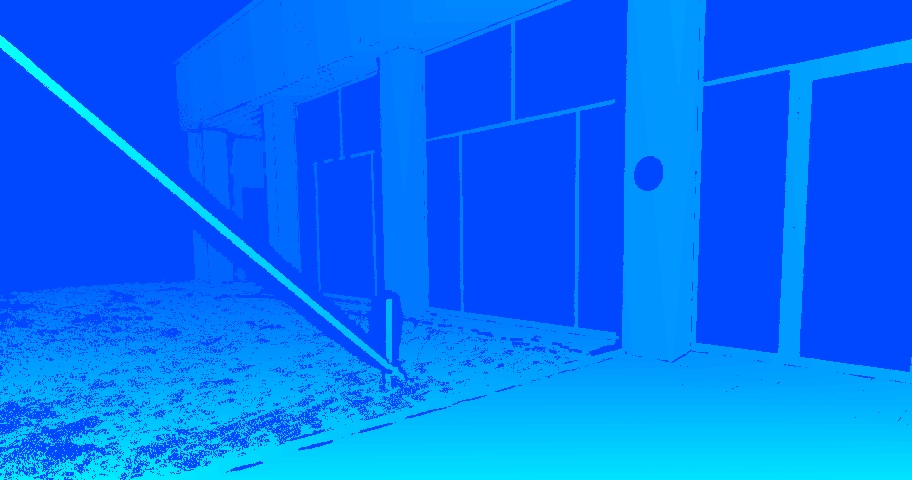}{GT}
        \plot{0}{-5.6\mywidth}{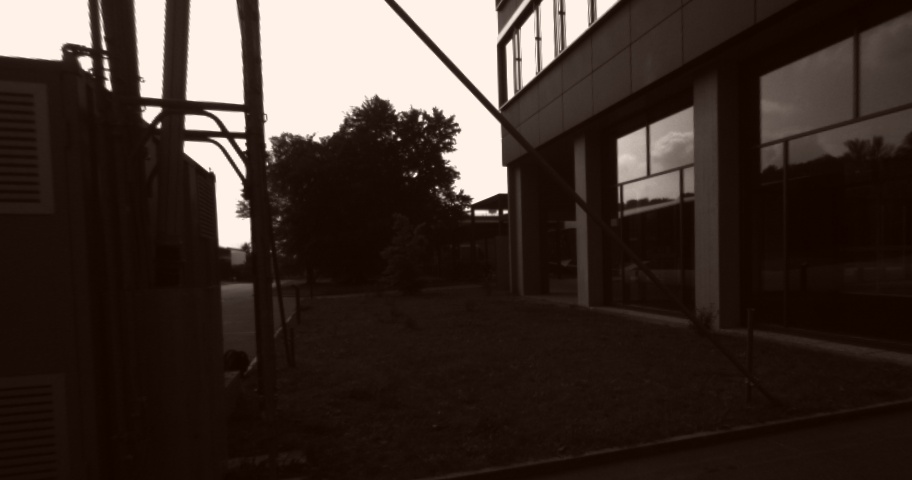}{RGB}
        \plot{1.02\mywidth}{-5.6\mywidth}{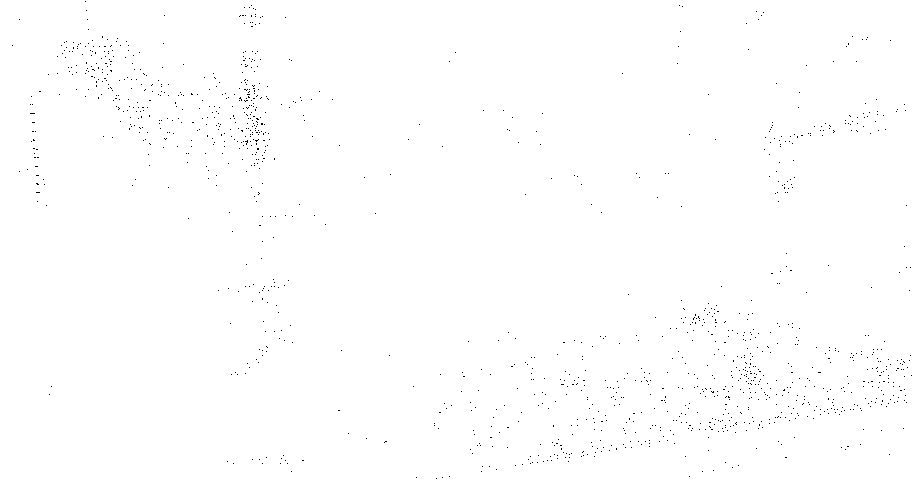}{H}
        \plot{2.04\mywidth}{-5.6\mywidth}{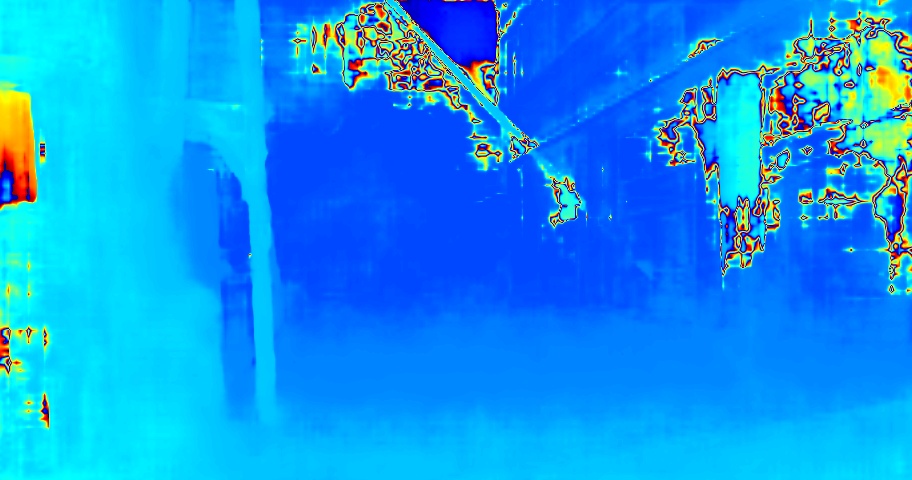}{disp}
        \plot{3.06\mywidth}{-5.6\mywidth}{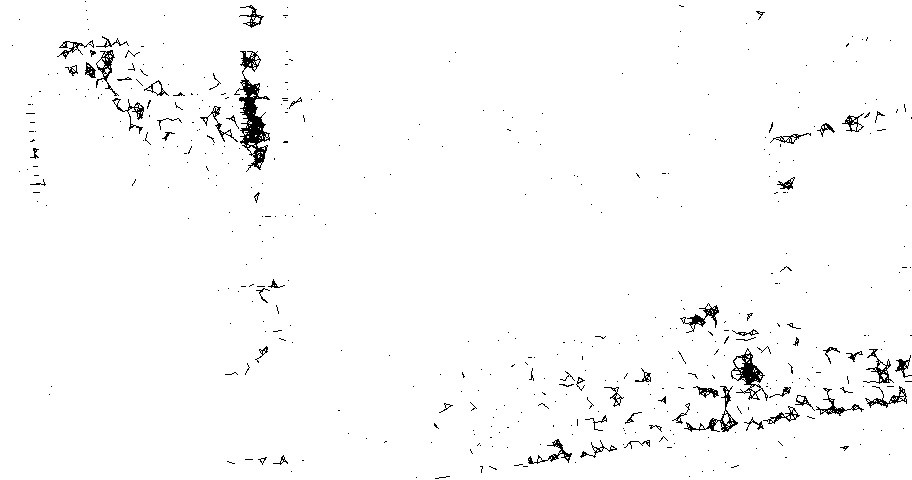}{H$_{\mathbf{3D} \mathcal{G} \mathbf{exp}}$}
        \plot{4.08\mywidth}{-5.6\mywidth}{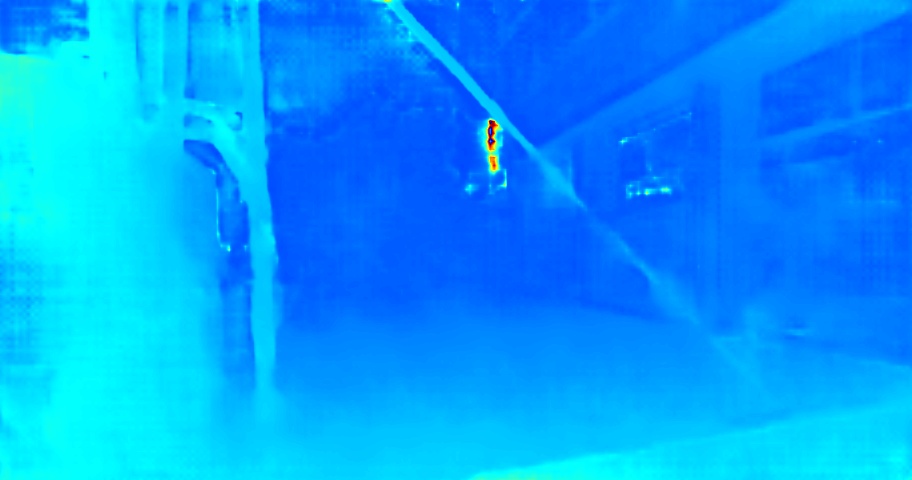}{disp$_{\mathrm{3D} \mathcal{G} \mathrm{exp}}$}
        \plot{5.10\mywidth}{-5.6\mywidth}{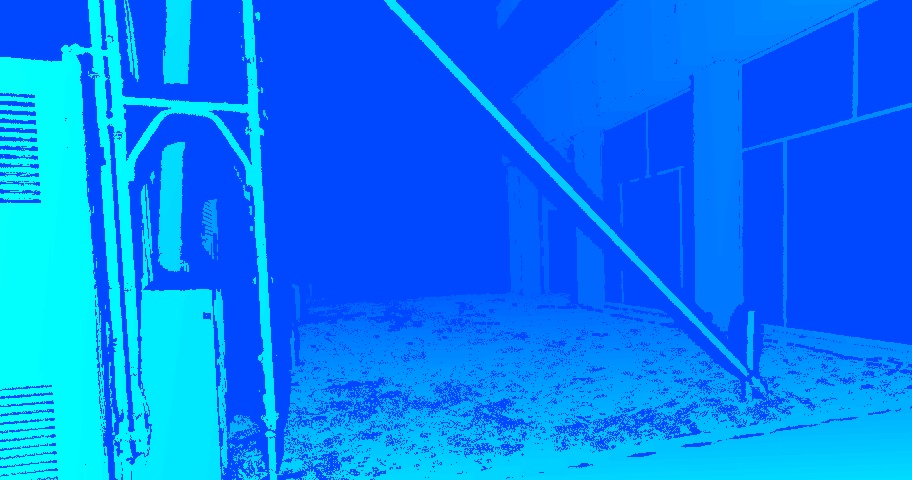}{GT}
        \plot{0}{-6.4\mywidth}{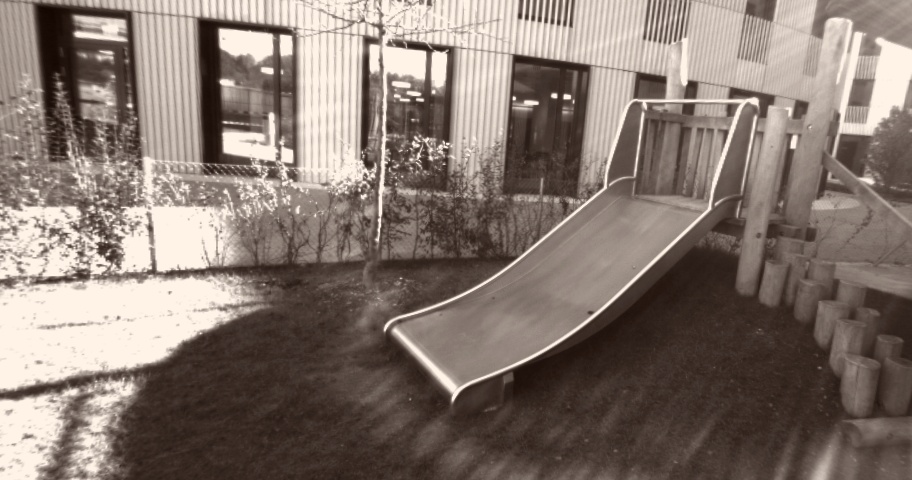}{RGB}
        \plot{1.02\mywidth}{-6.4\mywidth}{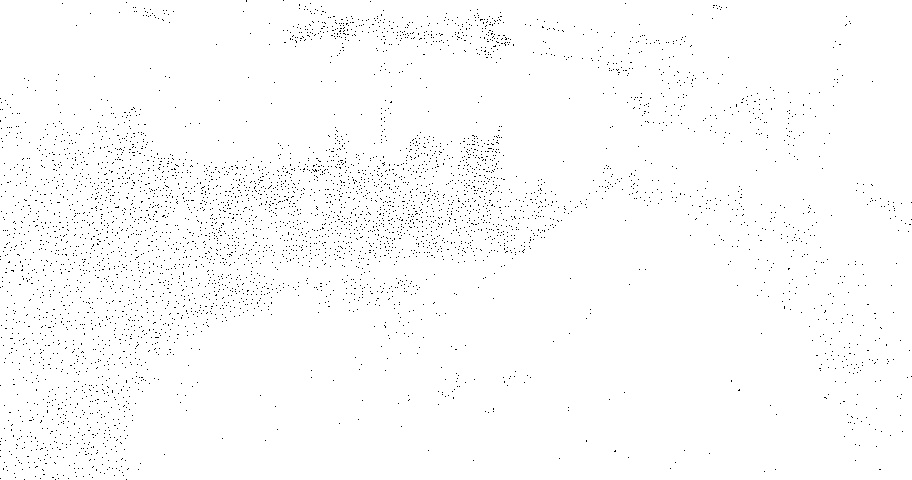}{H}
        \plot{2.04\mywidth}{-6.4\mywidth}{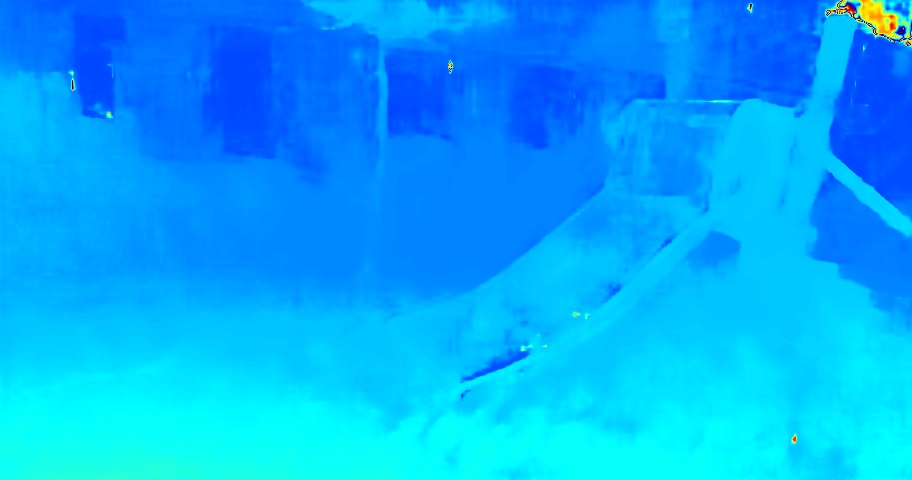}{disp}
        \plot{3.06\mywidth}{-6.4\mywidth}{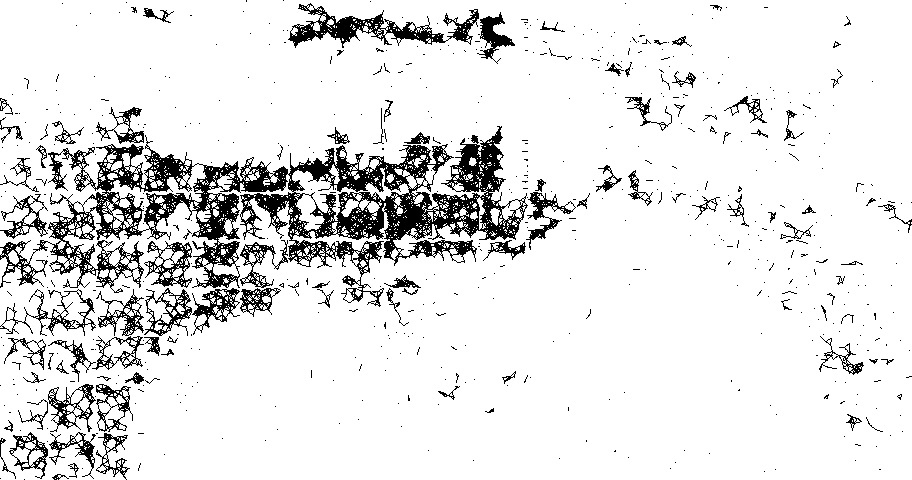}{H$_{\mathbf{3D} \mathcal{G} \mathbf{exp}}$}
        \plot{4.08\mywidth}{-6.4\mywidth}{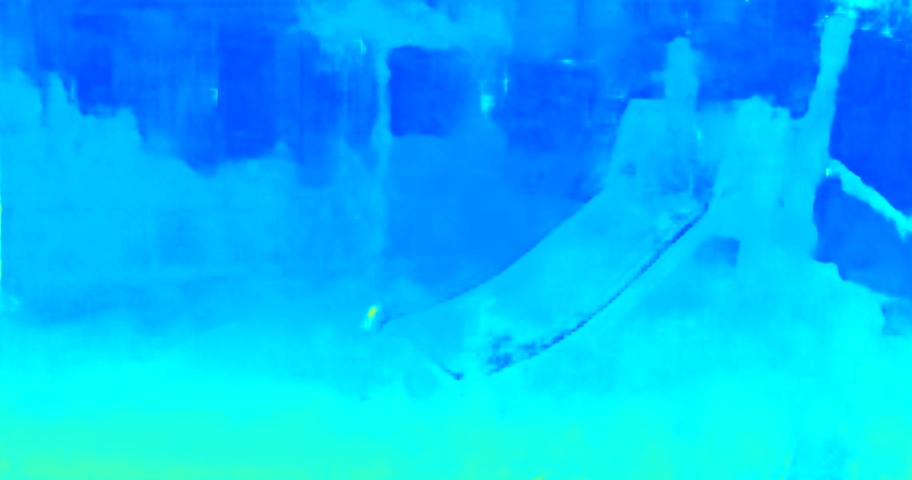}{disp$_{\mathrm{3D} \mathcal{G} \mathrm{exp}}$}
        \plot{5.10\mywidth}{-6.4\mywidth}{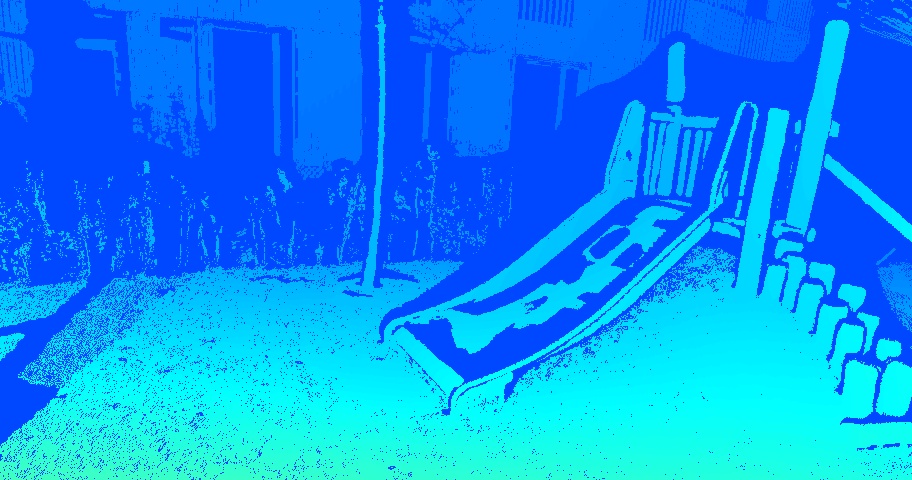}{GT}
    \end{tikzpicture}

    \hfill
    \caption{Qualitative results of PSMNet on {\sc ETH3D} data set. From left to right: $\mathbf{RGB}$ image, $\mathbf{H}$ sparse hints, $\mathbf{disp}$ predicted disparity guided with sparse hints, $\mathbf{H}_{\mathbf{3D} \mathcal{G} \mathbf{exp}}$ hints expanded with our 3D graph, $\mathbf{disp}_{\mathrm{3D} \mathcal{G} \mathrm{exp}}$ predicted disparity guided with expanded hints, $\mathbf{GT}$ disparity ground truth}
    \label{fig:supp_eth_psmnet}
\end{figure*}

\begin{figure*}
    \centering
    \setlength{\mywidth}{0.16\textwidth}%
  
  \newcommand{\plot}[4]{%
    \node[anchor=north west,draw=black!50,minimum width=\mywidth,
      minimum height=.75\mywidth,rounded corners=2pt,path picture={
      \node at (path picture bounding box.north west){
        \includegraphics[width=\mywidth]{#3}};
      }] (fig) at (#1,#2) {};
    \node[anchor=north east,draw=black!50,align=left,inner sep=1pt,
      fill=white,rounded corners=2pt,minimum height=10pt,opacity=.75] at 
      (fig.37) {\footnotesize #4};
  }
     \begin{tikzpicture}[inner sep=0,outer sep=0]  
        \plot{0}{0}{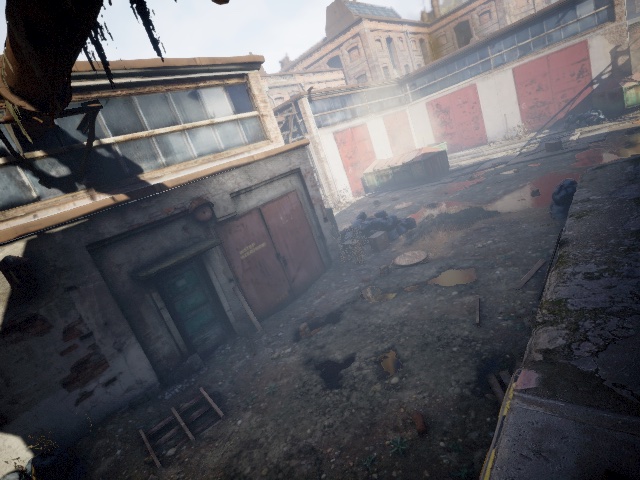}{RGB}
        \plot{1.02\mywidth}{0}{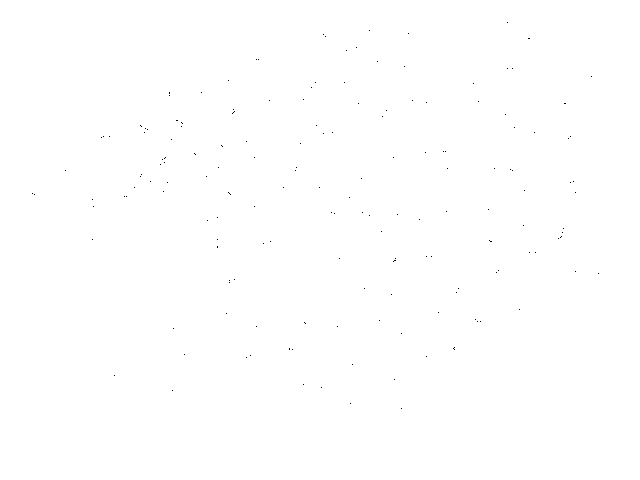}{H}
        \plot{2.04\mywidth}{0}{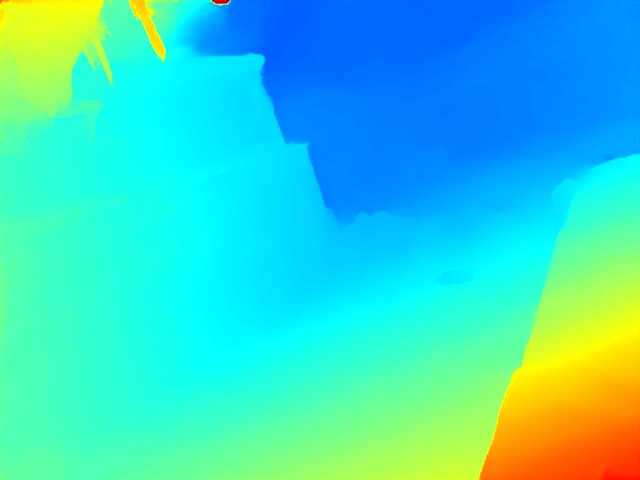}{disp}
        \plot{3.06\mywidth}{0}{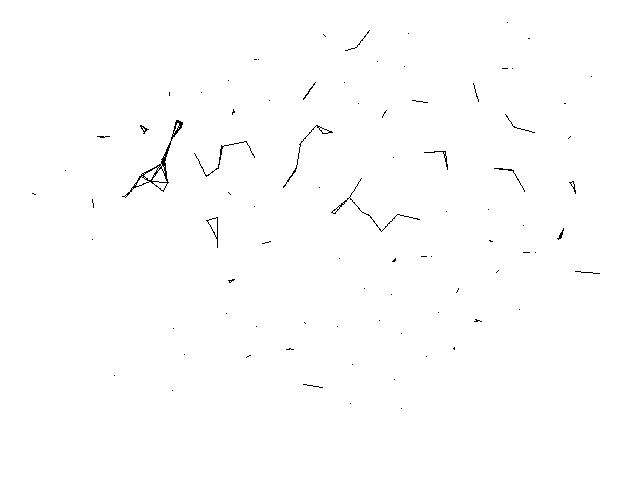}{H$_{\mathbf{3D} \mathcal{G} \mathbf{exp}}$}
        \plot{4.08\mywidth}{0}{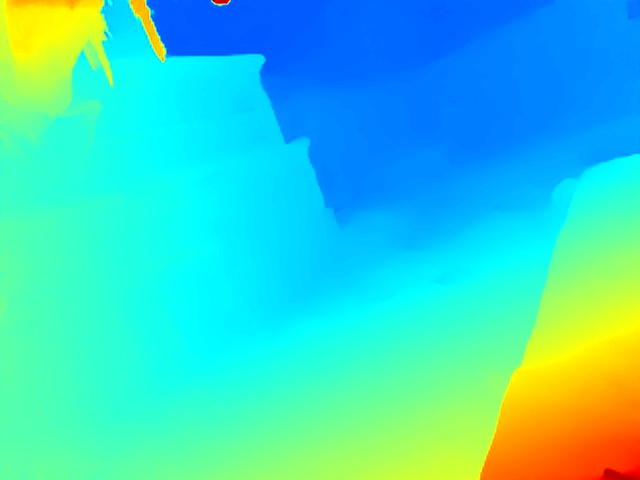}{disp$_{\mathrm{3D} \mathcal{G} \mathrm{exp}}$}
        \plot{5.10\mywidth}{0}{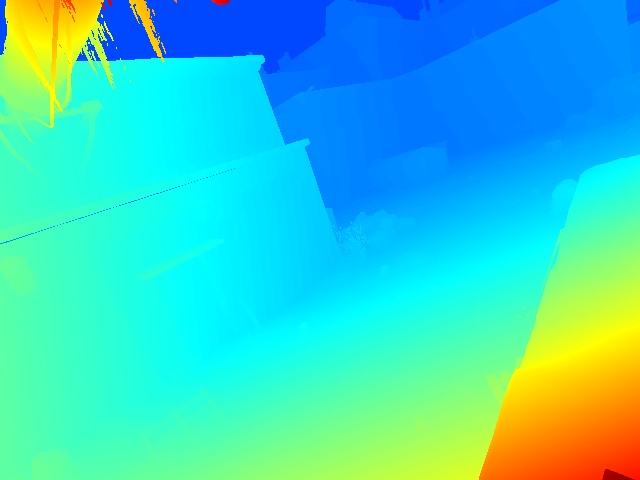}{GT}
        \plot{0}{0.8\mywidth}{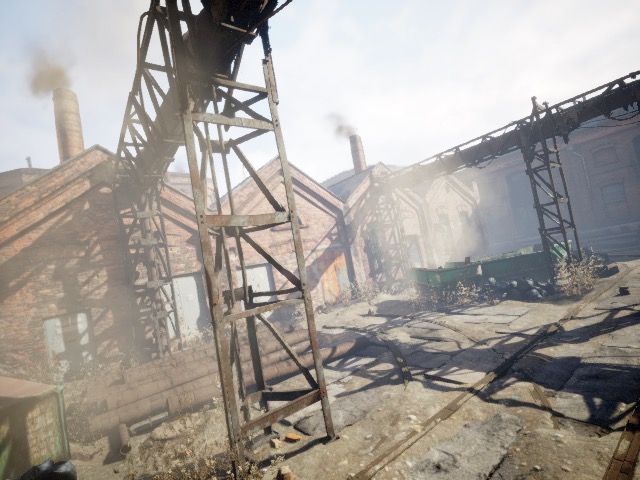}{RGB}
        \plot{1.02\mywidth}{0.8\mywidth}{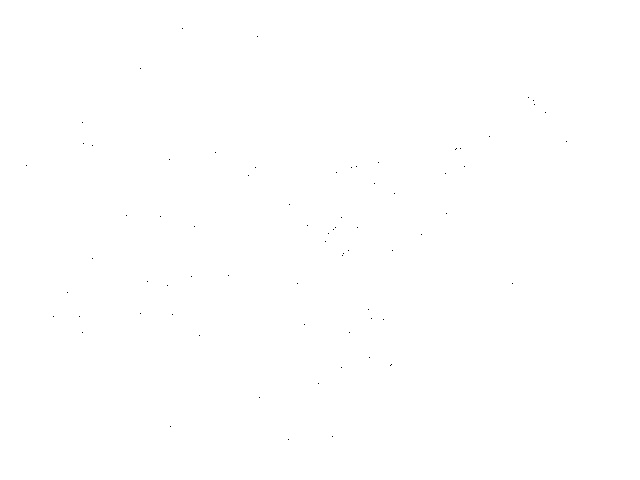}{H}
        \plot{2.04\mywidth}{0.8\mywidth}{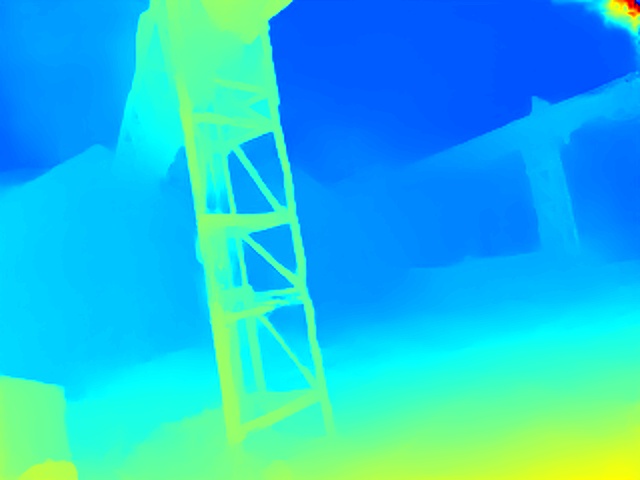}{disp}
        \plot{3.06\mywidth}{0.8\mywidth}{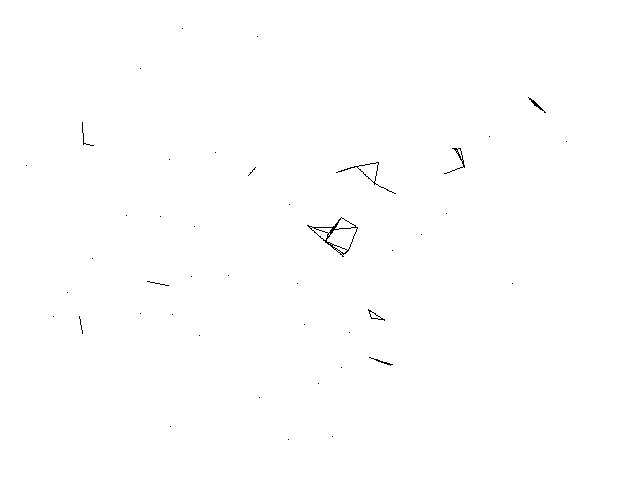}{H$_{\mathbf{3D} \mathcal{G} \mathbf{exp}}$}
        \plot{4.08\mywidth}{0.8\mywidth}{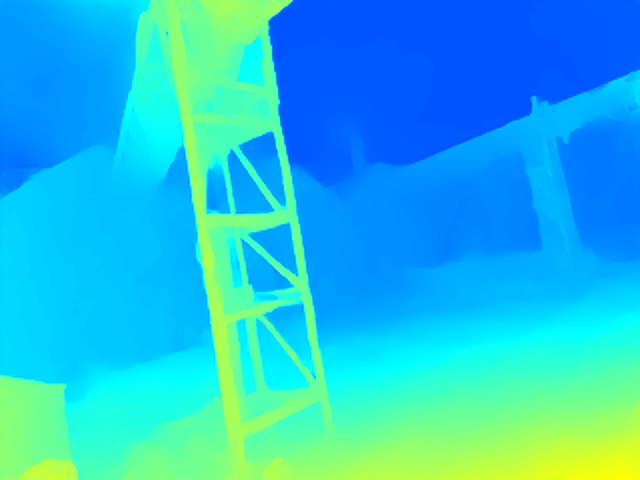}{disp$_{\mathrm{3D} \mathcal{G} \mathrm{exp}}$}
        \plot{5.10\mywidth}{0.8\mywidth}{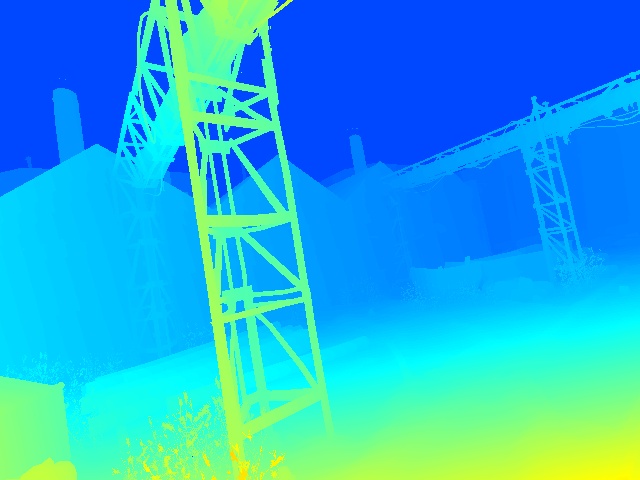}{GT}
        \plot{0}{1.6\mywidth}{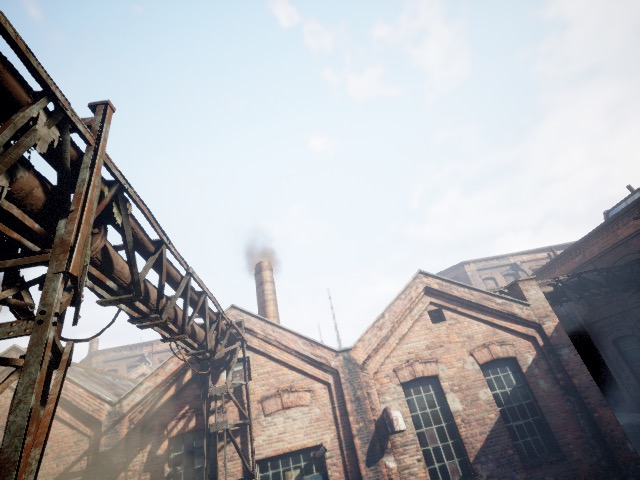}{RGB}
        \plot{1.02\mywidth}{1.6\mywidth}{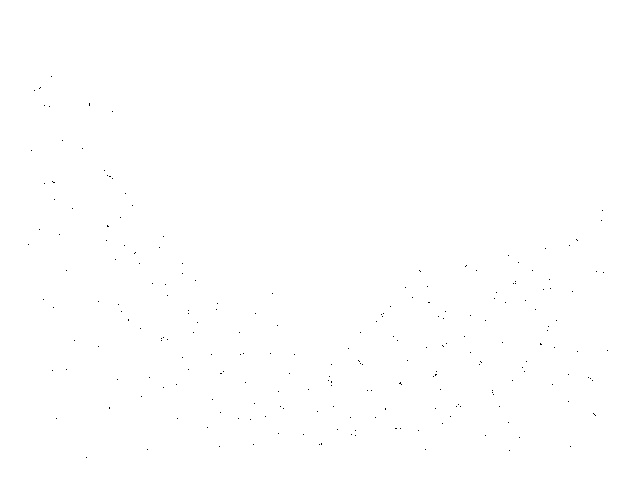}{H}
        \plot{2.04\mywidth}{1.6\mywidth}{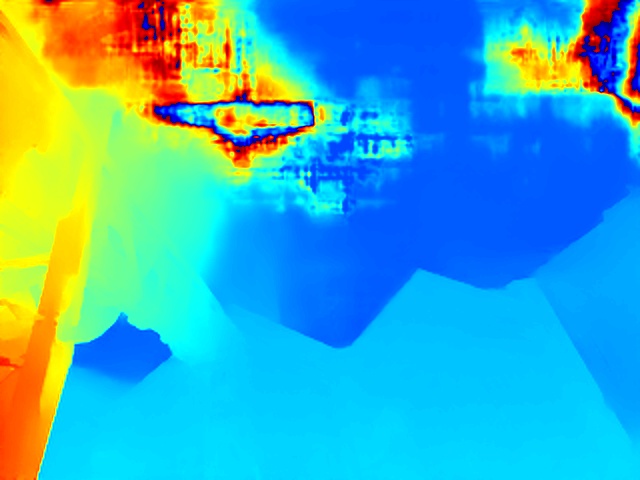}{disp}
        \plot{3.06\mywidth}{1.6\mywidth}{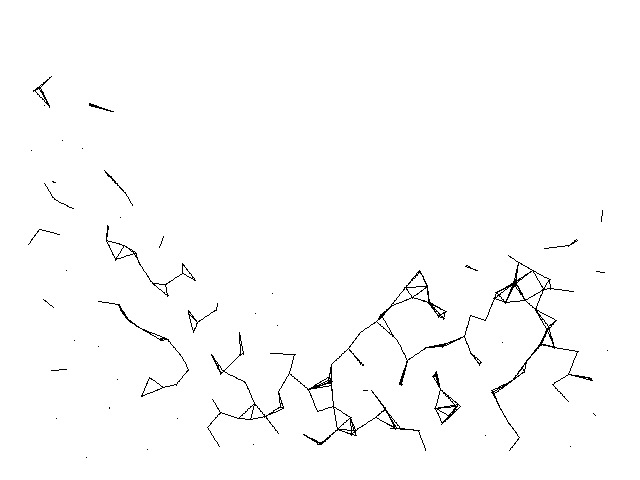}{H$_{\mathbf{3D} \mathcal{G} \mathbf{exp}}$}
        \plot{4.08\mywidth}{1.6\mywidth}{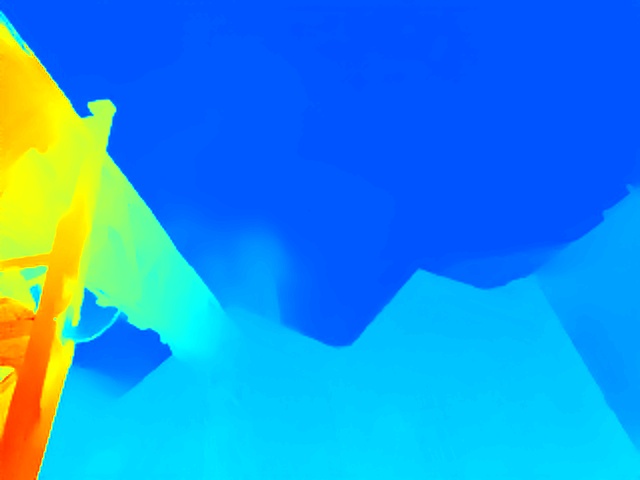}{disp$_{\mathrm{3D} \mathcal{G} \mathrm{exp}}$}
        \plot{5.10\mywidth}{1.6\mywidth}{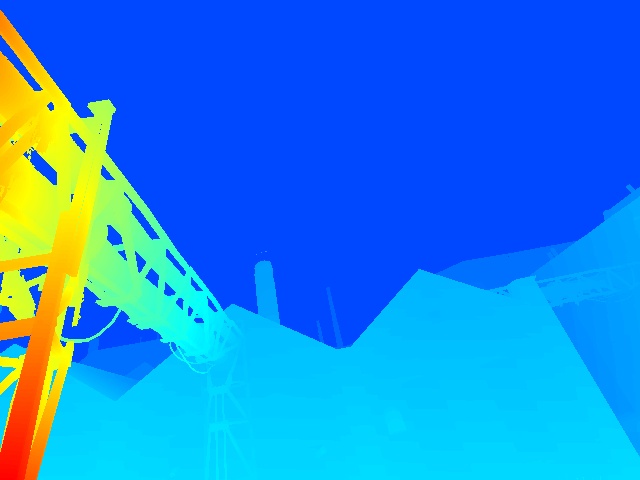}{GT}
        \plot{0}{2.4\mywidth}{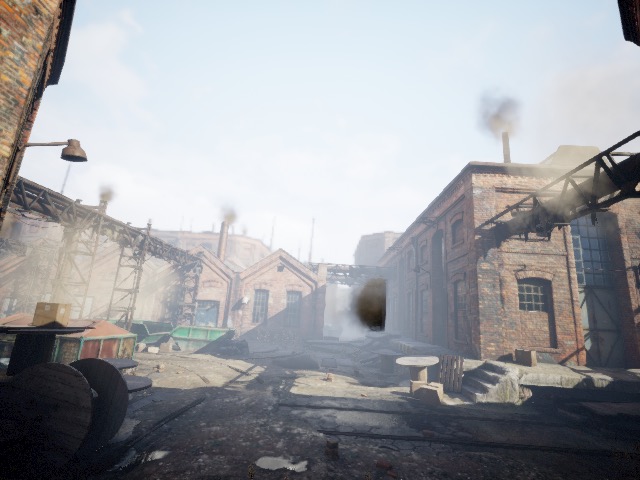}{RGB}
        \plot{1.02\mywidth}{2.4\mywidth}{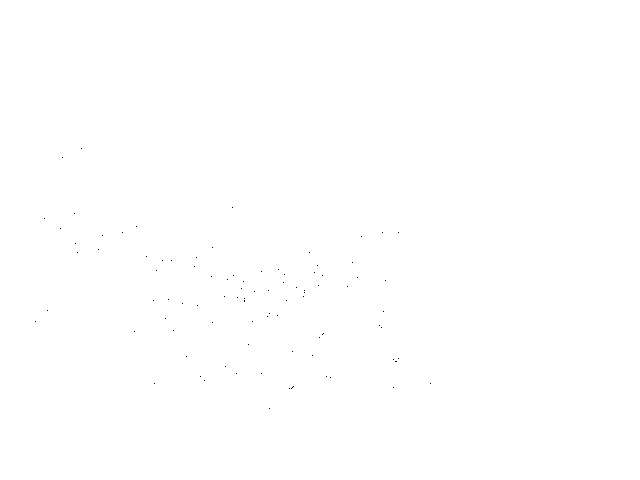}{H}
        \plot{2.04\mywidth}{2.4\mywidth}{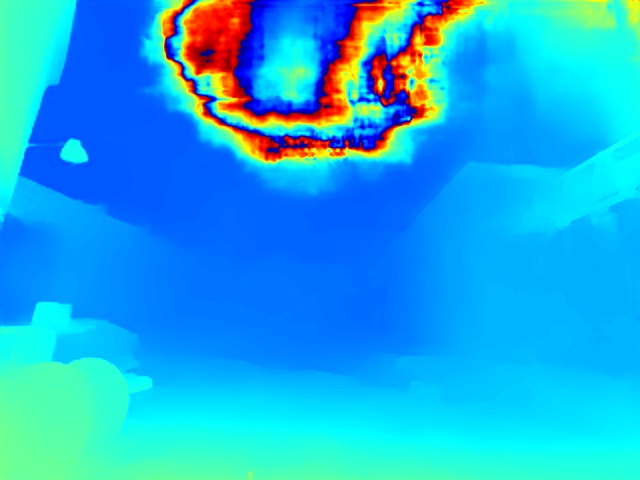}{disp}
        \plot{3.06\mywidth}{2.4\mywidth}{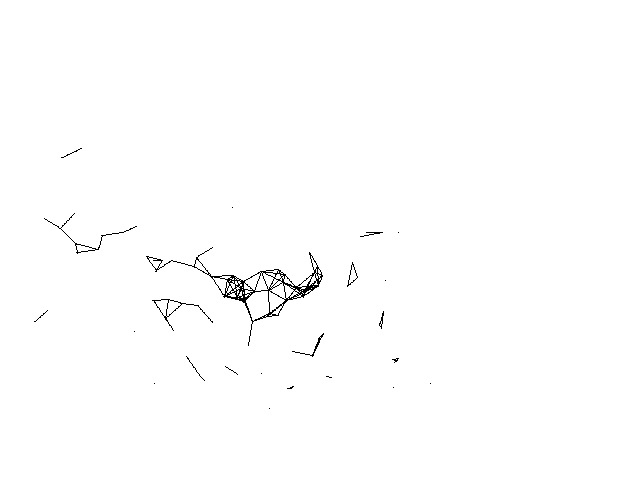}{H$_{\mathbf{3D} \mathcal{G} \mathbf{exp}}$}
        \plot{4.08\mywidth}{2.4\mywidth}{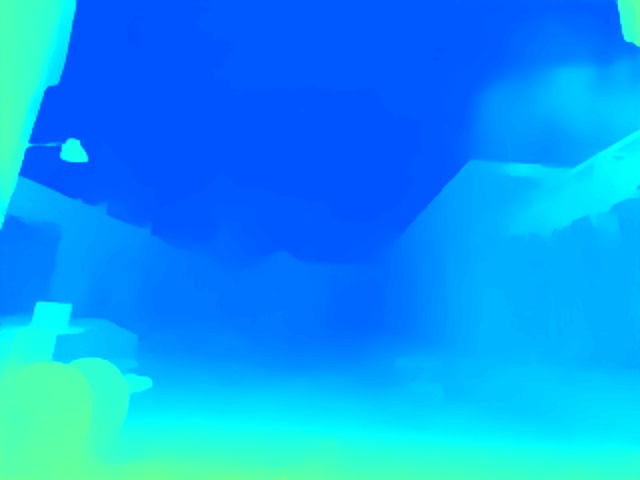}{disp$_{\mathrm{3D} \mathcal{G} \mathrm{exp}}$}
        \plot{5.10\mywidth}{2.4\mywidth}{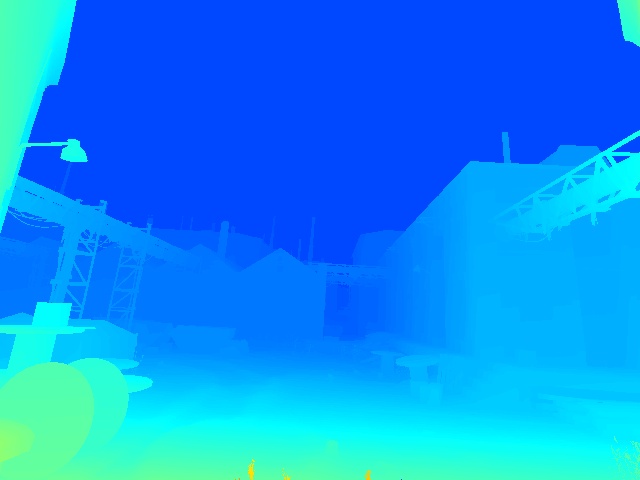}{GT}
        \plot{0}{3.2\mywidth}{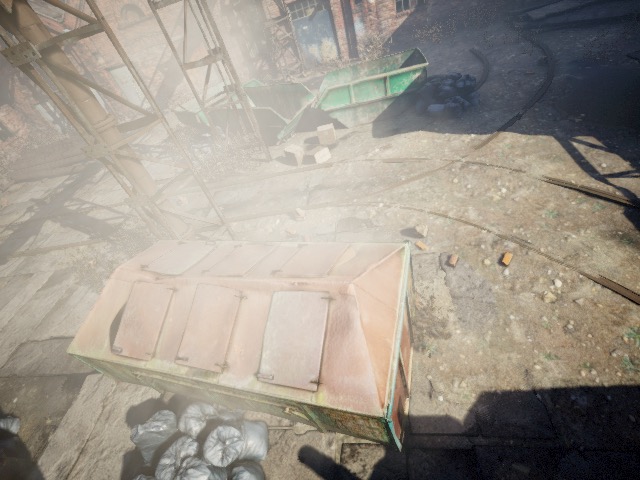}{RGB}
        \plot{1.02\mywidth}{3.2\mywidth}{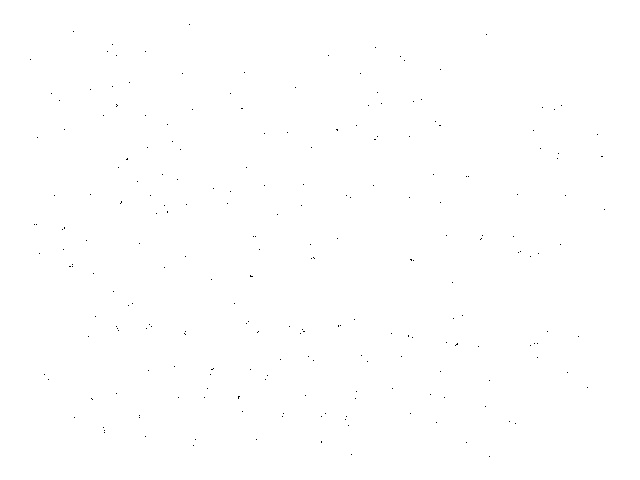}{H}
        \plot{2.04\mywidth}{3.2\mywidth}{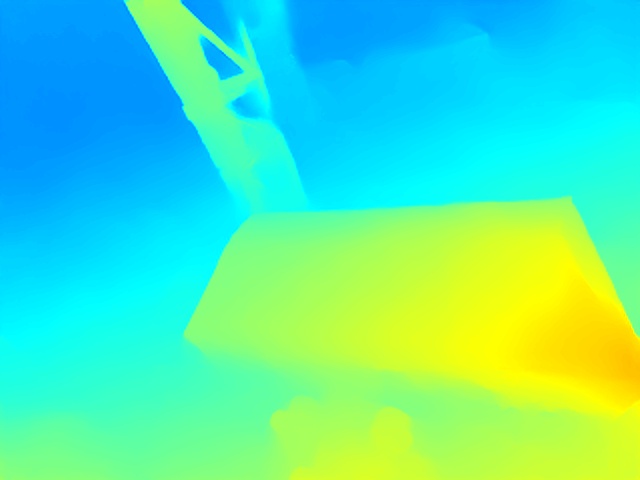}{disp}
        \plot{3.06\mywidth}{3.2\mywidth}{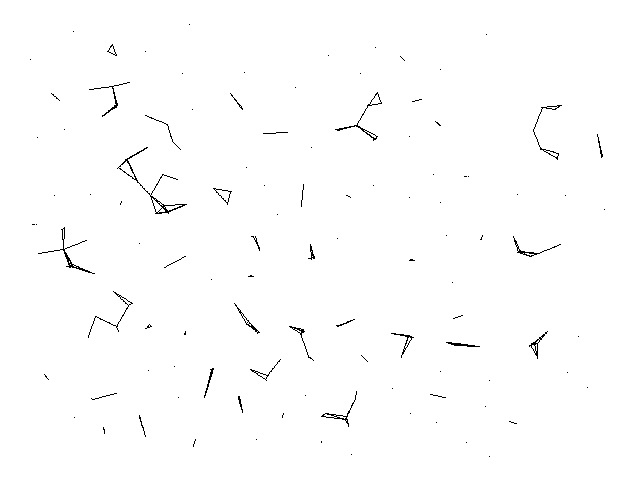}{H$_{\mathbf{3D} \mathcal{G} \mathbf{exp}}$}
        \plot{4.08\mywidth}{3.2\mywidth}{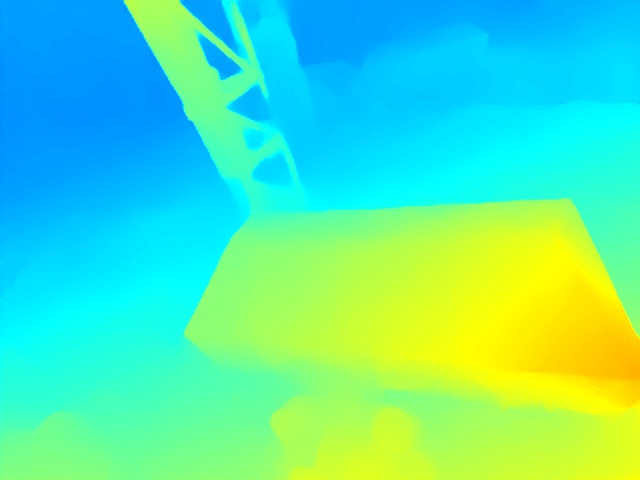}{disp$_{\mathrm{3D} \mathcal{G} \mathrm{exp}}$}
        \plot{5.10\mywidth}{3.2\mywidth}{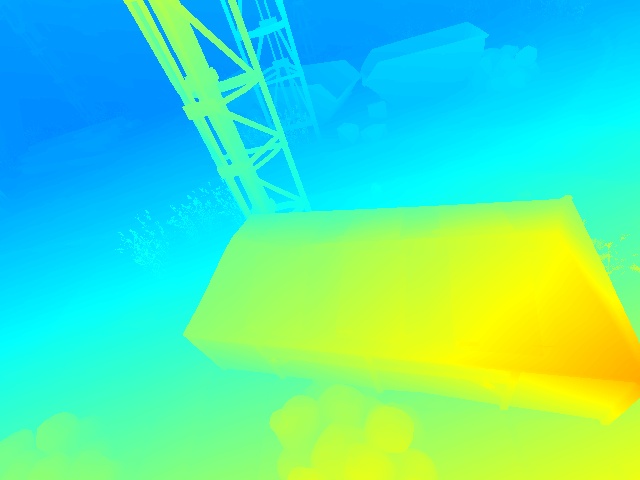}{GT}
        \plot{0}{4\mywidth}{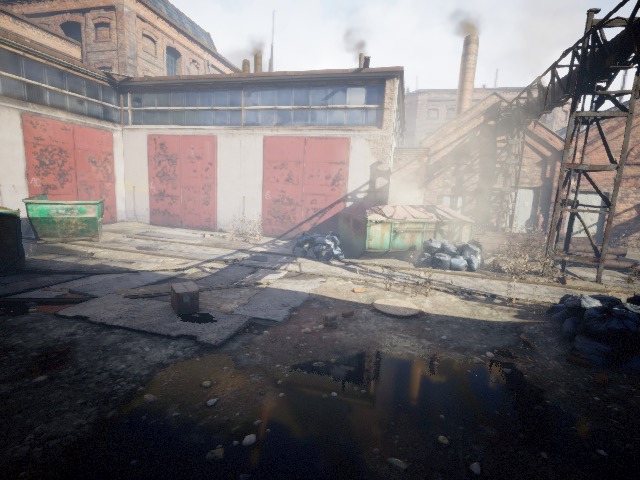}{RGB}
        \plot{1.02\mywidth}{4\mywidth}{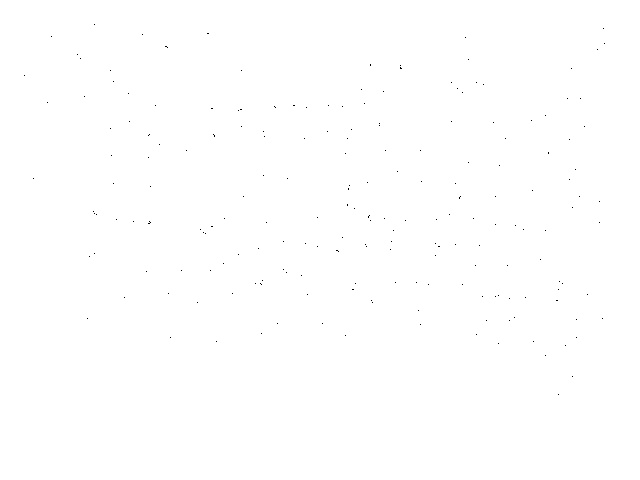}{H}
        \plot{2.04\mywidth}{4\mywidth}{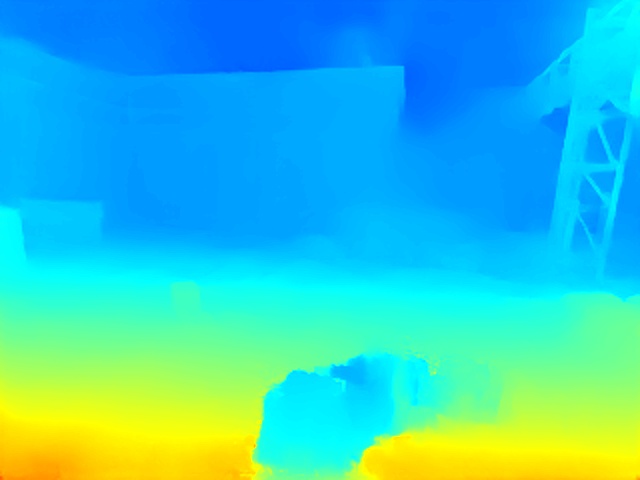}{disp}
        \plot{3.06\mywidth}{4\mywidth}{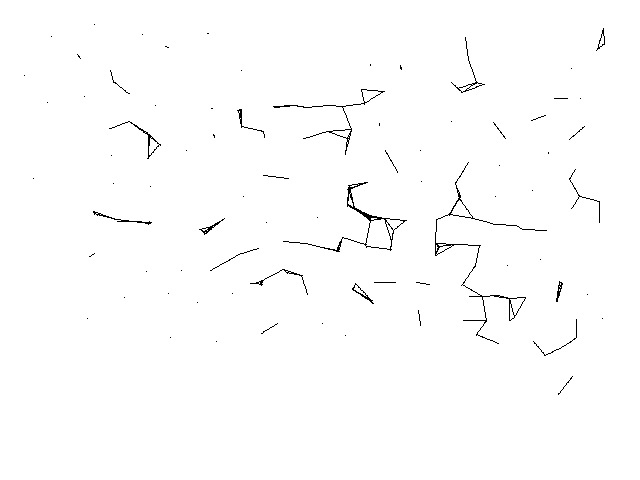}{H$_{\mathbf{3D} \mathcal{G} \mathbf{exp}}$}
        \plot{4.08\mywidth}{4\mywidth}{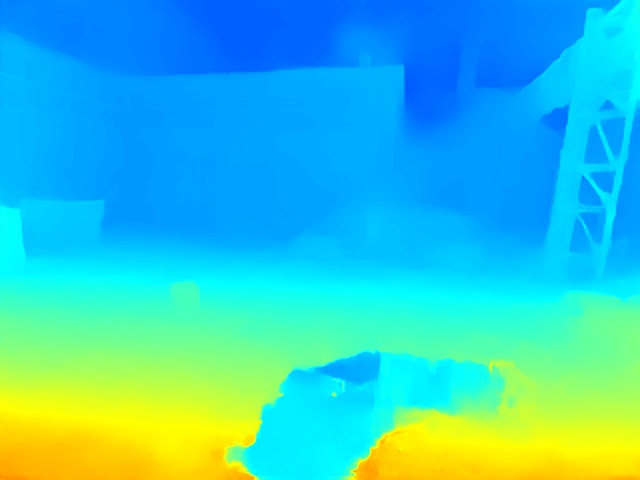}{disp$_{\mathrm{3D} \mathcal{G} \mathrm{exp}}$}
        \plot{5.10\mywidth}{4\mywidth}{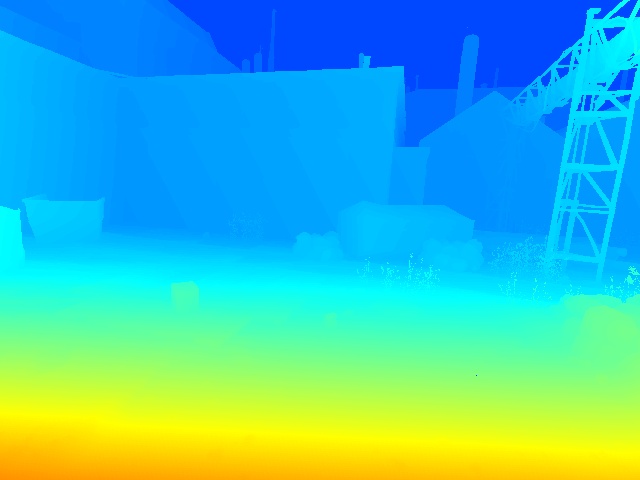}{GT}
        \plot{0}{4.8\mywidth}{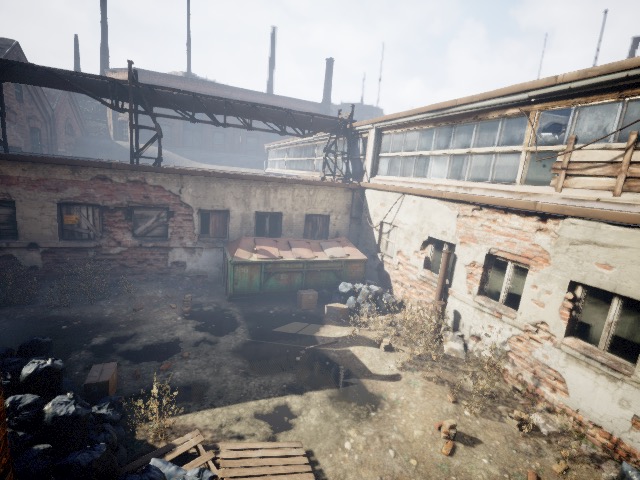}{RGB}
        \plot{1.02\mywidth}{4.8\mywidth}{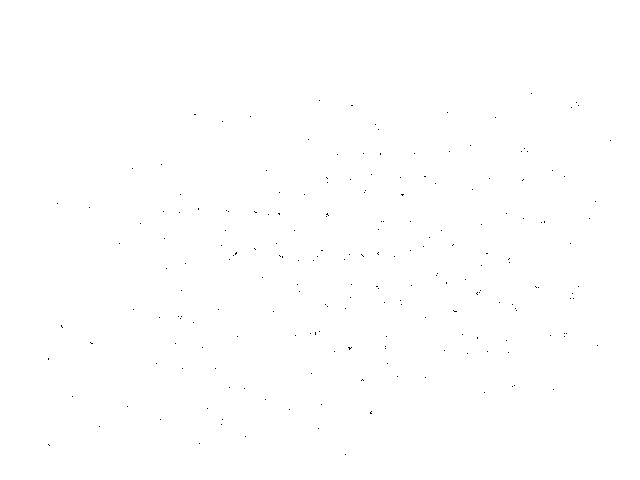}{H}
        \plot{2.04\mywidth}{4.8\mywidth}{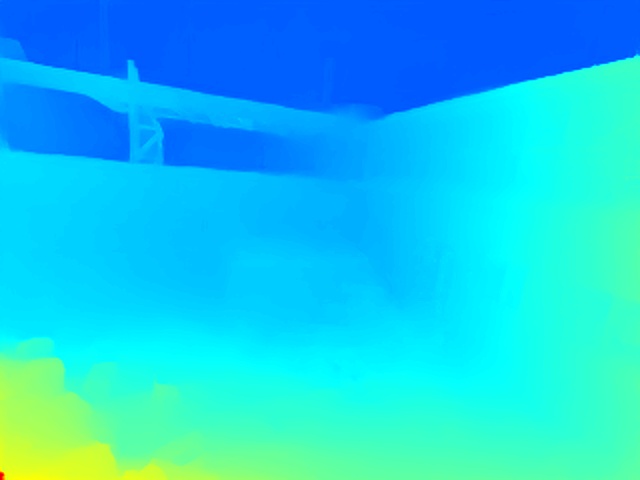}{disp}
        \plot{3.06\mywidth}{4.8\mywidth}{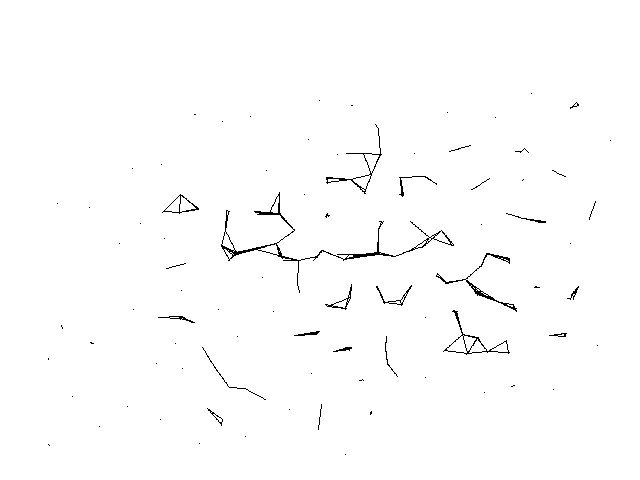}{H$_{\mathbf{3D} \mathcal{G} \mathbf{exp}}$}
        \plot{4.08\mywidth}{4.8\mywidth}{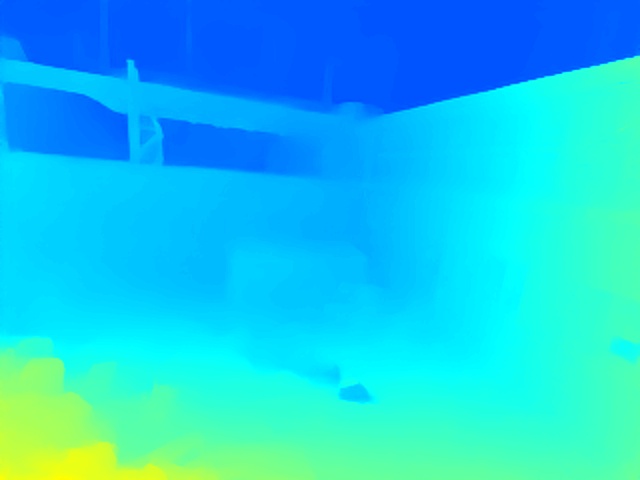}{disp$_{\mathrm{3D} \mathcal{G} \mathrm{exp}}$}
        \plot{5.10\mywidth}{4.8\mywidth}{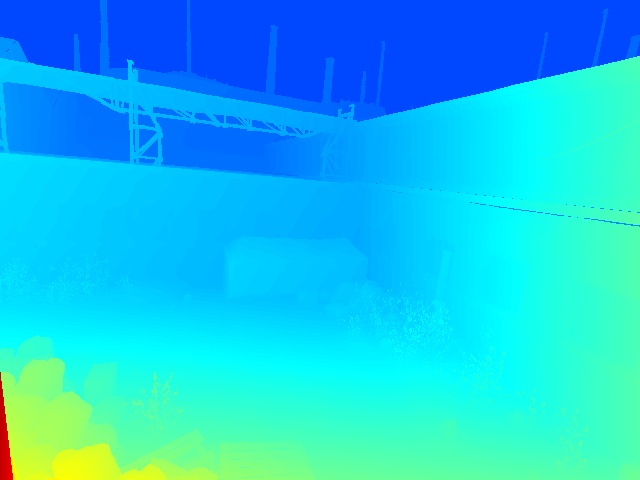}{GT}
        \plot{0}{5.6\mywidth}{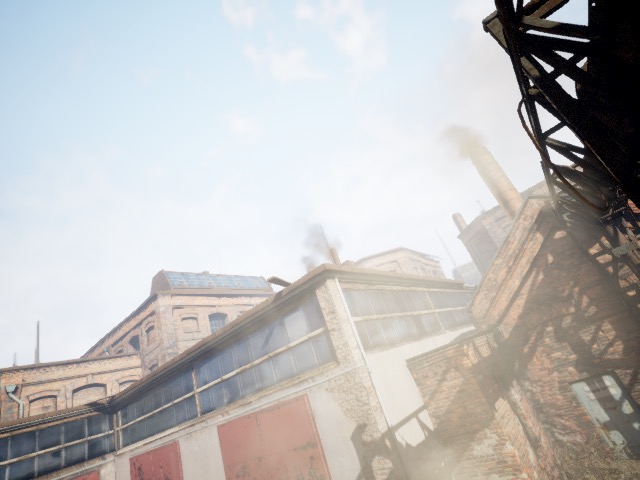}{RGB}
        \plot{1.02\mywidth}{5.6\mywidth}{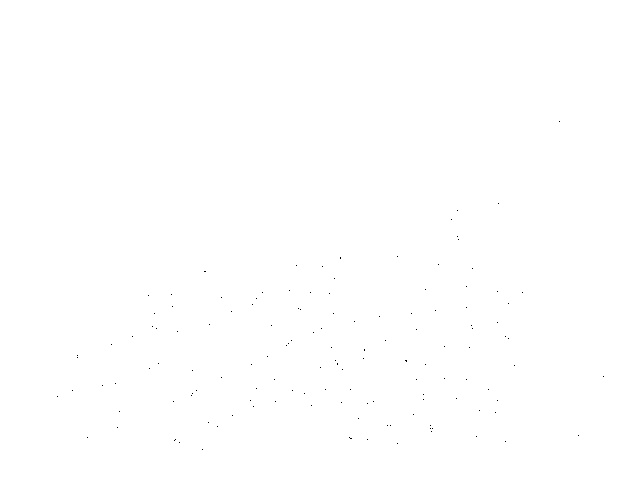}{H}
        \plot{2.04\mywidth}{5.6\mywidth}{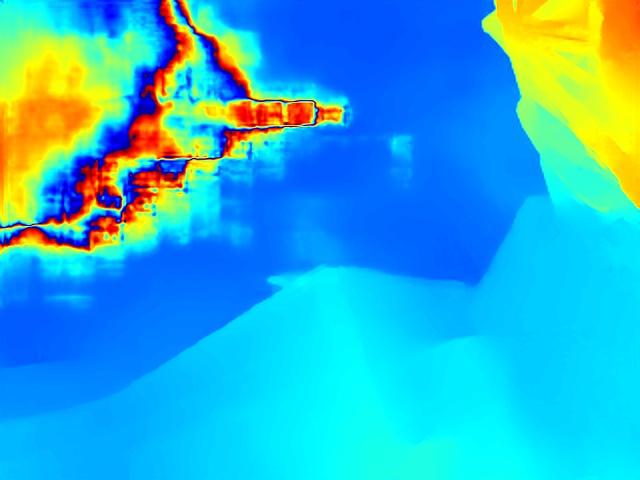}{disp}
        \plot{3.06\mywidth}{5.6\mywidth}{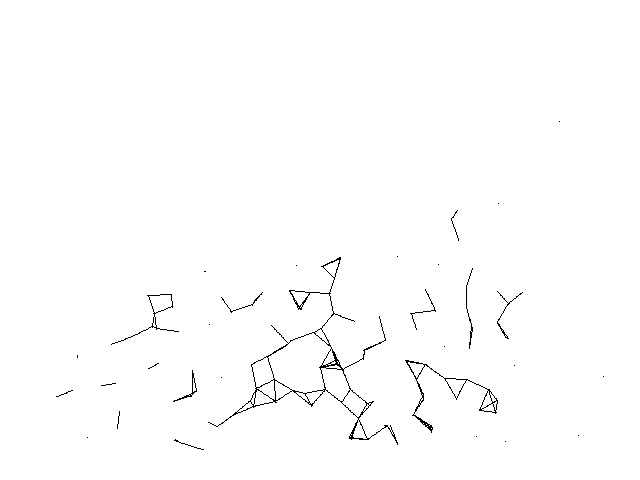}{H$_{\mathbf{3D} \mathcal{G} \mathbf{exp}}$}
        \plot{4.08\mywidth}{5.6\mywidth}{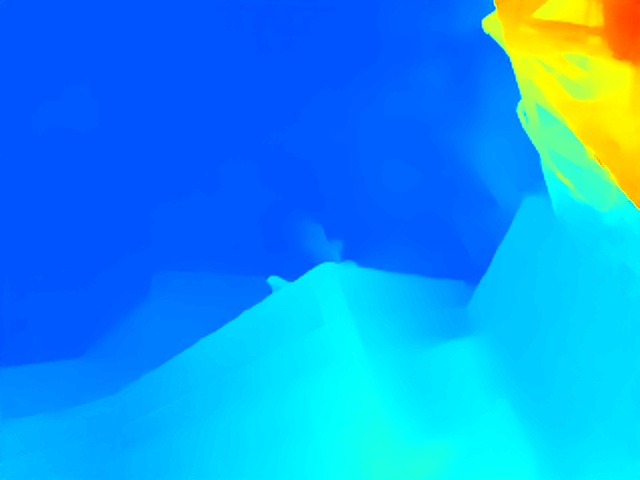}{disp$_{\mathrm{3D} \mathcal{G} \mathrm{exp}}$}
        \plot{5.10\mywidth}{5.6\mywidth}{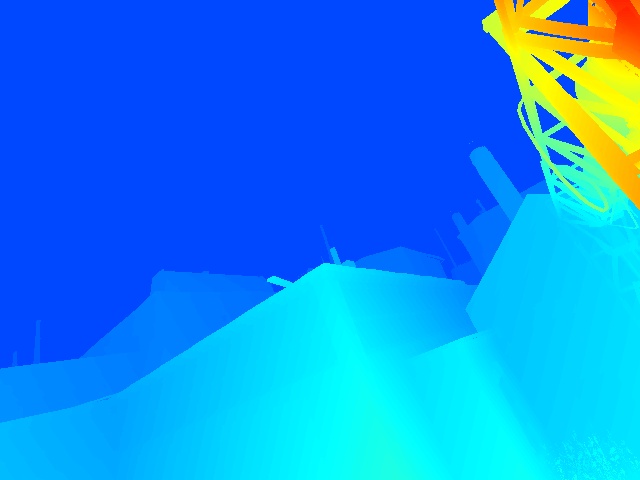}{GT}
    \end{tikzpicture}

    \caption{Qualitative results of DeepPruner on {\sc Tartan} data set. From left to right: $\mathbf{RGB}$ image, $\mathbf{H}$ sparse hints, $\mathbf{disp}$ predicted disparity guided with sparse hints, $\mathbf{H}_{\mathbf{3D} \mathcal{G} \mathbf{exp}}$ hints expanded with our 3D graph, $\mathbf{disp}_{\mathrm{3D} \mathcal{G} \mathrm{exp}}$ predicted disparity guided with expanded hints, $\mathbf{GT}$ disparity ground truth}
    \label{fig:supp_tartan_deeppruner}
\end{figure*}

Additional qualitative results are presented in \cref{fig:supp_eth_psmnet} for {\sc ETH3D} data set and in \cref{fig:supp_tartan_deeppruner} for {\sc Tartan} data set. From the sparse hints $\mathbf{H}$ visualizations, it is easy to grasp how sparse the hints are, in particular for {\sc Tartan}. This is even more evident from the densified hints $\mathbf{H}_{\mathrm{3D} \mathcal{G} \mathrm{exp}}$ where some areas of {\sc ETH3D} become dense, while it is never the case for {\sc Tartan}. Overall, the hints remove areas with large errors, that we believe may be caused by domain-shift, and also sharpen the details of some predictions. 
We provide a 3D point cloud visualization of \cref{fig:teaser} in image \cref{fig:supp_pointcloud}, note how the artifacts are greatly reduced with our proposed guidance and the prediction range does not explode remaining similar to GT.

\end{document}